\documentclass[10pt,twocolumn,letterpaper]{article}

\usepackage{iccv}              % To produce the CAMERA-READY version

\usepackage{times}
\usepackage[numbers,sort,compress]{natbib}
\usepackage{graphicx}
\usepackage{amsmath}
\usepackage{amssymb}
\usepackage{booktabs}
\usepackage{multirow}
\usepackage{siunitx}
\usepackage[nice]{nicefrac}
\usepackage{caption}

\RequirePackage{algorithm}
\usepackage[noend]{algpseudocode}               % ADDED THIS

\usepackage[pagebackref,breaklinks,colorlinks]{hyperref}

\usepackage{tikz}
\usetikzlibrary{positioning}
\usepackage{xcolor}

\newif\ifcomments
\commentstrue

\newif\ifarxiv
\arxivfalse

\newcommand{\position}{\boldsymbol{x}}
\newcommand{\positionDom}{\mathcal{X}}
\newcommand{\density}{\boldsymbol{r}}
\newcommand{\densityDom}{\mathcal{R}}
\newcommand{\nerfNet}{\boldsymbol{\nu}}
\newcommand{\pointEncoderParams}{{\boldsymbol{w}}}
\newcommand{\pointEncoderParamsDom}{{\mathcal{W}}}
\newcommand{\pointEncoder}{\boldsymbol{\gamma}}
\newcommand{\encoderRange}{\boldsymbol{\Gamma}}

\newcommand{\textToken}{\boldsymbol{c}}
\newcommand{\textTokenDom}{\mathcal{C}}

\newcommand{\textEmbedding}{\boldsymbol{z}}
\newcommand{\vectorEmbedding}{\boldsymbol{v}}
\newcommand{\mappingNet}{\boldsymbol{m}}%\boldsymbol{\zeta}}
\newcommand{\noise}{\boldsymbol{\epsilon}}

\newcommand{\loss}{\mathcal{L}}

\usepackage[capitalize]{cleveref}
\crefname{section}{Sec.}{Secs.}
\Crefname{section}{Section}{Sections}
\Crefname{table}{Table}{Tables}
\crefname{table}{Tab.}{Tabs.}

\iccvfinalcopy % *** Uncomment this line for the final submission
\def\iccvPaperID{5316} % *** Enter the ICCV Paper ID here

\ificcvfinal\fi

\newcommand{\ourTitle}{ATT3D: Amortized Text-to-3D Object Synthesis}
\begin{document}
    
    \title{\ourTitle}
    
    \author{
        Jonathan Lorraine \quad Kevin Xie \quad Xiaohui Zeng \quad Chen-Hsuan Lin \quad Towaki Takikawa \\
        Nicholas Sharp \quad Tsung-Yi Lin \quad Ming-Yu Liu \quad Sanja Fidler \quad James Lucas \\
        NVIDIA Corporation\\
        %{\tt\small us@nvidia.com}
        %\and
        %Us 2\\
    }
    \twocolumn[{
        \renewcommand\twocolumn[1][]{#1}
        \maketitle
        \begin{center}

    \vspace{-0.035\textheight}
    \centering
    \begin{tikzpicture}
        \centering
        % \hspace{-0.05\textwidth}
        % \node (img10){\includegraphics[trim={1.5cm .5cm 7.75cm 2.25cm},clip,width=.01\linewidth]{cvpr2023-author_kit-v1_1-1/latex/images/squirrel_comp.png}};
        \node (img11){\includegraphics[trim={.2cm .5cm .2cm .5cm},clip,width=.98\linewidth]{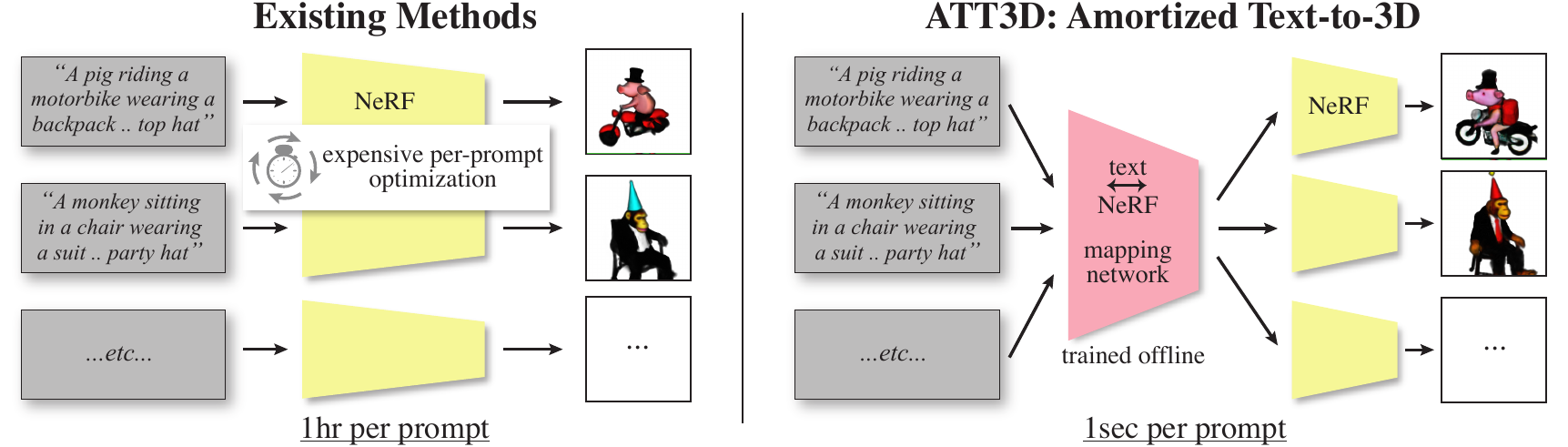}};

        \node[above=of img11, node distance=0cm, xshift=-4cm, yshift=-1.2cm,font=\color{black}]{Existing Methods};
        \node[above=of img11, node distance=0cm, xshift=4cm, yshift=-1.2cm,font=\color{black}]{ATT3D: Amortized Text-to-3D};

        %\node[below=of img11, node distance=0cm, xshift=-4cm, yshift=1.2cm,font=\color{black}]{Requires 8 GPUs and 1 hour};
        %\node[below=of img11, node distance=0cm, xshift=4cm, yshift=1.2cm,font=\color{black}]{Requires 1 GPU and $< 1$ sec};
        \node[below=of img11, node distance=0cm, xshift=-4cm, yshift=1.1cm,font=\color{black}]{Requires 1 hour};
        \node[below=of img11, node distance=0cm, xshift=4cm, yshift=1.1cm,font=\color{black}]{Requires $< 1$ sec};
        
        %\node [right=of img11, xshift=-1.25cm](img12){\includegraphics[trim={0cm 0cm 0cm 0cm},clip,width=.45\linewidth]{images/teaserV2/amortized.png}};
    \end{tikzpicture}
    \vspace{-0.015\textheight}
    \captionof{figure}{
        Our method initially trains one network to output 3D objects consistent with various text prompts.
        After, when we receive an unseen prompt, we produce an accurate object in $< \num{1}$ second, with $1$ GPU.
        Existing methods re-train the entire network for every prompt, requiring a long delay for the optimization to complete.
        Further, we can interpolate between prompts for user-guided asset generation (Fig.~\ref{fig:interpolation}).
        We include a \href{https://research.nvidia.com/labs/toronto-ai/ATT3D/}{project webpage} with an overview and videos.
        %We show qualitative results for assorted animal prompts, with full results in Sec.~\ref{sec:generalizeNewPrompt} and Fig.~\ref{fig:all_quantitative}.
        %\textbf{Takeaway:}
        %    By training a single model on many text prompts we are able to generalize to unseen prompts without any extra optimization.
        %We show quantitative results for this set of $2400$ animal prompts in Fig.~\ref{fig:all_quantitative}, where we achieve a higher quality for any compute budget on both training \& testing prompts.
        %Notably, when training on only ${\color{blue}50\%}$ or ${\color{green}12.5\%}$ of the prompts, the testing prompts-- which cost no optimization -- perform favorably with respect to the {\color{red}single-prompt} method, which must optimize on the test data.
    }
    \vspace{-0.00\textheight}
    \label{fig:method_teaser}
        \end{center}
    }]
    %\maketitle
    
    \vspace{-0.005\textheight}
\begin{abstract}\vspace{-0.01\textheight}
    Text-to-3D modelling has seen exciting progress by combining generative text-to-image models with image-to-3D methods like Neural Radiance Fields.
    DreamFusion recently achieved high-quality results but requires a lengthy, per-prompt optimization to create 3D objects.
    To address this, we amortize optimization over text prompts by training on many prompts simultaneously with a unified model, instead of separately.
    With this, we share computation across a prompt set, training in less time than per-prompt optimization.
    Our framework -- Amortized text-to-3D (ATT3D) -- enables knowledge sharing between prompts to generalize to unseen setups and smooth interpolations between text for novel assets and simple animations.
\end{abstract}
\vspace{-0.02\textheight}
    %%%%%%%%% BODY TEXT
\vspace{-0.005\textheight}
\section{Introduction}\label{sec:intro}
\vspace{-0.005\textheight}
    3D content creation is important because it allows for more immersive and engaging experiences in industries such as entertainment, education, and marketing.
    However, 3D design is challenging due to technical complexity of the 3D modeling software, 
    %, hardware requirements,
    and the artistic skills required to create high-quality models and animations.
    Text-to-3D (TT3D) generative tools have the potential to democratize 3D content creation by relieving these limitations.
    To make this technology successful, we desire tools that provide fast responses to users while being inexpensive for the operator.
    %Making these real-time will \TODO{TALK IN MEETING....} %: talk about making seamless content generation tools, or interactive objects in digital worlds

    %\Jon{Topic: Problems with current methods for interactive TT3D}
    Recent TT3D methods~\citep{poole2022dreamfusion, lin2022magic3d} allow users to generate high-quality 3D models from text-prompts but use a lengthy ($\sim \!\! \num{15}$ minute to $>\!\! \num{1}$ hour~\citep{lin2022magic3d, poole2022dreamfusion}) per-prompt optimization.
    Having users wait between each iteration of prompt engineering results in a sporadic and time-consuming design process.
    Further, generation for a new prompt requires multiple GPUs and uses large text-to-image models~\citep{balaji2022ediffi, rombach2022high, saharia2022photorealistic}, creating a prohibitive cost for the pipeline operator.

    %\Jon{Topic: Our solution for interactive TT3D}
    %Our method, Amortized Text-to-3D (ATT3D), trains one network to output NeRFs representing 3D objects consistent with various text prompts.
    We split the TT3D process into two stages.
    First, we optimize one model offline to generate 3D objects for many different text prompts simultaneously.
    This \emph{amortizes optimization} over the prompts, by sharing work between similar instances.
    The second, user-facing stage uses our amortized model in a simple feed-forward pass to quickly generate an object given text, with no further optimization required.
    %We train our model offline on many different setups.
    %\Kevin{By training simultaneously on many different prompts, work can be shared between similar prompts leading to faster training overall.}

     %\Jon{Topic: Benefits of our method}
    Our method, Amortized text-to-3D (ATT3D), produces a model which can generate an accurate 3D object in $< \num{1}$ second, with only $1$ consumer-grade GPU.
    This TT3D pipeline can be deployed more cheaply, with a real-time user experience.
    Our offline stage trains the ATT3D model significantly faster than optimizing prompts individually while retaining or even surpassing quality, by leveraging compositionality in the parts underlying each 3D object.
    We also gain a new user-interaction ability to interpolate between prompts for novel asset generation and animations.
    %Further, the amortized framework allows us to use add extra inputs $x$ to our ATT3D mapping, resulting in a (text+$x$)-to-3D mapping.
    %We use this to make other design choices -- like our loss function choice, or how we want to interpolate between prompts -- produce results quickly.
    % \SF{We show that the offline stage trains our ATT3D model significantly faster than running optimization per individual prompt, while retaining or even surpassing quality. }
    % \SF{Here we should also highlight compositionality?}

    \begin{figure*}%[h!]
    \vspace{-0.04\textheight}
    \centering
    \begin{tikzpicture}
        \centering
        \node (img1){\includegraphics[trim={.95cm 0.0cm .0cm .95cm},clip,width=.48\linewidth]{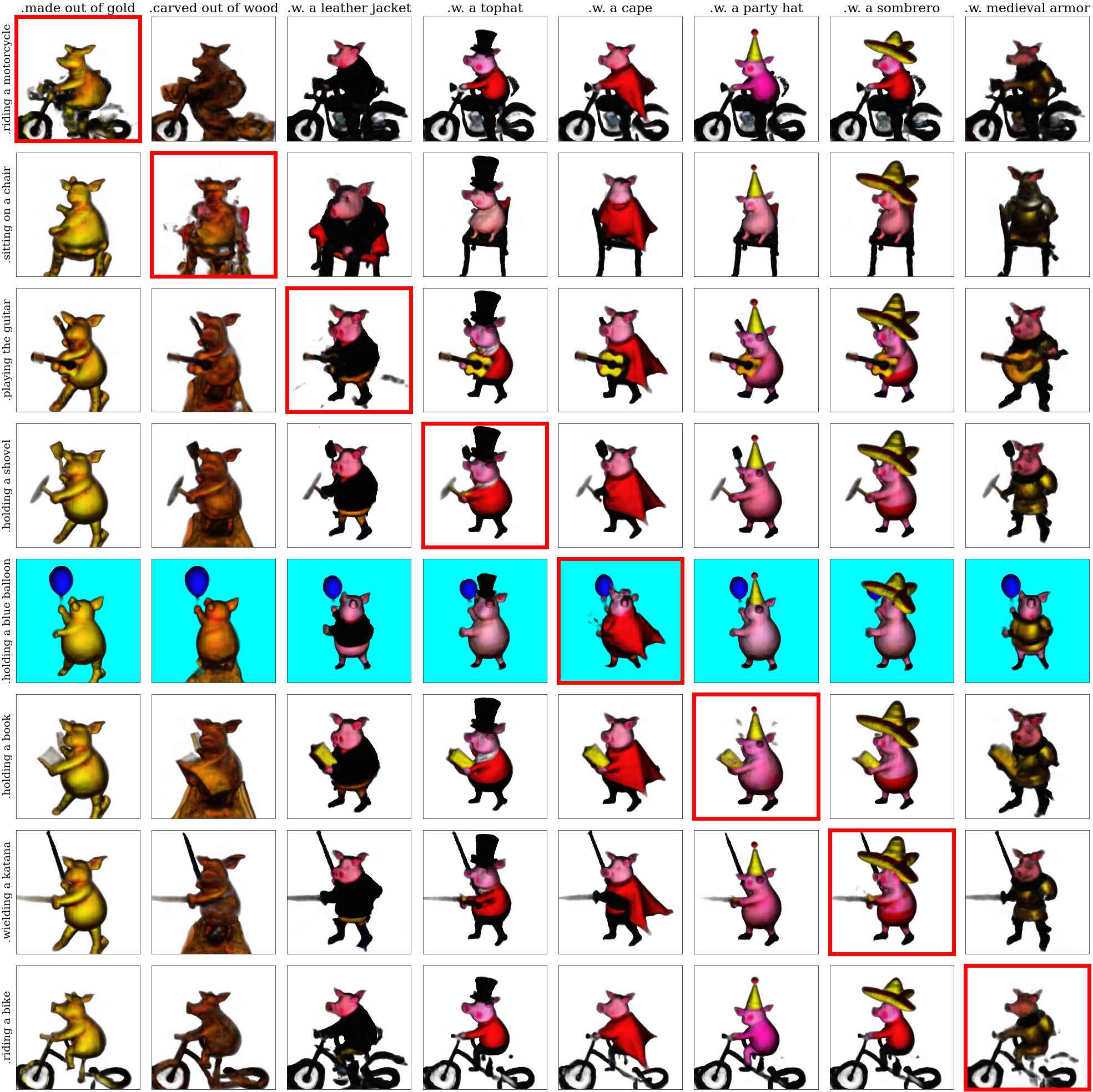}};
        \node[above=of img1, node distance=0cm, xshift=-.1cm, yshift=-1.2cm,font=\color{black}]{ATT3D};
        
        %\node [right=of img1, xshift=-1cm](img2){\includegraphics[trim={.0cm .0cm .0cm .0cm},clip,width=.48\linewidth]{images/compositional/compositional_single_prompt.png}};
        
        \node [right=of img1, xshift=-1cm](img2){\includegraphics[trim={.95cm 0.0cm .0cm .95cm},clip,width=.48\linewidth]{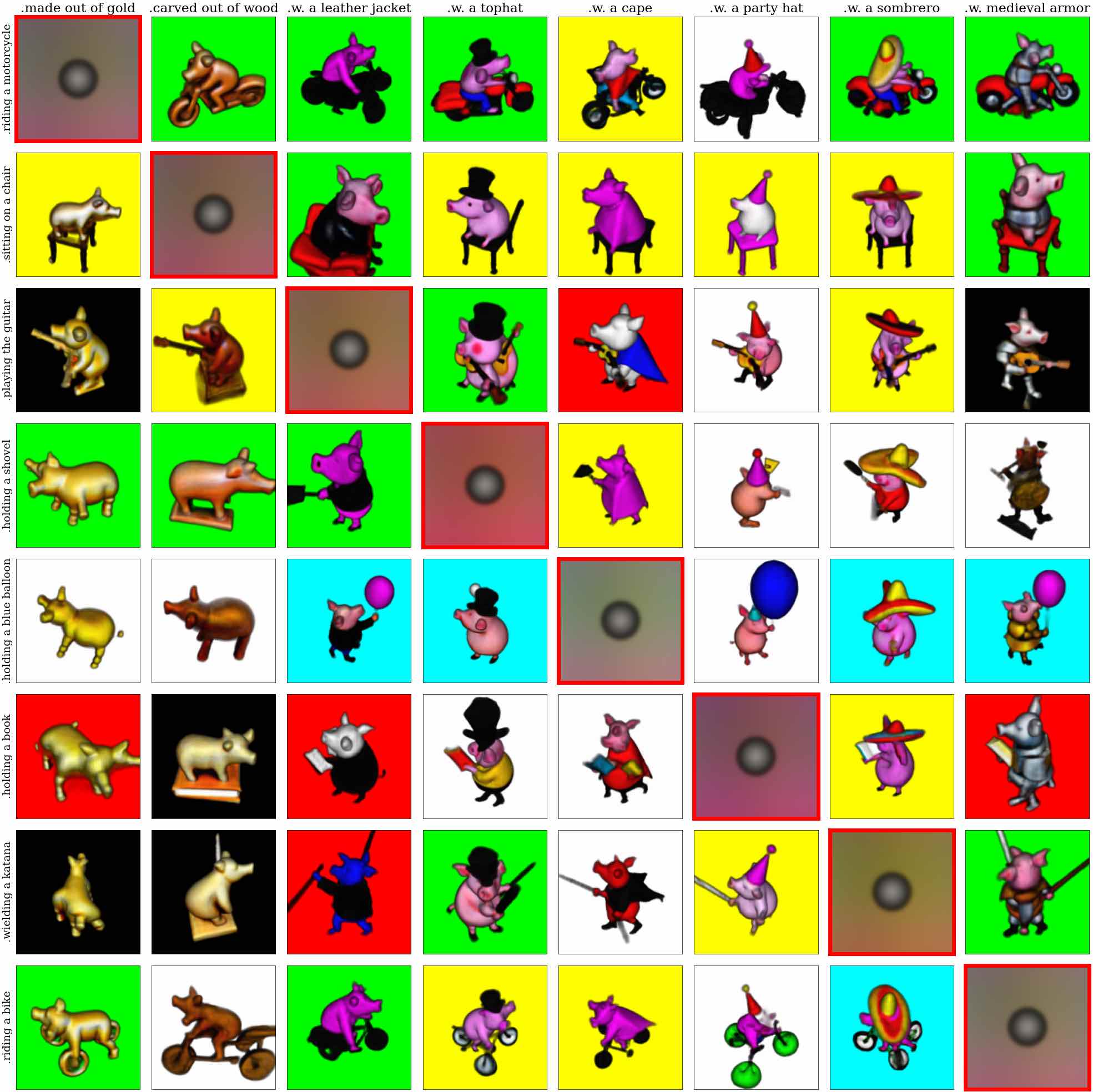}};
        %\node[left=of img1, node distance=0cm, rotate=90, xshift=2.0cm, yshift=-.5cm, font=\color{black}] {y-axis};
        \node[above=of img2, node distance=0cm, xshift=-.1cm, yshift=-1.2cm,font=\color{black}]{Per-prompt Training};
    \end{tikzpicture}
    \vspace{-0.015\textheight}
    \caption{
        We show results on a compositional prompt set.
        Each row has a different activity, while each column has a theme, which we combine into the prompt ``\emph{a pig} \{\texttt{activity}\} \{\texttt{theme}\}." while we evaluate generalization on a held-out set of unseen testing prompts in {\color{red}red} on the diagonal.
        \emph{Left:}
            Our method.
            Interestingly, the amortized objects have a unified orientation.
        \emph{Right:}
            The per-prompt training baseline~\citep{poole2022dreamfusion}, with a random initialization for unseen prompts to align compute budgets.
        \textbf{Takeaway:} 
            Our model performs comparably to per-prompt training on the seen prompts, with a far smaller compute budget (Fig.~\ref{fig:all_quantitative}).
            Importantly, we perform strongly on {\color{red}unseen prompts} with no extra training, unlike per-prompt training.
        % \Kevin{do we need to cut pig grid to rectangle?}
        % \Jon{Q: This is all qualitative? Ex., could we report high r-prec in an App. table.} 
        % \Kevin{We should have this def before deadline. I think today}
    }
    \vspace{-0.00\textheight}
    \label{fig:compositional_amortization}
\end{figure*}

\begin{figure*}[ht!]
    \vspace{-0.015\textheight}
    \centering
    \begin{tikzpicture}
        \centering
        % \node (img11){\includegraphics[trim={.0cm .0cm .0cm .0cm}, clip, width=.15\linewidth]{images/interpolation_set_1/frame0.png}};
        % \node [right=of img11, xshift=-1.5cm](img12){\includegraphics[trim={.0cm .0cm .0cm .0cm}, clip, width=.15\linewidth]{images/interpolation_set_1/frame1.png}};
        % \node [right=of img12, xshift=-1.5cm](img13){\includegraphics[trim={.0cm .0cm .0cm .0cm}, clip, width=.15\linewidth]{images/interpolation_set_1/frame2.png}};
        % \node [right=of img13, xshift=-1.5cm](img14){\includegraphics[trim={.0cm .0cm .0cm .0cm}, clip, width=.15\linewidth]{images/interpolation_set_1/frame3.png}};
        % \node [right=of img14, xshift=-1.5cm](img15){\includegraphics[trim={.0cm .0cm .0cm .0cm}, clip, width=.15\linewidth]{images/interpolation_set_1/frame4.png}};
        % \node [right=of img15, xshift=-1.5cm](img16){\includegraphics[trim={.0cm .0cm .0cm .0cm}, clip, width=.15\linewidth]{images/interpolation_set_1/frame5.png}};
        % \node [right=of img16, xshift=-1.5cm](img17){\includegraphics[trim={.0cm .0cm .0cm .0cm}, clip, width=.15\linewidth]{images/interpolation_set_1/frame6.png}};
        % \node[above=of img11, node distance=0cm, xshift=-.1cm, yshift=-1.2cm,font=\color{black}]{Hamburger};
        % \node[above=of img17, node distance=0cm, xshift=-.1cm, yshift=-1.2cm,font=\color{black}]{Pineapple};
        
        \node (img21){\includegraphics[trim={.0cm .3cm .0cm .3cm}, clip, width=.142\linewidth]{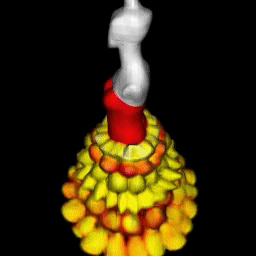}};
        \node [right=of img21, xshift=-1.26cm](img22){\includegraphics[trim={.0cm .3cm .0cm .3cm}, clip, width=.142\linewidth]{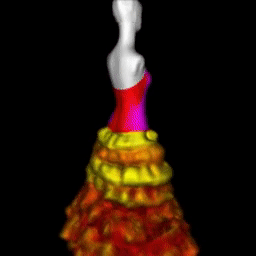}};
        \node [right=of img22, xshift=-1.26cm](img23){\includegraphics[trim={.0cm .3cm .0cm .3cm}, clip, width=.142\linewidth]{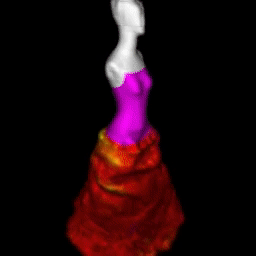}};
        \node [right=of img23, xshift=-1.26cm](img24){\includegraphics[trim={.0cm .3cm .0cm .3cm}, clip, width=.142\linewidth]{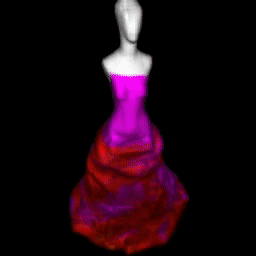}};
        \node [right=of img24, xshift=-1.26cm](img25){\includegraphics[trim={.0cm .3cm .0cm .3cm}, clip, width=.142\linewidth]{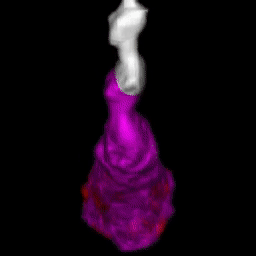}};
        \node[below=of img21, node distance=0cm, xshift=1.0cm, yshift=1.2cm,font=\color{black}]{``\emph{... dress made of fruit ...}"};
        \node[below=of img25, node distance=0cm, xshift=-1.5cm, yshift=1.2cm,font=\color{black}]{``\emph{... dress made of garbage bags...}"};
        \node[above=of img24, node distance=0cm, xshift=-.1cm, yshift=-1.2cm,font=\color{black}]{\footnotesize{Rendered frames from ATT3D with text embedding $(1 - \alpha) \textToken_1 + \alpha \textToken_2$ for $\alpha \in [0, 1]$}};

                \node [right=of img25, xshift=.55cm] (img111){\includegraphics[trim={.0cm .0cm .0cm .0cm},clip,width=.08\linewidth]{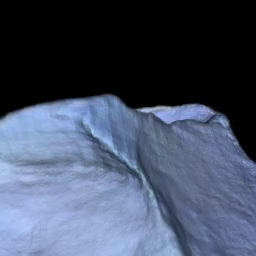}};
                \node[above=of img111, node distance=0cm, xshift=-.0cm, yshift=-1.2cm,font=\color{black}]{\footnotesize{``\emph{snowy rock}"}};
                
                \node [below=of img111, yshift=1.25cm, xshift=-.04\linewidth](img121){\includegraphics[trim={.0cm .0cm .0cm .0cm},clip,width=.08\linewidth]{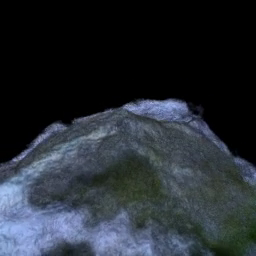}};
                \node [right=of img121, xshift=-1.25cm](img22){\includegraphics[trim={.0cm .0cm .0cm .0cm},clip,width=.08\linewidth]{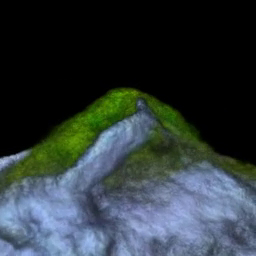}};

                \node [below=of img121, yshift=1.25cm, xshift=-.04\linewidth](img131){\includegraphics[trim={.0cm .0cm .0cm .0cm},clip,width=.08\linewidth]{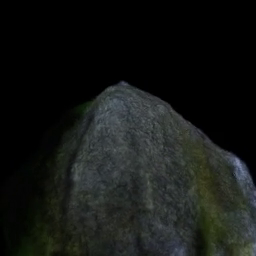}};
                \node [right=of img131, xshift=-1.25cm](img132){\includegraphics[trim={.0cm .0cm .0cm .0cm},clip,width=.08\linewidth]{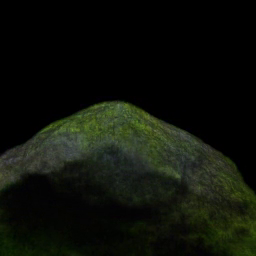}};
                \node [right=of img132, xshift=-1.25cm](img133){\includegraphics[trim={.0cm .0cm .0cm .0cm},clip,width=.08\linewidth]{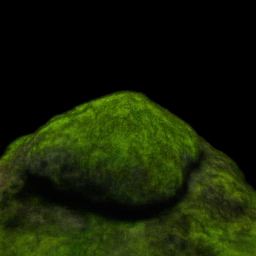}};
                \node[below=of img131, node distance=0cm, xshift=.1cm, yshift=1.2cm,font=\color{black}]{\footnotesize{``\emph{jagged rock}"}};
                \node[below=of img133, node distance=0cm, xshift=-.1cm, yshift=1.2cm,font=\color{black}]{\footnotesize{``\emph{mossy rock}"}};

        \node [below=of img21, yshift=.85cm](img31){\includegraphics[trim={.0cm .3cm .0cm .3cm}, clip, width=.142\linewidth]{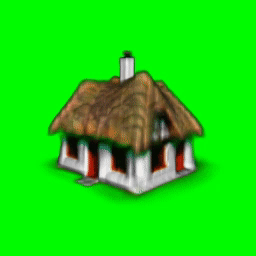}};
        \node [right=of img31, xshift=-1.26cm](img32){\includegraphics[trim={.0cm .3cm .0cm .3cm}, clip, width=.142\linewidth]{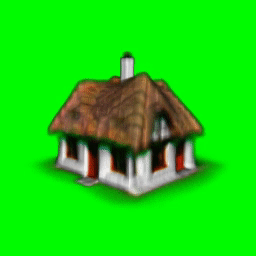}};
        \node [right=of img32, xshift=-1.26cm](img33){\includegraphics[trim={.0cm .3cm .0cm .3cm}, clip, width=.142\linewidth]{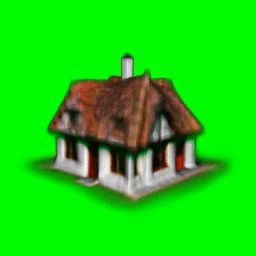}};
        \node [right=of img33, xshift=-1.26cm](img34){\includegraphics[trim={.0cm .3cm .0cm .3cm}, clip, width=.142\linewidth]{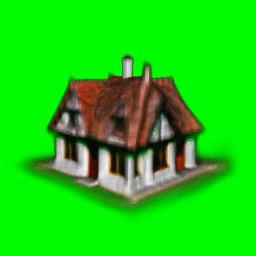}};
        \node [right=of img34, xshift=-1.26cm](img35){\includegraphics[trim={.0cm .3cm .0cm .3cm}, clip, width=.142\linewidth]{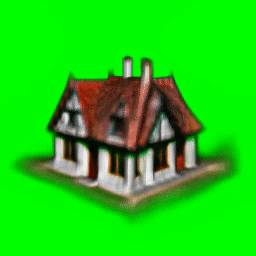}};
        \node[below=of img31, node distance=0cm, xshift=1.35cm, yshift=1.2cm,font=\color{black}]{``\emph{... cottage with a thatched roof}"};
        \node[below=of img35, node distance=0cm, xshift=-1.0cm, yshift=1.2cm,font=\color{black}]{``\emph{...  house in Tudor Style}"};

        % \node [below=of img31, yshift=.85cm](img41){\includegraphics[trim={.0cm .3cm .0cm .3cm}, clip, width=.142\linewidth]{images/interpolation_characters/frame0.png}};
        % \node [right=of img41, xshift=-1.26cm](img42){\includegraphics[trim={.0cm .3cm .0cm .3cm}, clip, width=.142\linewidth]{images/interpolation_characters/frame1.png}};
        % \node [right=of img42, xshift=-1.26cm](img43){\includegraphics[trim={.0cm .3cm .0cm .3cm}, clip, width=.142\linewidth]{images/interpolation_characters/frame2.png}};
        % \node [right=of img43, xshift=-1.26cm](img44){\includegraphics[trim={.0cm .3cm .0cm .3cm}, clip, width=.142\linewidth]{images/interpolation_characters/frame3.png}};
        % \node [right=of img44, xshift=-1.26cm](img45){\includegraphics[trim={.0cm .3cm .0cm .3cm}, clip, width=.142\linewidth]{images/interpolation_characters/frame4.png}};
        % \node [right=of img45, xshift=-1.26cm](img46){\includegraphics[trim={.0cm .3cm .0cm .3cm}, clip, width=.142\linewidth]{images/interpolation_characters/frame5.png}};
        % \node [right=of img46, xshift=-1.26cm](img47){\includegraphics[trim={.0cm .3cm .0cm .3cm}, clip, width=.142\linewidth]{images/interpolation_characters/frame6.png}};
        % \node[below=of img41, node distance=0cm, xshift=.75cm, yshift=1.2cm,font=\color{black}]{``\emph{a frog wearing a sweater}"};
        % \node[below=of img47, node distance=0cm, xshift=-1.25cm, yshift=1.2cm,font=\color{black}]{``\emph{a bear dressed as a lumberjack}"};

        \node [below=of img31, yshift=.85cm](img51){\includegraphics[trim={.0cm .3cm .0cm .3cm}, clip, width=.142\linewidth]{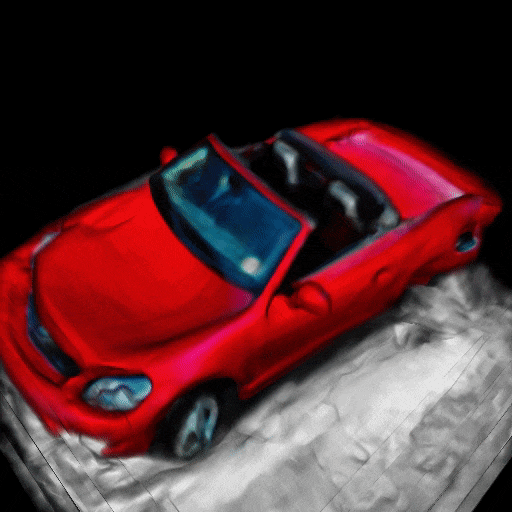}};
        \node [right=of img51, xshift=-1.26cm](img52){\includegraphics[trim={.0cm .3cm .0cm .3cm}, clip, width=.142\linewidth]{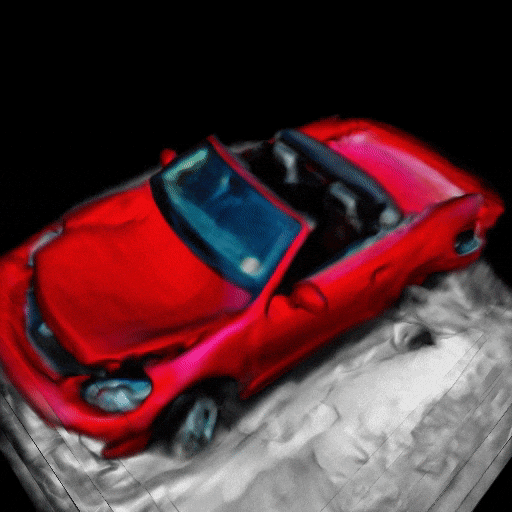}};
        \node [right=of img52, xshift=-1.26cm](img53){\includegraphics[trim={.0cm .3cm .0cm .3cm}, clip, width=.142\linewidth]{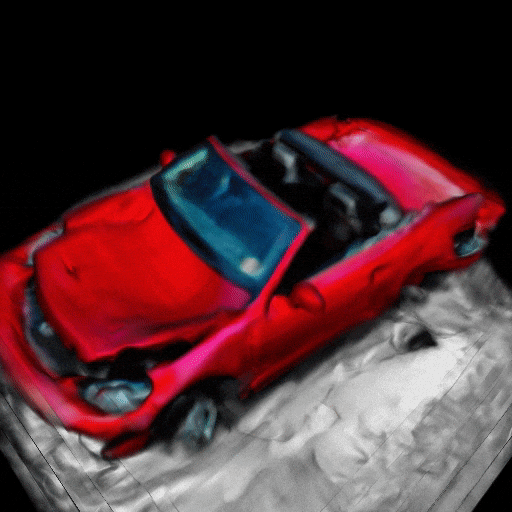}};
        \node [right=of img53, xshift=-1.26cm](img54){\includegraphics[trim={.0cm .3cm .0cm .3cm}, clip, width=.142\linewidth]{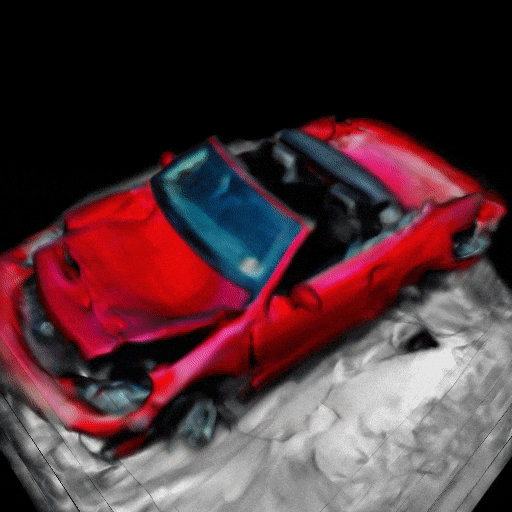}};
        \node [right=of img54, xshift=-1.26cm](img55){\includegraphics[trim={.0cm .3cm .0cm .3cm}, clip, width=.142\linewidth]{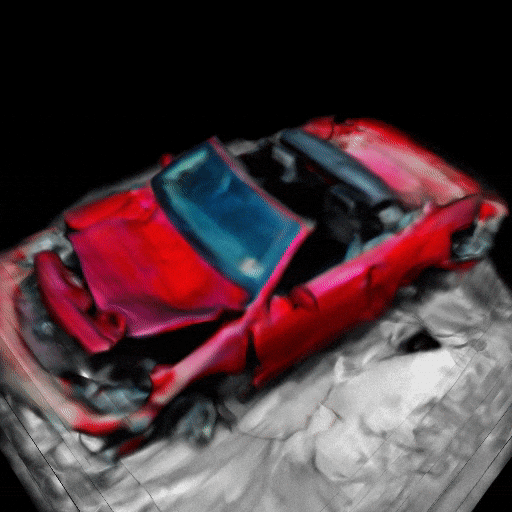}};
        \node [right=of img55, xshift=-1.26cm](img56){\includegraphics[trim={.0cm .3cm .0cm .3cm}, clip, width=.142\linewidth]{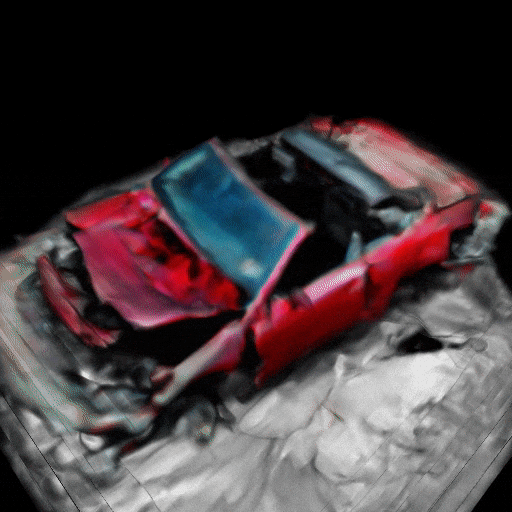}};
        \node [right=of img56, xshift=-1.26cm](img57){\includegraphics[trim={.0cm .3cm .0cm .3cm}, clip, width=.142\linewidth]{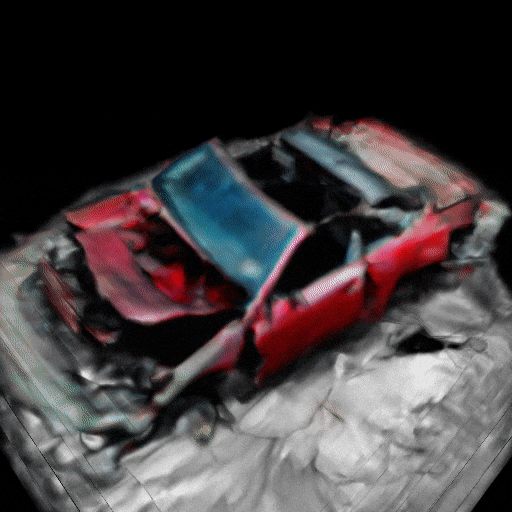}};
        \node[below=of img51, node distance=0cm, xshift=.25cm, yshift=1.2cm,font=\color{black}]{``\emph{... red convertible}"};
        \node[below=of img57, node distance=0cm, xshift=-.25cm, yshift=1.2cm,font=\color{black}]{``\emph{... destroyed car}"};

        \node [below=of img51, yshift=.85cm](img81){\includegraphics[trim={.0cm .3cm .0cm .3cm}, clip, width=.142\linewidth]{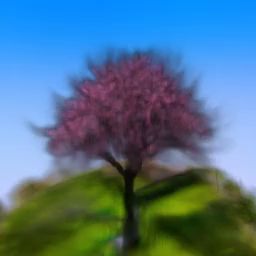}};
        \node [right=of img81, xshift=-1.26cm](img82){\includegraphics[trim={.0cm .3cm .0cm .3cm}, clip, width=.142\linewidth]{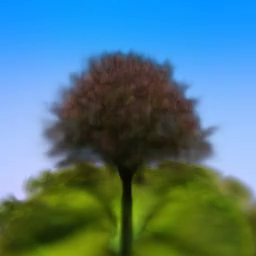}};
        \node [right=of img82, xshift=-1.26cm](img83){\includegraphics[trim={.0cm .3cm .0cm .3cm}, clip, width=.142\linewidth]{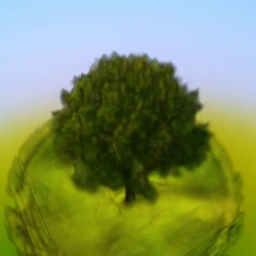}};
        \node [right=of img83, xshift=-1.26cm](img84){\includegraphics[trim={.0cm .3cm .0cm .3cm}, clip, width=.142\linewidth]{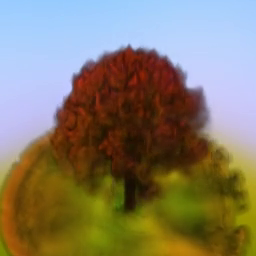}};
        \node [right=of img84, xshift=-1.26cm](img85){\includegraphics[trim={.0cm .3cm .0cm .3cm}, clip, width=.142\linewidth]{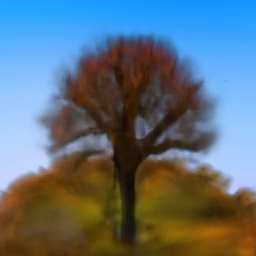}};
        \node [right=of img85, xshift=-1.26cm](img86){\includegraphics[trim={.0cm .3cm .0cm .3cm}, clip, width=.142\linewidth]{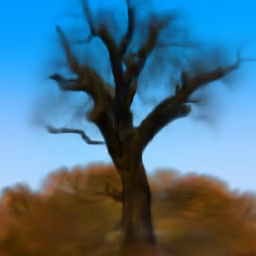}};
        \node [right=of img86, xshift=-1.26cm](img87){\includegraphics[trim={.0cm .3cm .0cm .3cm}, clip, width=.142\linewidth]{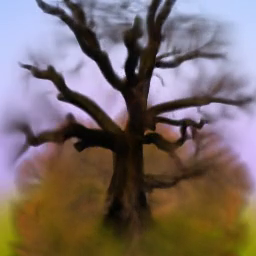}};
        \node[below=of img81, node distance=0cm, xshift=.25cm, yshift=1.2cm,font=\color{black}]{``\emph{...in the spring}"};
        \node[below=of img83, node distance=0cm, xshift=.0cm, yshift=1.2cm,font=\color{black}]{``\emph{...in the summer}"};
        \node[below=of img85, node distance=0cm, xshift=.0cm, yshift=1.2cm,font=\color{black}]{``\emph{...in the fall}"};
        \node[below=of img87, node distance=0cm, xshift=-.1cm, yshift=1.2cm,font=\color{black}]{``\emph{...in the winter}"};
    \end{tikzpicture}
    \vspace{-0.035\textheight}
    \caption{
        % We show text embedding interpolations for a model trained with our method.
        % \Kevin{unless I'm misunderstanding, with no additional training, could be misleading here. I think we should cut it? We still need to train over interpolants?}\James{Agreed}
        We show renders of our model's output on interpolated text embeddings $(1 - \alpha) \textToken_1 + \alpha \textToken_2$.
        We generate a continuum of landscape, clothing, building, and vehicle assets, and use chains of prompts for animations, like seasonality in a tree.
        % \Jon{Re-add to App + fix view?}
    }
    \vspace{-0.04\textheight}
    \label{fig:interpolation}
\end{figure*}
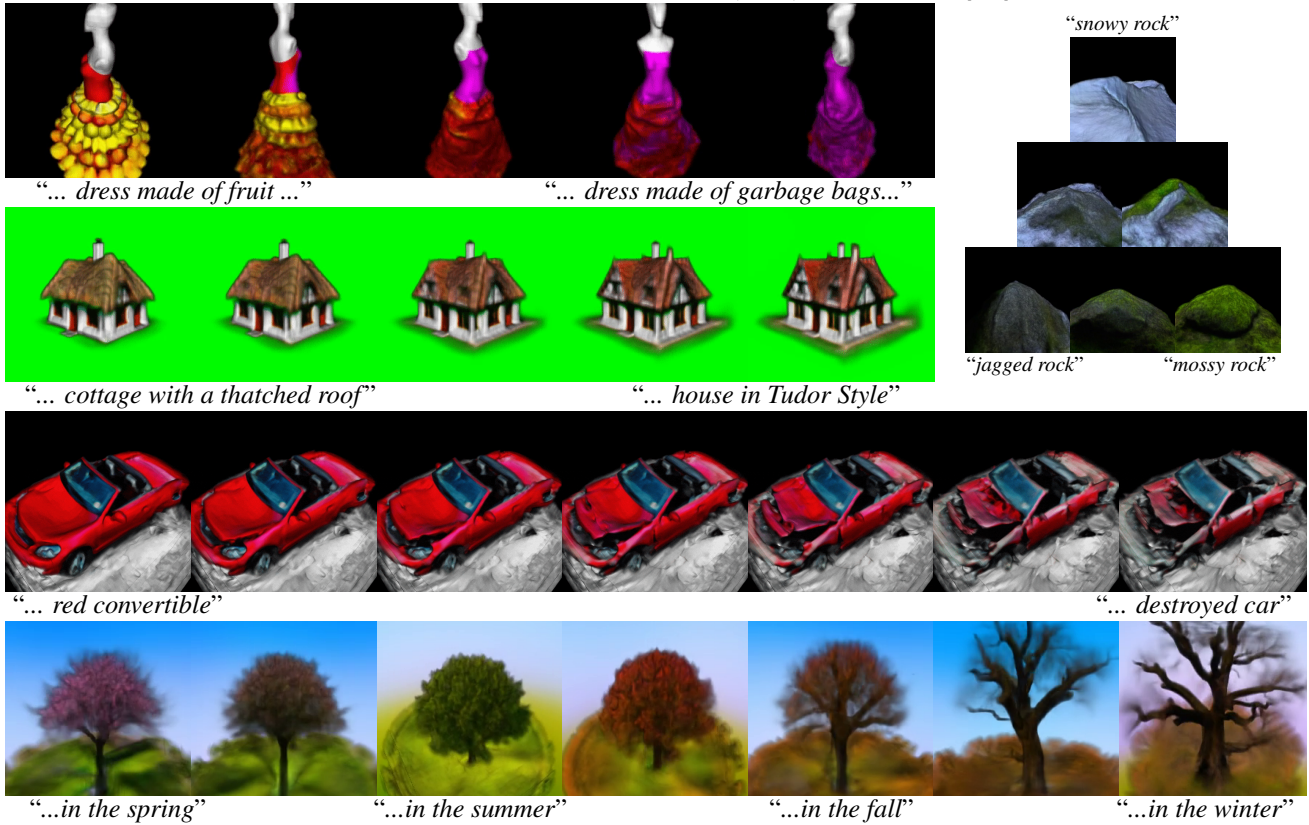

    % We find that the challenging training dynamics can cause problems for a na\"ive implementation of amortized optimization but these can be remedied by leveraging techniques from the bilevel optimization community. The result is a robust training scheme that allows us to learn a text-to-3D model trained over multiple text prompts that can generate novel 3D scenes near-instantly.
    
    \newcommand{\contributionSpaceShrink}{\vspace{-0.01\textheight}}  
    \subsection{Contributions}
        %\noindent
        We present a method to synthesize 3D objects from text prompts immediately.
        By using amortized optimization we can:
        \contributionSpaceShrink
        \begin{itemize}
            \item Generalize to new prompts -- Fig.~\ref{fig:compositional_amortization}.
            \contributionSpaceShrink
            \item Interpolate between prompts -- Fig.~\ref{fig:interpolation}.
            \contributionSpaceShrink
            \item Amortize over settings other than text prompts -- Sec.~\ref{sec:AmortizedInterpolation}.
            \contributionSpaceShrink
            \item Reduce overall training time -- Fig.~\ref{fig:all_quantitative}.
            %\item Rapidly finetune on arbitrary prompts -- Fig.~\ref{fig:finetuning_qualitative}.
            % \contributionSpaceShrink
            % \item Reduce artifacts from overfitting prompts -- Fig.~\ref{fig:janus_reduction}.
        \end{itemize}
    
    % \subsection{Contribution Impact}
    %     In recent years, text-to-X generation-as-a-service has become very popular (HuggingFace, Stability, Runway, DALLE-2, etc…) and we are seeing a rise in text-to-X generation happening at a massive scale over many diverse sets of prompts being submitted daily.
    %     This has huge implications for the service providers in terms of energy and compute usage, as the total-cost-of-ownership for standard single prompt text-to-X generation systems scales linearly by the number of prompts.
    %     This is also an environmental concern… [cite studies of rainforests being burned down by GPUs].
    %     In our work, we show that we are able to significantly reduce costs over batches of prompts, showing sublinear scaling of text-to-X generation over prompts.
    %\input{sections/teaser_figure.tex}
    \section{Background}\label{sec:background}
        %Put notation, etc. in this Sec.
        This section contains concepts and prior work relevant to our method, with notation in App. Table~\ref{tab:TableOfNotation}.
        
        \subsection{NeRFs for Image-to-3D}
            %\noindent
            NeRFs~\citep{mildenhall2021nerf} represent 3D scenes via a radiance field parameterized by a neural network.
            We denote 3D coordinates with $\position \! = \! [x, y, z] \in \positionDom$ and the radiance values with $\density \! =\!  [\sigma, r, g, b] \in \densityDom$.
            NeRFs are trained to output radiance fields to render frames similar to multi-view images with camera information.
            Simple NeRFs map locations $\position$ to radiances $\density$ via an MLP-parameterized function.
            Recent NeRFs use spatial grids storing parameters queried per location \citep{muller2022instant, takikawa2022variable, Chan_2022_CVPR}, integrating spatial inductive biases.
            We view this as a \emph{point-encoder} function $\pointEncoder_{\pointEncoderParams} \!\! : \!\positionDom \! \to \!  \encoderRange$ with parameters $\pointEncoderParams$ encoding a location $\position$ before the final MLP $\nerfNet \! : \! \encoderRange \! \to \! \densityDom$.
            \begin{equation}\label{eq:point_encoder}
                \density = \nerfNet\left(\pointEncoder_{\pointEncoderParams}\left(\position\right)\right)
            \end{equation}
        
        \subsection{Text-to-Image Generation}\label{sec:text-to-image}
            % Describe our DDM.  Other text-to-image (GAN, whatever) go in related work.
            The wide availability of captioned image datasets has enabled the development of powerful text-to-image generative models.
            We use a DDM with comparable architecture to recent large-scale methods \citep{balaji2022ediffi, rombach2022high, saharia2022photorealistic}.
            We train for score-matching, where (roughly) input images have noise added to them \citep{ho2020denoising, song2020score} that the DDM predicts.
            Critically, these models can be conditioned on text to generate matching images via classifier-free guidance\cite{ho2022classifier}.
            We use pre-trained T5-XXL~\citep{JMLR:v21:20-074} and CLIP~\citep{pmlr-v139-radford21a} encoders to generate text embeddings, which the DDM conditions on via cross-attention with latent image features.
            Crucially, we reuse the text token embeddings -- denoted $\textToken$ -- for modulating our NeRF.
            % \Kevin{can't we refer to same pipeline as magic3d directly in this part now?}\James{We can, but we should still include full details in supplementary at least. In case magic3D paper is updated with some changes to the pipeline in future.}
        
        \subsection{Text-to-3D (TT3D) Generation}
\begin{figure*}%[h]
    \vspace{-0.03\textheight}
    \centering\includegraphics{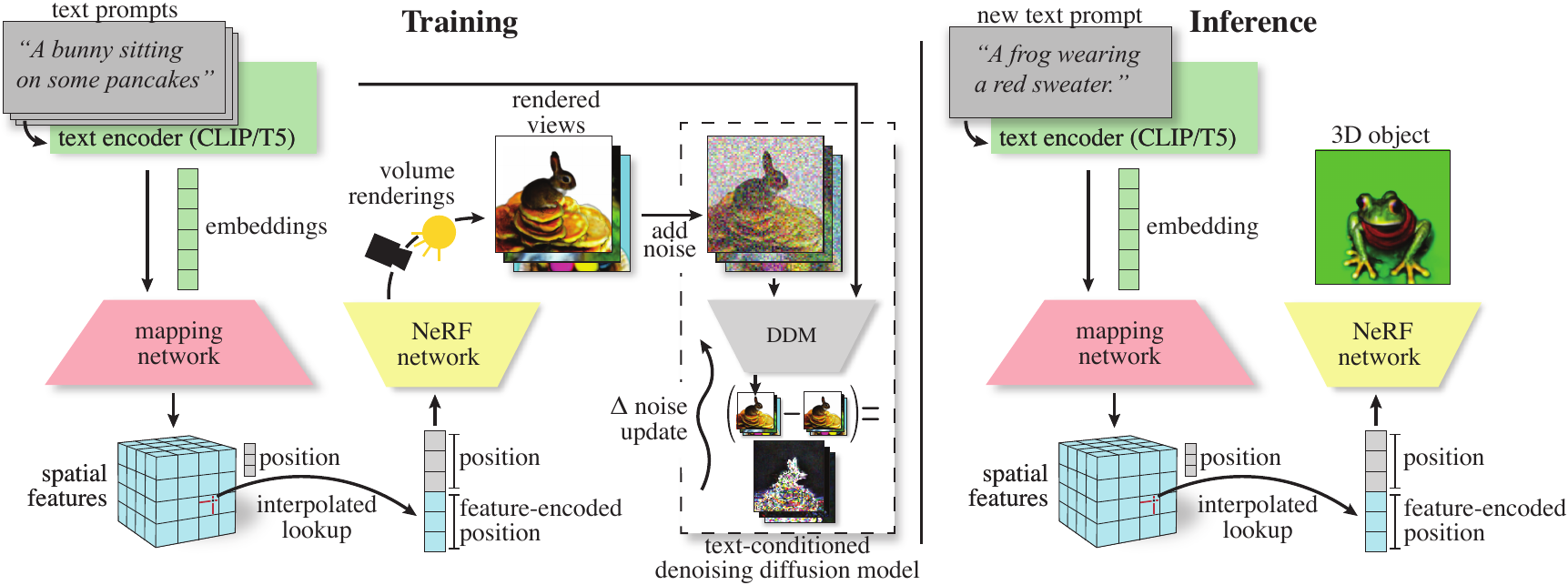}
    \vspace{-0.025\textheight} % 0.02
    % \vspace{-0.0\textheight}
    \caption{
        We show a schematic of our text-to-3D pipeline with changes from DreamFusion's pipeline~\citep{poole2022dreamfusion} shown in {\color{red}red} and pseudocode in Alg.~\ref{alg:amortized_training}.
        The text encoder (in {\color{green}green}) provides its -- potentially cached -- text embedding ${\color{green}\textToken}$ to the text-to-image DDM and now also to the mapping network ${\color{red}\mappingNet}$ (in {\color{red}red}).
        We use a spatial point-encoder ${\color{blue}\pointEncoder}_{{\color{red}\mappingNet(}{\color{green}\textToken}{\color{red})}}$ (in {\color{blue}blue}) for our position $\position$, whose parameters are modulations from the mapping network ${\color{red}\mappingNet(}{\color{green}\textToken}{\color{red})}$.
        The final NeRF MLP $\nerfNet$ outputs a radiance $\density$ given the point encoding: $\density = \nerfNet({\color{blue}\pointEncoder}_{{\color{red}\mappingNet(}{\color{green}\textToken}{\color{red})}}(\position))$, which we render into views.
        \emph{Left:}
            At training time, the rendered views are input to the DDM to provide a training update.
            The NeRF network $\nerfNet$, mapping network ${\color{red}\mappingNet}$, and (effectively) the spatial point encoding ${\color{blue}\pointEncoder}_{{\color{red}\mappingNet(}{\color{green}\textToken}{\color{red})}}$ are optimized.
        \emph{Right}:
            At inference time, we use the pipeline up to the NeRF for representing the 3D object.
            %Note that the spatial features need only be computed once per prompt.
    }
    \vspace{-0.015\textheight}
    \label{fig:amortized-text-to-3d-pipeline}
\end{figure*}

            % Dreamfusion, etc. Other text-to-3d-methods with less quality go in related work.
            %\noindent
            Prior works rely on per-prompt optimization to generate 3D scenes. Recent TT3D methods \cite{poole2022dreamfusion, wang2022score} use text-to-image generative models to train NeRFs. To do so,
            they render a view and add noise.
            The DDM, conditioned on a text prompt, approximates $\noise$ with $\hat{\noise}$, using the difference $\hat{\noise} - \noise$ to update NeRF parameters.
            We outline this method in Alg.~\ref{alg:amortized_training} and Fig.~\ref{fig:amortized-text-to-3d-pipeline} and refer to DreamFusion Sec. 3 for more details. 
        
        \subsection{Amortized Optimization}
            % Talk about modulations approaches we look at. Other modulation less relevant in related work.
            Amortized optimization methods use learning to predict solutions when we repeatedly solve similar instances of the same problem~\citep{amos2022tutorial}.
            Current TT3D independently optimizes prompts, whereas, in Sec.~\ref{sec:method}, we use amortized methods.
            
            A typical amortization strategy is to find a problem context -- denoted $\textEmbedding$ -- to change our optimization, with some strategies specialized for NeRFs~\citep{rebain2022attention}.
            For example, concatenating the context to the NeRF's MLP:
                $\density(\position, \textEmbedding) = \nerfNet(\pointEncoder(\position), \textEmbedding)$
            Or, having a \emph{mapping network} $\mappingNet$ outputting modulations to the weights or hidden units:
            \\[-0.01\textheight]
            \begin{equation}\label{eq:mapping_net}
                \density\left(\position, \textEmbedding\right) = \nerfNet\left(\smash{\pointEncoder_{\mappingNet\left(\textEmbedding\right)}}\left(\position\right)\right)
            \end{equation}
            \\[-0.02\textheight]
            %Sentence to lead into the next section.... our method, where we use the text-token for the modulation.
            But, designing useful contexts, $\textEmbedding$, can be non-trivial.
            %\Kevin{maybe designing instead of finding, also some might disagree since we just reuse the text encoding}

\begin{figure}%[H]
    \vspace{0.03\textheight}
    \centering
    \begin{tikzpicture}
        \centering
        \node (img11){\includegraphics[trim={.0cm .0cm .0cm .0cm},clip,width=.27\linewidth]{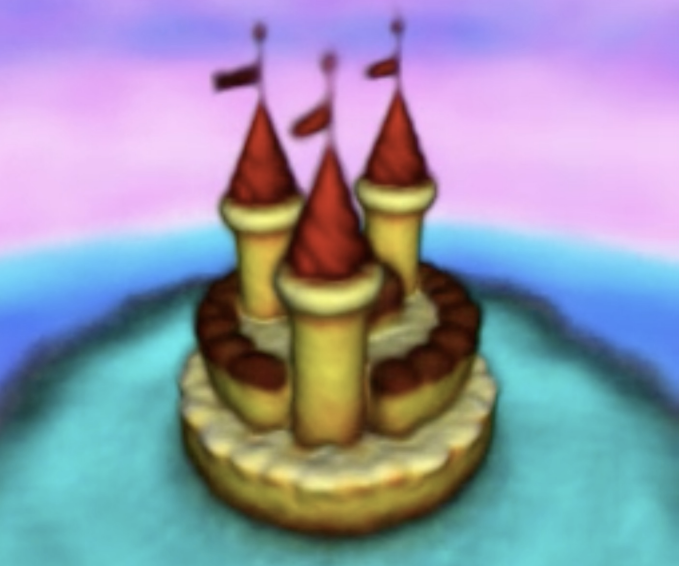}};
        \node [right=of img11, xshift=-1cm](img12){\includegraphics[trim={.0cm .0cm .0cm .0cm},clip,width=.27\linewidth]{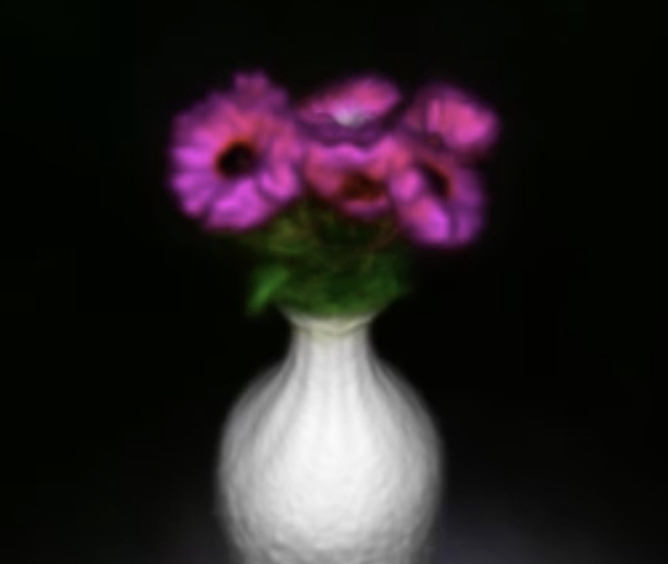}};
        \node [right=of img12, xshift=-1cm](img13){\includegraphics[trim={.0cm .0cm .0cm .0cm},clip,width=.27\linewidth]{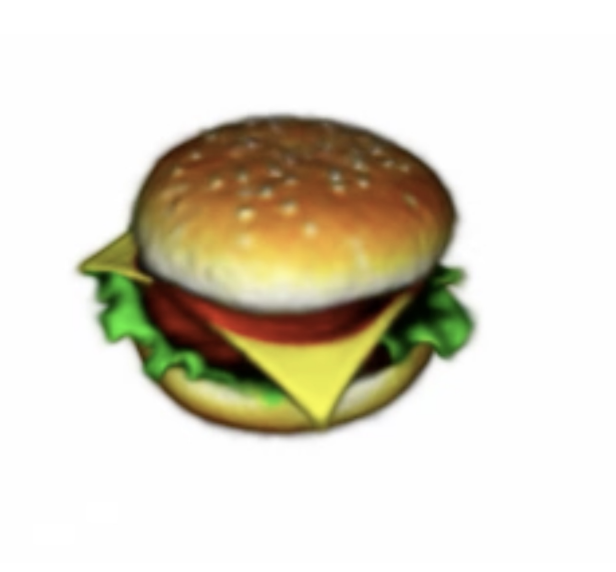}};
        \node[left=of img11, node distance=0cm, rotate=90, xshift=1.0cm, yshift=-.75cm, font=\color{black}] {{\footnotesize{DreamFusion}}};
        
        \node [below=of img11, yshift=1cm](img21){\includegraphics[trim={.0cm .0cm .0cm .0cm},clip,width=.27\linewidth]{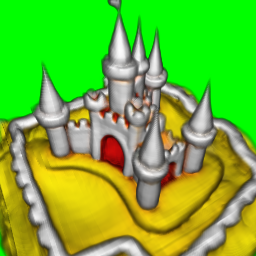}};
        \node [right=of img21, xshift=-1cm](img22){\includegraphics[trim={.0cm .0cm .0cm .0cm},clip,width=.27\linewidth]{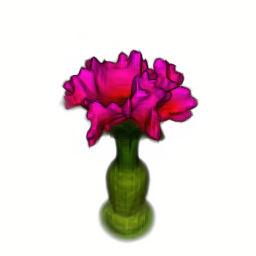}};
        \node [right=of img22, xshift=-1cm](img23){\includegraphics[trim={.0cm .0cm .0cm .0cm},clip,width=.27\linewidth]{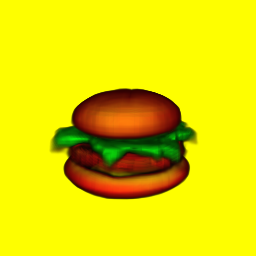}};
        \node[left=of img21, node distance=0cm, rotate=90, xshift=1.15cm, yshift=-.75cm, font=\color{black}](row1label) {{\scriptsize{DreamFusion reimpl.}}};
        
        \node[left=of row1label, node distance=0cm, rotate=90, xshift=3.5cm, yshift=-.65cm, font=\color{black}] {{\footnotesize{Per-prompt Training}}};
        % \node[left=of img21, node distance=0cm, rotate=90, xshift=1.5cm, yshift=-.75cm, font=\color{black}, label={[align=center, rotate=90]{\footnotesize{Dreamfusion reimpl.}}\\{\footnotesize{=Single-prompt train}}}] {};
        
        \node [below=of img21, yshift=1cm](img31){\includegraphics[trim={.0cm .0cm .0cm .0cm},clip,width=.27\linewidth]{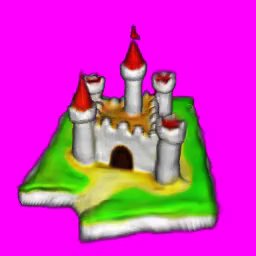}};
        \node [right=of img31, xshift=-1cm](img32){\includegraphics[trim={.0cm .0cm .0cm .0cm},clip,width=.27\linewidth]{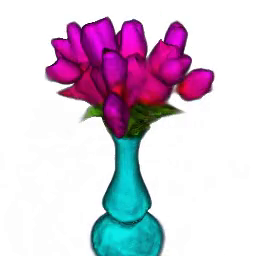}};
        \node [right=of img32, xshift=-1cm](img33){\includegraphics[trim={.0cm .0cm .0cm .0cm},clip,width=.27\linewidth]{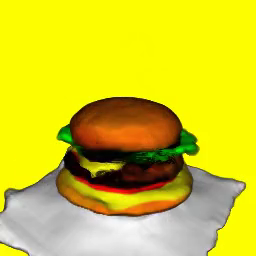}};
        %\node[left=of img31, node distance=0cm, rotate=90, xshift=1.0cm, yshift=-.75cm, font=\color{black}, label={[align=center, rotate=90]{\footnotesize{Amortized Train}}\\{\footnotesize{=Our Method}}}] {};
        \node[left=of img31, node distance=0cm, rotate=90, xshift=1.25cm, yshift=-.75cm, font=\color{black}](row2label) {{\scriptsize{= Our Method, ATT3D}}};
        
        % \node [below=of img31, yshift=1cm](img41){\includegraphics[trim={.0cm .0cm .0cm .0cm},clip,width=.27\linewidth]{example-image-a}};
        % \node [right=of img41, xshift=-1cm](img42){\includegraphics[trim={.0cm .0cm .0cm .0cm},clip,width=.27\linewidth]{example-image-a}};
        % \node [right=of img42, xshift=-1cm](img43){\includegraphics[trim={.0cm .0cm .0cm .0cm},clip,width=.27\linewidth]{example-image-a}};
        % \node[left=of img41, node distance=0cm, rotate=90, xshift=.75cm, yshift=-.75cm, font=\color{black}]{{\footnotesize{+ Finetune}}};
        
        \node[left=of row2label, node distance=0cm, rotate=90, xshift=2.455cm, yshift=-.65cm, font=\color{black}]{{\scriptsize{Amortized Training}}};
        %\node[left=of img41, node distance=0cm, rotate=90, xshift=1.0cm, yshift=-.75cm, font=\color{black}, label={[align=center, rotate=90]{\footnotesize{Amortized Train + }}\\{\footnotesize{Finetune}}}] {};
        
        \node[below=of img31, node distance=0cm, xshift=-.1cm, yshift=.05cm,font=\color{black}, label={[align=center]{\tiny{matte painting of a}}\\[-0.01\textheight]{\tiny{castle made of cheesecake}}\\[-0.01\textheight]{\tiny{ surrounded by a moat}}\\[-0.01\textheight]{\tiny{made of ice cream}}}]{};
        \node[below=of img32, node distance=0cm, xshift=-.1cm, yshift=.1cm,font=\color{black}, label={[align=center]{\footnotesize{a vase with}}\\[-0.005\textheight]{\footnotesize{pink flowers}}}]{};
        \node[below=of img33, node distance=0cm, xshift=-.1cm, yshift=.3cm,font=\color{black}, label={[align=center]{\footnotesize{a hamburger}}}]{};
    \end{tikzpicture}
    \vspace{-0.025\textheight}
    \caption{
        Here we qualitatively assess our method relative to the baseline per-prompt training -- i.e., DreamFusion's method.
        A public DreamFusion implementation is not available.
        \textbf{Takeaway:}
            Our re-implementation achieves similar quality to the original.
            Also, our amortized method performs comparably to per-prompt training.
        %\TODO{Reference this in text?}
    }
    \vspace{-0.0\textheight}
    \label{fig:usVsDF}
\end{figure}
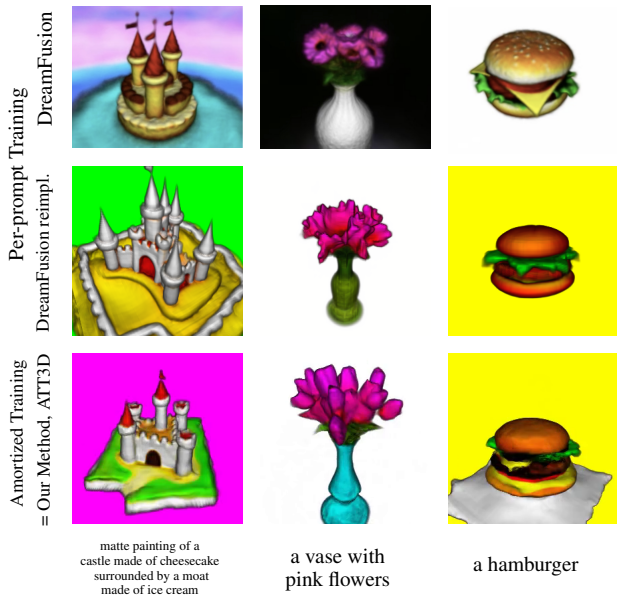
    
\section{Our Method: Amortized Text-to-3D}\label{sec:method}
    %Use notation from problem setting section
    %\noindent
    Our method has an initial training stage using amortized optimization, after which we perform cheap inference on new prompts.
    % Our method adds an additional training stage to the existing text-to-3D pipeline via amortized optimization techniques.
    We first describe the ATT3D architecture and its use during inference, then the training procedure.

    \subsection{The Amortized Model used at Inference}
        At inference, our model consists of a \emph{mapping network} $\mappingNet$, a NeRF $\nerfNet$, and a spatial grid of features $\pointEncoder_{\pointEncoderParams}$ with parameters $\pointEncoderParams$ (Fig.~\ref{fig:amortized-text-to-3d-pipeline}).
        The mapping network takes in an (encoded) text prompt $\textToken$ and produces feature grid \emph{modulations}: $\pointEncoder_{\mappingNet(\textToken)}$. 
        Our final NeRF module $\nerfNet$ is a small MLP acting on encoded points $\pointEncoder_{\mappingNet(\textToken)}(\position)$ -- Eq.~\ref{eq:point_encoder} -- representing a 3D object for the text prompt with the modulated feature grid.
        Full details are in App. Sec.~\ref{sec:app-implementation-details} and summarized here.

        \noindent
        \textbf{Architectural details:}
            We followed Instant NGP~\citep{muller2022instant} for our NeRF, notably using multi-resolution voxel/hash grids for our point-encoder $\pointEncoder$.
            We use hypernetwork modulations for implementation and computational simplicity, with alternatives of concatenation and attention considered in App.~\ref{sec:app_mapping_network}.
            % We considered concatenation, hypernetwork, and attention approaches for modulation~\citep{rebain2022attention}.
            % Concatenation was prohibitively expensive by scaling the final per-point NeRF MLP $\nerfNet$ cost.
            % Attention offered unnecessary complexity.
            %We do not use the simplest strategy of directly making $\textToken$ an input to our NeRF -- i.e, $\nerfNet(\pointEncoder(\position), \textToken)$ -- to maintain fast rendering times for training and inference.
            % 
            Hypernetwork approaches output the point-encoder parameters $\pointEncoderParams$ from a text embedding $\textToken$:
            \begin{equation}
                \pointEncoderParams = \textnormal{Hypernetwork}(\textToken)
            \end{equation}
            We simply output via a vector $\vectorEmbedding$ from the text embeddings, which is used to output the parameters via linear maps.
            % \\[-0.03\textheight]
            \begin{equation}
                \vectorEmbedding = \textnormal{SiLU}(\smash{\textnormal{linear}^{\textnormal{spec.norm}}_{\textnormal{w/ bias}}}(\textnormal{flatten}(\textToken)))
            \end{equation}
            % \\[-0.03\textheight]
            \begin{equation}
                \pointEncoderParams = \textnormal{reshape}(\smash{\textnormal{linear}^{\textnormal{spec.norm}}_{\textnormal{no bias}}}(\vectorEmbedding))
            \end{equation}
            % \\[-0.02\textheight]
            % We saw no significant benefit for using attention modulations (or even deeper networks) over this extremely simple hypernetwork approach, which both effectively solved our prompt sets.
            This $\pointEncoderParams$ parameterizes the point-encoder $\pointEncoder_{\pointEncoderParams}$, which is used to evaluate radiances per-point as per Eq.~\ref{eq:point_encoder}.
            This simple approach solved our prompt sets, so we used it in all results.
            Using more sophisticated hypernetworks performed comparably but was slower.
            However, this may be necessary for scaling to more complicated sets of prompts.
            
            Designing larger prompt sets was challenging because the per-prompt baselines could not effectively handle open-domain text prompts.
            We partially overcame this limitation by creating compositional prompt sets using prompt components that the underlying model effectively handled.

            % \Jon{
            %     Constructing more complicated prompts sets is a non-trivial problem because the per-prompt baselines we amortize can handle arbitrary open-domain text-prompts.
            %     We could search for more prompts on which they work, or improve the underlying model's robustness to prompts.
            %     Instead, we construct compositional sets out of properties that work.
            % }
            % The per-prompt baselines we amortize can not effectively handle arbitrary open-domain text prompts, which makes finding large prompt sets -- which may require sophisticated architectures -- to amortize non-trivial.

    % \vspace{-0.005\textheight}
    \subsection{Amortized Text-to-3D Training}\label{sec:amortization-addition-to-base}
    % \vspace{-0.005\textheight}
        %Topic: Describe the idea optimizing a network to optimize multiple text-embeddings simultaneously, by making them as input.
        % We use a spectral-normalized mapping (hyper)network $\mappingNet$ (Eq.~\ref{eq:mapping_net}/\ref{eq:our_mapping_net}) to modulate the point-encoder's spatial parameters $\pointEncoderParams$ depending on the text embeddings $\textToken$.
        % \begin{equation}\label{eq:our_mapping_net}
        % \density(\position, \textEmbedding) = \nerfNet(\pointEncoder_{\mappingNet({\color{red}\textToken})}(\position))
        % \end{equation}
        Alg.~\ref{alg:amortized_training} overviews our training procedure. 
        In each optimization step, we sample several prompts and produce their -- potentially cached -- text embeddings $\textEmbedding$, which we use to compute the modulations $\mappingNet(\textToken)$.
        We also sample camera poses and rendering conditions.
        These are combined with the NeRF module to render our images.
        We then use the Score Distillation Sampling loss~\citep{poole2022dreamfusion} to update the NeRF.
        
        As in prior work, we augment text prompts depending on camera position -- \emph{``\dots, front/side/rear view"}.
        We provide the text embeddings (without augmentation) to the mapping network to modulate the NeRF.

        \vspace{-0.01\textheight}
        \subsubsection{Stabilizing Optimization}\label{sec:stability_tricks}
        % \vspace{-0.005\textheight}
            The NeRF's loss is specified by a denoising diffusion model (DDM) and thus changes during training akin to bilevel setups like GANs~\citep{goodfellow2020generative, miyato2018spectral, brock2018large} and actor-critic models~\citep{pfau2016connecting}.
            We use techniques from nested optimization to stabilize training motivated by observing similar failure modes.
            Specifically, we required spectral normalization~\citep{miyato2018spectral} -- crucial for large-scale GANs~\citep{brock2018large} -- to mitigate numerical instability.
            
            Removing optimization momentum helped minimize oscillations from complex dynamics as in nested optimization~\citep{gidel2019negative, lorraine2022complex}.
            Unlike DreamFusion, we did not benefit from Distributed Shampoo~\citep{anil2020scalable} and, instead, use Adam~\citep{kingma2014adam}.

        \subsubsection{Amortizing Over Other Settings}\label{sec:AmortizedInterpolation}
            So far, we described amortizing optimization over many prompts. 
            More generally, we can amortize over other variables like the choice of guidance weight, regularizers, data augmentation, or other aspects of the loss function.
            We use this to explore techniques for allowing semantically meaningful prompt interpolations, which is a valuable property of generative models like GANs~\citep{goodfellow2020generative} and VAEs~\citep{kingma2013auto}.
            
            There are various prompt interpolation strategies we can amortize over, like, between text embeddings, guidance weights, or loss functions; see App. Fig.~\ref{fig:interpolation_hamburgers} for specifics.
            To sample an interpolated setup, we sample prompt (embedding) pairs $\textToken_1, \textToken_2$ and an interpolant weight $\alpha \in [0, 1]$.
            %\James{maybe clearer to just say: ``we must sample prompt pairs and an interpolation weight during training''}
            We must give this information to our mapping network - ex., by making it an input $\mappingNet(\textToken_1, \textToken_2, \alpha)$.
            Instead, we input interpolated embeddings, allowing an unmodified architecture and incorporating prompt permutation invariance:\footnote{By invariance we actually mean $\mappingNet(\!\textToken_1 \!,\! \textToken_2, \!\alpha\!) = \mappingNet(\!\textToken_2 ,\! \textToken_1 \!, \!1 - \alpha\!)$.}
            \begin{equation}
                \mappingNet\left(\left(1 - \alpha\right) \textToken_1 + \alpha \textToken_2\right)
            \end{equation}
            
            In addition to the text prompts distribution, we must choose the interpolant weights $\alpha$'s distribution.
            For example, we could sample uniform $\alpha \in [0, 1]$, or a binary $\alpha \in \{0, 1\}$ -- i.e., training without interpolants -- which are both special cases of a Dirichlet distribution.
            The Dirichlet concentration coefficient is another user choice to change results qualitatively -- see App. Fig.~\ref{fig:interpolation_ships}.
            We show examples of various loss interpolations in Figs.~\ref{fig:interpolation} and \ref{fig:interpolation_other}.
            The interpolation setup is further details in App. Sec.~\ref{sec:app-interpolation-exp}.

    \subsection{Why We Amortize}\label{sec:why-we-do-this}
        %Why we do this
        %We receive various benefits via amortization:
        
\begin{figure*}%[H]
    \vspace{-0.01\textheight}
    \centering
    \begin{tikzpicture}
        \centering
        \node (img11){\includegraphics[trim={.0cm .75cm .0cm .75cm},clip,width=.33\linewidth]{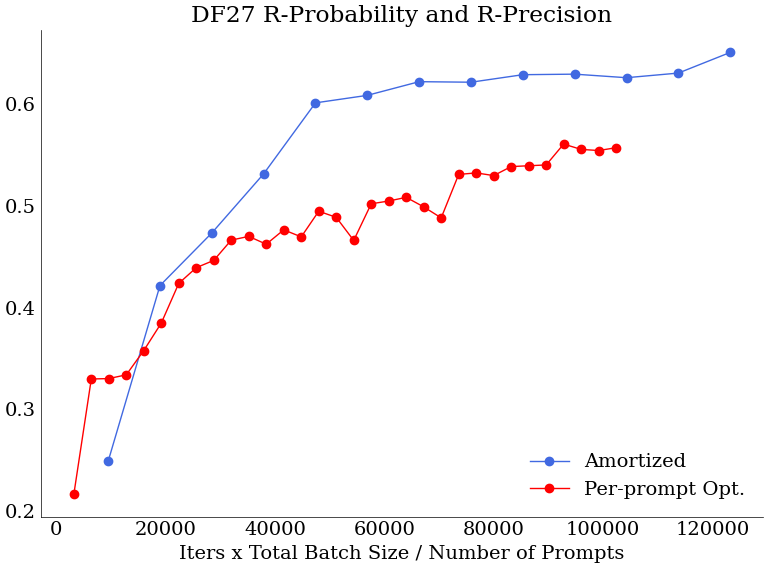}};
        % WEBPAGE CAPTION 1
        % \node[left=of img11, node distance=0cm, rotate=90, xshift=.6cm, yshift=-.9cm, font=\color{black}] {Quality};

        % WEBPAGE CAPTION 2
        % \node[left=of img11, node distance=0cm, rotate=90, xshift=.6cm, yshift=-.55cm, font=\color{black}] {Quality};
        % \node[left=of img11, node distance=0cm, rotate=90, xshift=1.75cm, yshift=-.9cm, font=\color{black}] {= Average R-probability};

        % ORIGINAL CAPTION
        \node[left=of img11, node distance=0cm, rotate=90, xshift=1.75cm, yshift=-.9cm, font=\color{black}] {Average R-probability};
        
        \node[above=of img11, node distance=0cm, xshift=-.1cm, yshift=-1.25cm,font=\color{black}]{\scriptsize{DF27 Prompts (Small)}};

        \node [right=of img11, xshift=-1.2cm](img12){\includegraphics[trim={.1cm .75cm .21cm 1.0cm},clip,width=.33\linewidth]{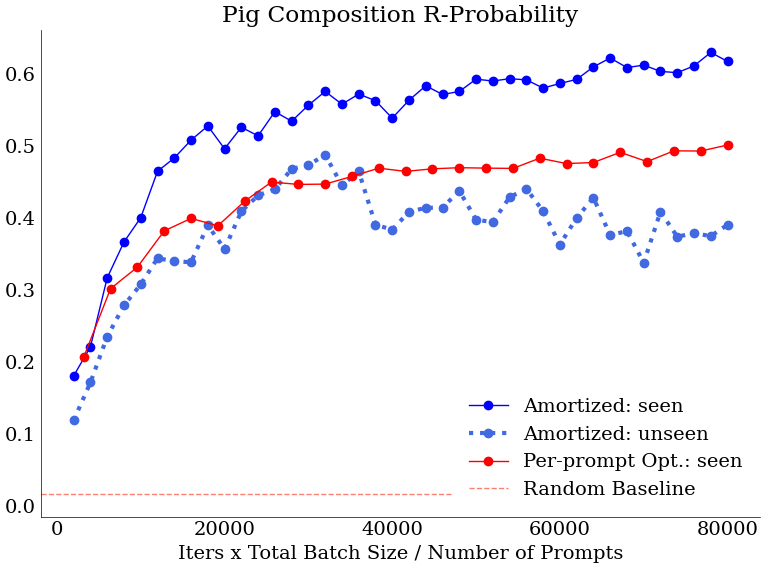}};
        \node[above=of img12, node distance=0cm, xshift=.0cm, yshift=-1.25cm,font=\color{black}]{\scriptsize{Pig Prompts (Small + Compositional)}};

        % WEBPAGE CAPTION 1
        % \node[below=of img11, node distance=0cm, xshift=.2cm, yshift=1.2cm,font=\color{black}]{Compute Budget};

        % ORIGINAL CAPTION
        \node[below=of img12, node distance=0cm, xshift=-.1cm, yshift=1.0cm,font=\color{black}]{Compute Budget = Number of rendered frames used in training per prompt};
        
        \node [right=of img12, xshift=-1.2cm](img13){\includegraphics[trim={.15cm .8cm .13cm .97cm},clip,width=.30\linewidth]{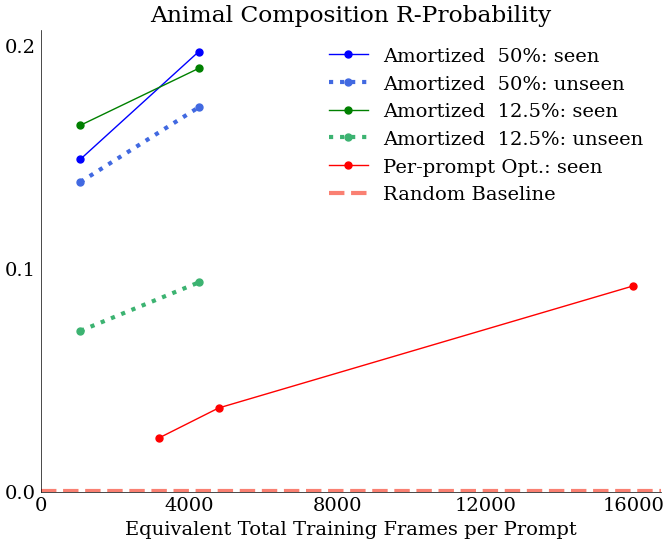}};
        \node[above=of img13, node distance=0cm, xshift=.0cm, yshift=-1.25cm,font=\color{black}]{\scriptsize{Animal Prompts (Large + Compositional)}};
    \end{tikzpicture}
    \vspace{-0.025\textheight}
    %\vspace{-0.0325\textheight}
    %
    %t{\color{gray}e}s{\color{gray}t}i{\color{gray}n}g
    \caption{
        We display the quality against compute budget for a split of \textbf{seen} \& unseen (dashed) prompts with our method (in {\color{blue}blue} and {\color{green}green}) \& existing work's {\color{red}per-prompt optimization} baseline (in {\color{red}red}).
        Our method is only trained on the seen split of the prompts.
        At a given training iteration, the amortized model is evaluated zero-shot on unseen prompts.
        \textbf{Takeaway:}
            For any compute budget, we achieve a higher quality on both the seen and unseen prompts.
            Our benefits grow for larger, compositional prompt sets.
        %Results with R-precision are similar (App. Fig.~\ref{fig:r-prec-vs-r-prob}).
        \emph{Left}:
            The $27$ prompts from DreamFusion (Fig.~\ref{fig:memorize-df27}).
        \emph{Middle}:
            The $\num{64}$ compositional pig prompts (Fig.~\ref{fig:compositional_amortization}).
            {\color{red}Per-prompt optimization} cannot perform zero-shot generation for unseen prompts, so we report the performance of a random initialization baseline.
            %so we report a random initialization comparable to a model trained on a random prompt to align compute budgets.
            %Fig.~\ref{fig:finetuning_qualitative} extends to finetuning on individual prompts at test time.
        \emph{Right}:
            The $\num{2400}$ compositional animal prompts (Fig.~\ref{fig:animals-compositional}), with varying prompt proportions used in training.
            The generalization gap is small when training on ${\color{blue}50\%}$ of the prompts.
            Notably, the cheap testing performance is better than the expensive {\color{red}per-prompt} method with only ${\color{green}12.5\%}$ of the prompts.
    }
    \vspace{-0.01\textheight}
    \label{fig:all_quantitative}
\end{figure*}
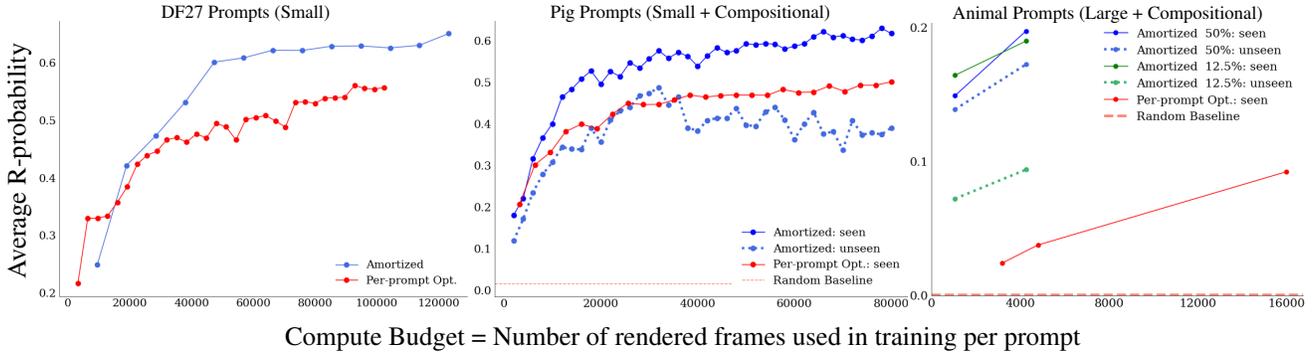

        % Reordered these to be inline with section 4?
        \noindent\textbf{Reduce training cost (Fig.~\ref{fig:all_quantitative}):}
            We train on text prompts for a fraction of the per-prompt cost.

        \noindent\textbf{Generalize to unseen prompts (Fig.~\ref{fig:compositional_amortization}, \ref{fig:animals-compositional}):}
            We seek strong performance when evaluating our model on unseen prompts during the amortized training without extra optimization.

        \noindent\textbf{Prompt interpolations (Fig.~\ref{fig:interpolation}):}
            Unlike current TT3D, we can interpolate between prompts, allowing:
                (a) generating a continuum of novel assets,
                or (b) creating 3D animations.
            %Furthermore, we can take the amortized model and finetune it to strong performance on a single prompt, as shown in Fig.~\ref{fig:finetuning_qualitative}
        
        % \noindent\textbf{Finetuning:}
        %     We want to rapidly finetune our model on arbitrary prompts, as shown in Fig.~\ref{fig:finetuning_qualitative}.
        
        % \noindent\textbf{Reduce visual artifacts:}
        %     The DreamFusion training has various standard failure modes, including (a) \emph{Janus-faces}, where a front view is rendered from all angles, and (b) mode collapse, including an all-black object, an empty scene, using the background to color components, or numerous others.
        %     Fig.~\ref{fig:janus_reduction} qualitatively shows that amortized training can help mitigate these issues.
            
     %\begin{figure}
        %\vspace{-0.0\textheight}
        %\centering
        %\begin{minipage}{0.35\textwidth}
            \begin{algorithm}%[H]
                \caption{
                    ATT3D Pseudocode for each update\\
                    Changes from DreamFusion Sec. 3 shown in {\color{red}red}
                    } \label{alg:amortized_training}
                \begin{algorithmic}[1]
                    \For{each loss term in batch}
                        \State {\color{red}sample a text and it's embedding $\textToken$}
                        \State {\color{red} compute the modulation $\mappingNet' = \mappingNet(\textToken)$}
                        \State sample camera position
                        \State add front/side/back to text, given camera
                        \State sample textureless/shadeless/full render
                        \State perform the render:
                        \State \quad create a ray for each pixel in the frame
                        \State \quad at each ray, sample multiple points $\position$
                        \State \quad at each point, compute encoding $\pointEncoder' \! =\! \pointEncoder_{\!{\color{red}\mappingNet'}}(\position)\!$
                        \State \quad at each point, compute the radiance $\nerfNet(\pointEncoder')$
                        \State \quad composite radiance into a frame
                        \State add noise to frame %$\noise \sim \mathcal{N}(0, \identity)$ to frame
                        \State compute denoised frame with the DDM via $\hat{\noise}$
                        %\State backprop so rendered close to denoised$-\epsilon$ frame
                        \State compute gradient using $\hat{\noise} - \noise$ as per SDS
                        %\State \emph{Note: DDM Jacobian approximated with $\identity$}
                    \EndFor
                \end{algorithmic}
            \end{algorithm}
            %\vspace{-0.01\textheight}
        %\vspace{-0.0\textheight}
        % \caption{
        %     TODO
        % }
    %\end{figure}
    % \section{Experimental Setup}\label{sec:experimental-setup}
%     %\noindent
%     We describe our problem setting in Sec.~\ref{sec:specific_problem_setting}, how we test our method in Sec.~\ref{sec:how_we_test}, which we combine into results in Sec.~\ref{sec:results}.
    
%     \subsection{Problem Setting}\label{sec:specific_problem_setting}
%         App. Sec.~\ref{sec:app-implementation-details} contains the full details for reproducing our setup, which is summarized here.

\section{Results and Discussion}\label{sec:results}
    %Show the results of combining our method with the experimental setup.
    %\noindent
    Here, we investigate our method's potential benefits.
    %The baseline is DreamFusion~\citep{poole2022dreamfusion}, which we equate to using our method on a single prompt -- see Fig.~\ref{fig:usVsDF}.
    We refer to the baseline as ``per-prompt optimization'', which follows existing works using separate optimization for each prompt.
    The specific NeRF rendering and SDS loss implementation are equivalent between the baseline and our method -- see Fig.~\ref{fig:usVsDF}.
    % \Kevin{maybe better way to describe the DF comparison in Fig 4}
    App. Sec.~\ref{sec:app-results} contains additional experiments, ablations, and visualizations.

    \vspace{0.005\textheight}
    \subsection{How We Evaluate ATT3D}\label{sec:how_we_test}
    %\vspace{-0.005\textheight}
        We first describe the datasets we use, then our metrics for quality and cost.
        \vspace{0.005\textheight}
        \subsubsection{Our Text Prompt Datasets}\label{sec:datasets}
        \vspace{0.005\textheight}
            %\vspace{-0.01\textheight}
            \textbf{DreamFusion (DF):}
                %We draft our first dataset from the closest related work -- DreamFusion (DF).
                The DF$27$ dataset consists of the $27$ prompts from DreamFusion's main paper, while DF$411$ has $411$ prompts from the project page.
                We explore memorizing these datasets but find them unsuitable for generalization.
            
            \noindent\textbf{Compositional:}
                To test generalization, we design a compositional prompt set by composing fragments with the template ``\emph{a} \{\texttt{animal}\} \{\texttt{activity}\} \{\texttt{theme}\}" and hold out a subset of ``unseen'' prompts. %animals/activities/themes not seen together.
                Our model must generalize to unseen compositions that require nontrivial changes to geometry.
                %We selected a handful of options inspired by DF$411$ 
                Using this template, we created a small pig-prompts and a larger animal-prompts dataset detailed in App. Sec.~\ref{sec:thePrompts} and shown in Figs.~\ref{fig:compositional_amortization} and \ref{fig:animals-compositional}.
                We hold out $\num{8}$ out of the $\num{64}$ pig prompts, as shown in Fig.~\ref{fig:compositional_amortization}.
                For the animals, the held-out prompts are sampled homogeneously and we investigate holding out larger fractions of the prompts.

        %\vspace{-0.015\textheight}
        \subsubsection{Our Evaluation Metrics}\label{sec:cost_metrics}
        \vspace{0.005\textheight}
        %\vspace{-0.01\textheight}
            \noindent
            \textbf{Cost:}
                We measure the computational cost of training per-prompt models versus our amortized approach.
                Wall-clock time and number of iterations are insufficient because we train with varying compute setups and numbers of GPUs -- see App. Sec.~\ref{sec:app-compute-requirements}.
                To account for this difference, we measure the number of rendered frames used for training (normalized by the number of prompts).
                Specifically, this is the number of optimization iterations times batch size divided by the total number of prompts in the dataset.

            \noindent
            \textbf{Quality:}
                \emph{CLIP R-(prec.)ision} is a text-to-3D correspondence metric introduced in Dream Fields~\citep{jain2021dreamfields}, defined as the CLIP model's accuracy at classifying the correct text input of a rendered image from amongst a set of distractor prompts (i.e., the \emph{query set}).
                \emph{CLIP R-(prob.)ability} is the probability assigned to the correct prompt instead of the binary accuracy, preserving information about confidence, and reducing noise. 
                %, which is important when we have few prompts.
                We found that R- metrics track each other (App. Fig.~\ref{fig:r-prec-vs-r-prob}), so we focus on R-prob.
                We evaluate R-prob. averaged over the input prompt dataset and four distinct rendered views as in DreamFusion~\citep{poole2022dreamfusion}, using the entire dataset as our query set.
                %The query set consists of all other prompts in the prompt set which tests whether the generations can be adequately distinguished from each other.
                The queries in DF$27$ are highly dissimilar, so we make the metric harder by adding the DF$411$ prompts to the query set.

        % \subsubsection{Our Quality Metrics}\label{sec:quality_metrics}
            
        %     \noindent\textbf{CLIP R-probability} -- closely related to CLIP R-precision -- is the probability CLIP assigns to the correct prompt instead of just the binary accuracy.
        %     The R-probability preserves information about the CLIP model's confidence making it less noisy, which is increasingly important when we have a small number of prompts.

            % \noindent\textbf{Visual assessment:}
            % Existing metrics (like R-prob.) do not measure visual fidelity perfectly.
            % For example, attaining maximal scores (of $1.0$) while a text prompt is still clearly improving.
            % Also, some visual artifacts -- like those in Fig.~\ref{fig:janus_reduction} -- are not detected by existing quantitative methods, but are by visual assessment.
            % Creating metrics to fix these deficiencies is an exciting research direction.
            % As such, we also display renderings of our method for qualitative assessment.

    %\vspace{-0.005\textheight}
    \subsection{Can We Reduce Training Cost?}
    %\vspace{-0.005\textheight}
        %\noindent
        Before evaluating generalization, we see if our method can optimize a diverse prompt collection faster than optimizing individually.
        %First, we try to memorize text prompt sets for a reduced cost.
        Fig.~\ref{fig:all_quantitative} gives the R-probability against compute budget for our method \& per-prompt optimization, showing we achieved higher quality for any budget.
        % App. Fig.~\ref{fig:r-prec-vs-r-prob} displays the same setup with R-precision, showing similar results.
        App. Figs.~\ref{fig:memorize-df27} and \ref{fig:df400_full}, qualitatively show we accurately memorize all prompts in DreamFusion's main paper and extended prompt set for a reduced cost -- perhaps from component re-use as in App. Fig.~\ref{fig:feature-reuse}.
        So, we have a powerful optimization method that quickly memorizes training data.

        But does the performance generalize to unseen prompts?
        Current TT3D methods optimize $1$ prompt, so any generalization is a valuable contribution.
        %\Kevin{Although we find some preliminary generalization on DF411 through our investigation in App. Fig.~\ref{fig:df411-zero-shot}, these prompt sets are too limited in size compared to their diversity.} 
        App. Fig.~\ref{fig:df411-zero-shot} shows unseen composed and interpolated prompts, with promising results, which we improve in Secs.~\ref{sec:generalizeNewPrompt} and \ref{sec:interpolationExperiments} respectively.

        % \begin{figure}%[H]
        %     \vspace{-0.00\textheight}
        %     \centering
        %     \begin{tikzpicture}
        %         \centering
        %         \node (img11){\includegraphics[trim={.0cm .75cm .0cm .75cm},clip,width=.95\linewidth]{images/quantitative_results/df27_no_rprec.png}};
        %         \node[left=of img11, node distance=0cm, rotate=90, xshift=1.6cm, yshift=-.9cm, font=\color{black}] {Average R-probability};
        %         \node[below=of img11, node distance=0cm, xshift=-.1cm, yshift=1.1cm,font=\color{black}]{Total rendered frames used in training};
        %         %{Iterations $\cdot$ \# GPUs vs. Iterations $\cdot$ \# Prompts};
        %         \node[above=of img11, node distance=0cm, xshift=.1cm, yshift=-1.2cm,font=\color{black}]{\scriptsize{DF27 Training Results: {\color{blue}Our Method (ATT3D)} vs. {\color{red}per-prompt Training}}};
        %     \end{tikzpicture}
        %     \vspace{-0.0275\textheight}
        %     \caption{
        %         We display the average R-probability against the number of frames used for optimization with our method in {\color{blue}blue} \& per-prompt training in {\color{red}red}.
        %         We train on DF27 -- the prompts from DreamFusion's paper shown in Fig.~\ref{fig:memorize-df27}.
        %         Results with R-precision are similar and shown in App. Fig.~\ref{fig:r-prec-vs-r-prob}.
        %         \textbf{Takeaway:} For any computational budget, we achieve higher quality.
        %     }
        %     \vspace{-0.015\textheight}
        %     \label{fig:training_quantitative}
        % \end{figure}
    
    \subsection{Can We Generalize to Unseen Prompts?}\label{sec:generalizeNewPrompt}
        Next, we investigate generalizing to unseen prompts with no extra optimization. 
        We used compositional prompt datasets to evaluate (compositional) generalization in the smaller pig and larger animal prompt datasets.
        %in a small ($64$ prompts) and larger regime ($2400$ prompts).
        Fig.~\ref{fig:all_quantitative} shows R-probability against compute budget on both seen \& unseen prompts for our method \& per-prompt optimization showing that we achieved higher quality for any compute budget on both prompt sets.
        Our generalization is especially evident in the larger prompt set, where we held out a significant fraction of the training prompts.
        With $50\%$ of prompts withheld, we have a minimal generalization gap.
        With only $12.5\%$ ($300$) prompts seen during training, generalization to \emph{unseen prompts} was better than per-prompt optimization on \emph{seen prompts} with only $\nicefrac{1}{4}$ the per-prompt compute budget.
        
\begin{figure*}
    \vspace{-0.01\textheight}
    \centering
    \begin{tikzpicture}
        \centering
        \node (img11){\includegraphics[trim={.0cm .0cm .0cm .0cm},clip,width=.85\linewidth]{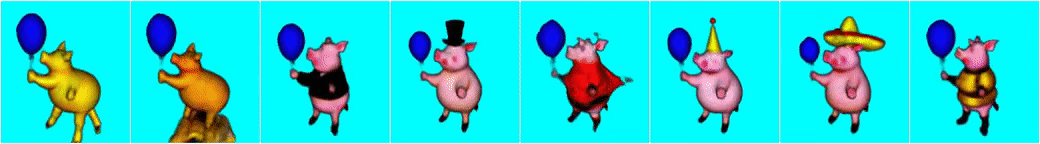}};
        %\node [right=of img11, xshift=-1cm](img12){\includegraphics[trim={.275cm .275cm .275cm .25cm},clip,width=.22\linewidth]{example-image-a}};
        \node[above=of img11, node distance=0cm, xshift=-.0cm, yshift=-1.2cm,font=\color{black}]{{\color{blue}Amortized Training}};
        \node[left=of img11, node distance=0cm, rotate=90, xshift=.9cm, yshift=-.9cm, font=\color{black}] {``\emph{...holding a blue balloon}"};
        
        \node [below=of img11, yshift=1cm](img21){\includegraphics[trim={.0cm .0cm .0cm .0cm},clip,width=.85\linewidth]{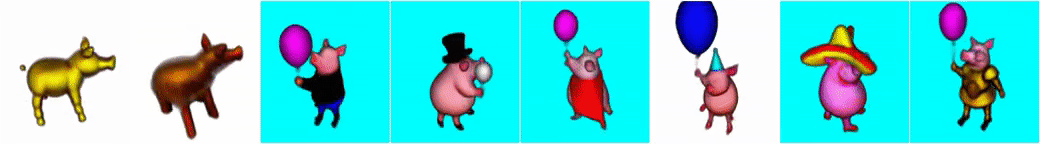}};
        \node[below=of img21, node distance=0cm, xshift=.0cm, yshift=1.2cm,font=\color{black}]{{\color{red}Per-prompt optimization}};
        %\node [right=of img21, xshift=-1cm](img22){\includegraphics[trim={.275cm .275cm .275cm .275cm},clip,width=.22\linewidth]{example-image-a}};
    \end{tikzpicture}
    \vspace{-0.015\textheight}
    \caption{
        We compare {\color{blue}amortized} and {\color{red}per-prompt} optimization on the prompts of the form ``\emph{...holding a blue balloon}."
        Amortization discovers a canonical orientation and always makes the balloon blue, while per-prompt training may only make the background blue or fail altogether, potentially explaining performance improvements in Fig.~\ref{fig:all_quantitative}.
    }
    \vspace{-0.01\textheight}
    \label{fig:pig-consistency}
\end{figure*}
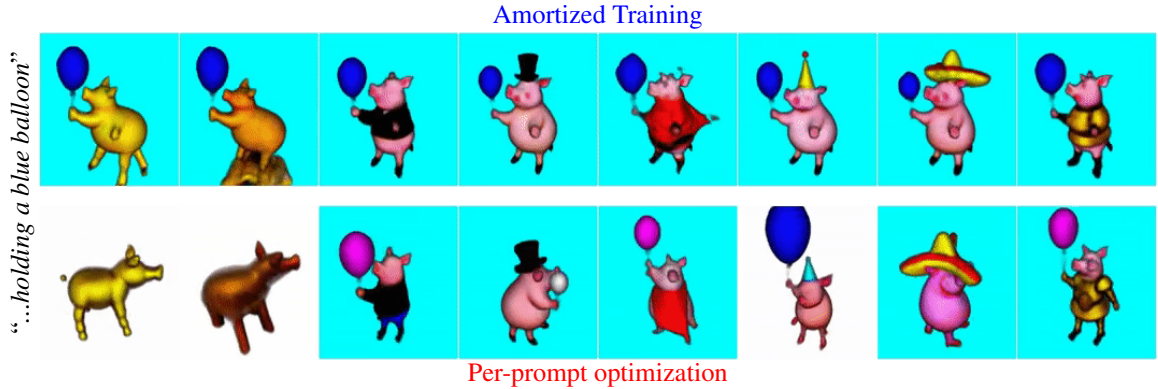

\begin{figure*}
        \vspace{-0.00\textheight}
        \centering
        \begin{tikzpicture}
            \centering
            % \hspace{-0.05\textwidth}
            % \node (img10){\includegraphics[trim={1.5cm .5cm 7.75cm 2.25cm},clip,width=.01\linewidth]{cvpr2023-author_kit-v1_1-1/latex/images/squirrel_comp.png}};

            \node (imgGrid){\includegraphics[trim={.0cm .0cm .0cm .0cm},clip,width=.54\linewidth]{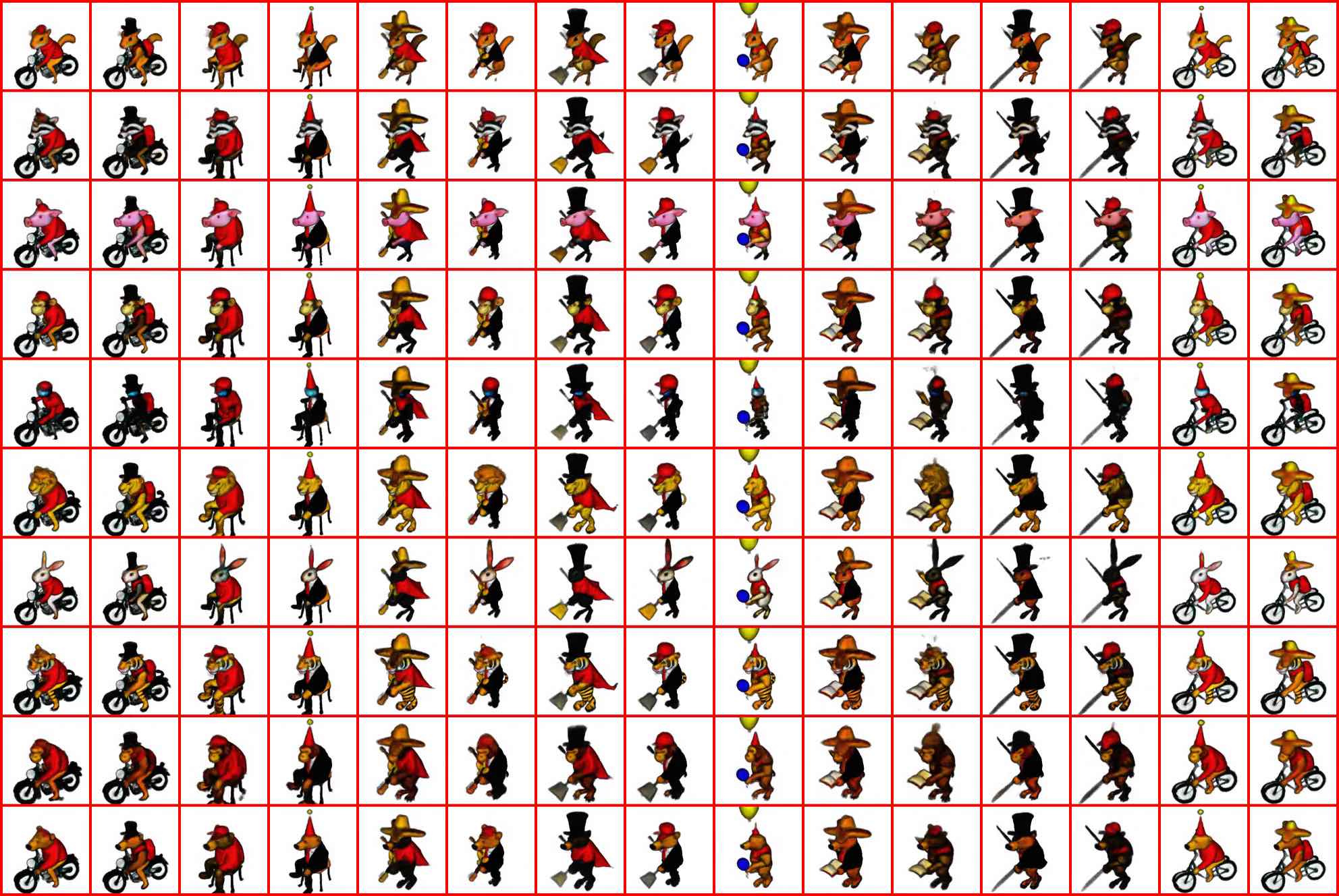}};
            \node[above=of imgGrid, node distance=0cm, xshift=.25cm, yshift=-1.2cm,font=\color{black}, rotate=0]{\footnotesize{Testing prompt for {\color{blue}Amortized 50\% split}, at $4800$}};

            \node [right=of imgGrid, xshift=-.75cm, yshift=1.5cm](img11){\includegraphics[trim={1.5cm .5cm 7.75cm 2.25cm},clip,width=.18\linewidth]{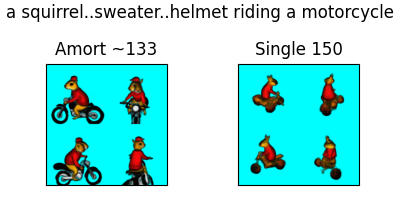}};
            \node[above=of img11, node distance=0cm, xshift=1.35cm, yshift=-1.2cm,font=\color{black}, rotate=0]{\footnotesize{Testing prompt for {\color{green}Amortized 12.5\% split}, at $4800$}};
            \node [right=of img11, xshift=-1.25cm](img12){\includegraphics[trim={1.5cm .5cm 14.75cm 2.0cm},clip,width=.16\linewidth]{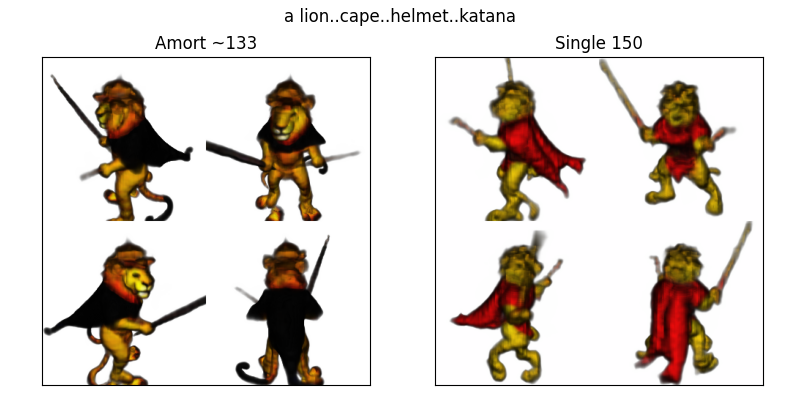}};
            
            \node [below=of img11, xshift=-0.05cm, yshift=1.0cm](img13){\includegraphics[trim={8.35cm .5cm 1.25cm 2.25cm},clip,width=.168\linewidth]{images/compositonal_animals/squirrel_comp.png}};
            \node [right=of img13, xshift=-1.1cm](img14){\includegraphics[trim={15.35cm .5cm 1.25cm 2.0cm},clip,width=.155\linewidth]{images/compositonal_animals/lion_comp.png}};
            \node[below=of img13, node distance=0cm, xshift=1.25cm, yshift=1.2cm,font=\color{black}, rotate=00]{\footnotesize{{\color{red}Per-prompt} at $4800$}};

        \end{tikzpicture}
        \vspace{-0.015\textheight}
        \captionof{figure}{
            We show quantitative results for the $2400$ animal prompts in Fig.~\ref{fig:all_quantitative}, where we achieve a higher quality for any compute budget on seen \& unseen prompts.
            Notably, when training on only ${\color{blue}50\%}$ or ${\color{green}12.5\%}$ of the prompts, the unseen prompts -- which cost no optimization -- perform stronger than the {\color{red}per-prompt} method, which must optimize on the data.
            \textbf{Takeaway:}
                By training a single model on many text prompts we generalize to unseen prompts without extra optimization.
        }
        \vspace{-0.01\textheight}
        \label{fig:animals-compositional}
     \end{figure*}

        To understand the superior performance, we visually compare a subset of pig prompts with the ``\emph{holding a blue balloon}'' activity in Fig.~\ref{fig:pig-consistency}. 
        ATT3D produced more consistent results than per-prompt optimization, potentially explaining our higher R-probability.
        Visualizations for the pig and animal experiments are in Figs.~\ref{fig:compositional_amortization} and \ref{fig:animals-compositional}, respectively.
        %We visualize optimization trajectories in App. Fig.~\ref{fig:pig_training}.
        This confirms we can achieve strong generalization performance with a sufficient prompt set.
        Further, quality can be improved with fine-tuning strategies (App. Fig.~\ref{fig:finetuning_qualitative}).

    \subsection{Can We Make Useful Interpolations?}\label{sec:interpolationExperiments}
        %\noindent
        Next, we investigate our method's ability to create objects as we interpolate between text prompts with no additional test-time optimization.
        In Fig.~\ref{fig:interpolation}, we show rendered outputs as we interpolate between different prompts.
        The output remains realistic with smooth transitions.
        
        For Fig.~\ref{fig:interpolation}, top right, we \emph{did not} use loss amortization and generalize to interpolants while only training on the $3$ rock prompts.
        But, some prompts gave suboptimal results without interpolant training (App. Fig.~\ref{fig:df411-zero-shot}) which we improved by interpolant amortization (Sec.~\ref{sec:AmortizedInterpolation}).
        We evaluated several prompt interpolation approaches.
        App. Fig.~\ref{fig:interpolation_hamburgers} compares $3$ interpolant amortization types: loss weightings, interpolated embeddings, and guidance weightings, showing various ways to control results.
        App. Fig.~\ref{fig:interpolation_ships} compares different interpolant sampling strategies during training, providing qualitatively different ways to generate assets.
        % \Jon{This section could use another set of eyes too.}

    \vspace{-0.005\textheight}
\section{Related Work}
\vspace{-0.005\textheight}
    %Compare and contrast. If method applicable to our setting, expect comparison.
    We cover the various fields our method combines:
        (a) text-to-image generation, then 
        (b) image-to-3D models, which lead to 
        (c) text-to-3D models, which we augment with 
        (d) amortized optimization.

    \vspace{-0.0175\textheight}
    \noindent
    \paragraph{Text-to-image Generation:}
        (A)TT3D methods~\citep{poole2022dreamfusion, lin2022magic3d, wang2022score} use large-scale text-conditional DDMs~\citep{ho2022imagen, ramesh2022hierarchical, rombach2022high, balaji2022ediffi, deepfloyd}, which train using classifier-free guidance to sample images matching text prompts~\citep{ho2022classifier}.
        While these models generate diverse and high-fidelity images for many prompts, they cannot provide view-consistent renderings of a single object and are thus incapable of making 3D assets directly.
        % Notably, the models condition on text-embeddings from text-encoding models like T5-XXL~\citep{JMLR:v21:20-074} and CLIP~\citep{pmlr-v139-radford21a}.

    \vspace{-0.0175\textheight}
    \noindent
    \paragraph{Image-to-3D Models:}
        %NeRFs and image-to-3D methods (mesh, etc?)
        Beyond using 3D assets to train 3D generative models, prior work has also used image datasets.
        Most of these methods use NeRFs~\citep{mildenhall2021nerf, rebain2022lolnerf, rebain2022attention, chan2021pi, melas2023realfusion, tang2023make} as a differentiable renderer optimized to produce image datasets.
        Differentiable mesh rendering is an alternative~\citep{pavllo2021learning, gao2022get3d, pavllo2020convolutional, chen2019learning}.
        \citet{Chan_2022_CVPR} are closely related in this category, using a StyleGAN generator modulated with a learned latent code to produce a triplanar grid that is spatially interpolated and fed through a NeRF producing a static image dataset.
        We also modulate spatially oriented feature grids, without relying on memory-intensive pre-trained generator backbones.
        These techniques may prove valuable in future work scaling to ultra-large prompt sets.
    
    %\vspace{0.02\textheight}
    \noindent
    \paragraph{Text-to-3D Generation:}
        %Dreamfusion and other text-to-3d-methods with less quality
        % \paragraph{Geometry-based methods} 
        %     Point clouds: LION, Point-voxel diffusion. Mesh-based: CLIP-mesh, GET3D, etc. 
        % \paragraph{Dreamfusion}
        Recent advances include CLIP-forge~\cite{sanghi2021clip}, CLIP-mesh~\cite{khalid2022clipmesh}, Latent-NeRF~\cite{metzer2023latent}, Dream Field~\cite{jain2021dreamfields}, Score-Jacobian-Chaining~\citep{wang2022score}, \& DreamFusion~\cite{poole2022dreamfusion}.
        In CLIP-forge~\cite{sanghi2021clip}, the model is trained for shapes conditioned on CLIP text embeddings from rendered images.
        During inference, the embedding is provided for the generative model to synthesize new shapes based on the text.
        CLIP-mesh~\cite{khalid2022clipmesh} and Dream Field~\cite{jain2021dreamfields} optimized the underlying 3D representation with the CLIP-based loss.
        % DreamFusion~\cite{poole2022dreamfusion} and Score-Jacobian-Chaining~\citep{wang2022score} optimize NeRFs with a loss from a text-to-image DDM~\cite{ho2022imagen}.
        Magic3D adds a finetuning phase with a textured-mesh model~\citep{shen2021deep}, allowing high resolutions.
        Future advances may arise by combining with techniques from unconditional 3D generation~\citep{bautista2022gaudi, zeng2022lion, zhou20213d}.
        Notable open-source contributions are Stable-Dreamfusion~\cite{stable-dreamfusion} and threestudio~\cite{threestudio2023}.
        Other concurrent works include Zero-1-to-3~\cite{liu2023zero}, Fantasia3D~\cite{chen2023fantasia3d}, Dream3D~\cite{xu2023dream3d}, DreamAvatar~\cite{cao2023dreamavatar}, and ProlificDreamer~\cite{wang2023prolificdreamer}. 
        However, we differ from all of these text-to-3D works, because we amortize over the text prompts.

    \noindent
    \paragraph{Amortized Optimization:}
        Amortized optimization~\citep{amos2022tutorial} is a tool of blossoming importance in learning to optimize~\citep{chen2021learning} and machine learning, with applications to meta-learning~\citep{hospedales2021meta}, hyperparameter optimization~\citep{lorraine2018stochastic, mackay2018self}, and generative modeling~\citep{kingma2013auto, rezende2014stochastic, cremer2018inference, wu2020meta}.
        Hypernetworks~\citep{ha2016hypernetworks} are a popular tool for amortization~\citep{lorraine2018stochastic, mackay2018self, zhang2018graph, knyazev2021parameter} and have also been used to modulate NeRFs~\citep{sitzmann2020metasdf, rebain2022attention, dupont2022data}, inspiring our strategy.
        Our method differs from prior works by modulating spatially oriented parameters, and our objective is from a (dynamic) DDM instead of a (static) dataset.

    \noindent
    \paragraph{Text-to-3D Animation:}
        Text-to-4D~\cite{singer2023text} is an approach for directly making 3D animations from text, instead of our interpolation strategy.
        This is done by generalizing TT3D to use a text-to-video model~\cite{singer2022make, ho2022imagen, blattmann2023align}, instead of a text-to-image model.
        However, unlike us, this requires text-to-video, which can require video data.
        
    \begin{figure} %[H]
        \vspace{-0.005\textheight}
        %\centering
        \begin{tikzpicture}
            %\centering
            \node (img11){\includegraphics[trim={.5cm .5cm 42.5cm .35cm},clip,width=.98\linewidth]{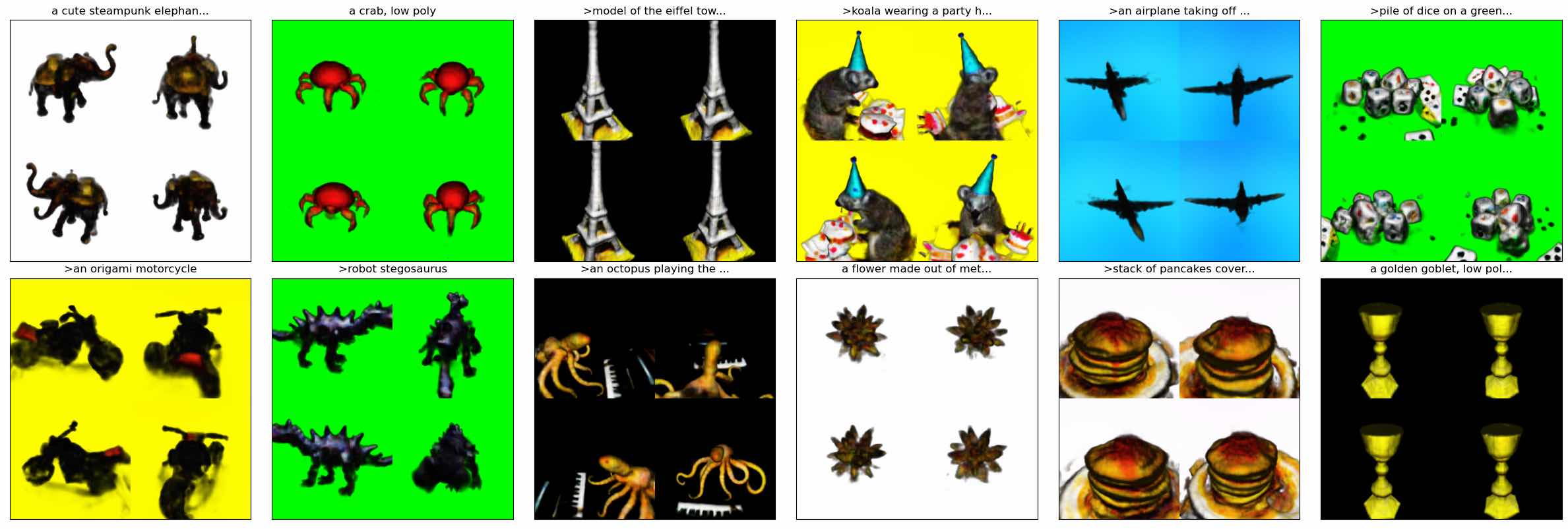}};
            %\node[above=of img11, node distance=0cm, xshift=-.1cm, yshift=-1cm,font=\color{black}]{per-prompt Training};
        \end{tikzpicture}
        \vspace{-0.025\textheight}
        \caption{
            Results for amortized training on DreamFusion's extended set of $411$ text prompts, DF$411$.
            See Fig.~\ref{fig:df400_full} for the full set.
            \textbf{Takeaway:} We scale to diverse prompt sets $>\!\!10\times$ larger than DF27 (Fig.~\ref{fig:memorize-df27}) with minor quality drop.
        }\label{fig:df400_qualitative}
        \vspace{-0.02\textheight}
    \end{figure}
    \newpage

\vspace{-0.005\textheight}
\section{Conclusion}\label{sec:conclusion}
    \vspace{-0.005\textheight}
    % \Kevin{
    We presented ATT3D, a method for amortized optimization of text-to-3D (TT3D) models.
    We use a mapping network from text to NERFs, enabling a single model to represent 3D objects of many different prompts.
    We experimentally validate our method on existing and new compositional prompt sets.
    We are faster at training than current TT3D methods by sharing the optimization cost across a prompt set.
    Once trained, our model generalizes by directly outputting objects for prompts unseen during training in a single forward pass.
    Furthermore, by amortizing over interpolation weights, we quickly generate a continuum of interpolations between prompts, enhancing user control.
    % }
    % Brief recap of paper
    %We presented a method for amortized optimization of text-to-3D models.
    %We empirically verified that our method:
    % (a) reduces compute cost on training prompts,
    % (b) generalizes to new prompts,
    % (d) allows continuous interpolations,
    % (e) easily amortizes over other setups.
    % \Kevin{
    
    Although ATT3D only represents a small step towards general and fast text-to-3D generation, we believe that the ideas presented are a promising avenue toward this future.
    % }
    % \Jon{I edited up the conclusion with Kevin's material.  This could use another set of eyes}

    %\vspace{-0.005\textheight}
    \paragraph{Limitations:}
        Our method builds on the existing text-to-3D optimization paradigm, so we share several limitations with these works:
        %We can improve our quality with more powerful DDMs. \Kevin{Quality and robustness of generations can depend a lot on the text-to-image DDM used.}
        More powerful text-to-image DDMs may be required for higher quality and robustness in results.
        The objective has high variance, and the system can be sensitive to prompt engineering.
        We also suffer from a lack of diversity, as in prior work. %, which may be alleviated by incorporating a generative modeling framework.
        % Also, our single prompt implementation under-performs the original qualitatively. \Kevin{This is not clear right now}\Jon{Happy to drop this.}
        We found that similar prompts can collapse to the same scene when amortizing.
        Finally, larger object-centric prompt sets are required to further test the scaling of amortized training.
    
    % Amazing things we want to do in the future
    %\paragraph{Future work:}
        % We amortized over text prompts, but it would be valuable for future work to re-purpose these tools to amortize over:
        % (i) noise for generative models of per-prompt 3D scenes and
        % (ii) over the time axis of text-to-video~\citep{singer2022make, ho2022imagen} to create text-to-3D-animations.
        %A critical extension would be finding more extensive and complex prompt sets to explore larger-scale generalization behavior.
        % \Jon{Other? Ex., detecting Janus faces}

    % We presented a method for
    % Our method has benefits of X, which we empirically validated
    % We are excited about 

    %\vspace{-0.005\textheight}
    \paragraph{Ethics Statement:}
        Text-to-image models carry ethical concerns for synthesizing images, which text-to-3D models like this share. 
        For example, we may inherit any biases in our underlying text-to-image model.
        These models could displace creative jobs or enable the growth and accessibility of 3D asset generation.
        Alternatively, 3D synthesis models could be used to generate misinformation by bad actors.
    
    %\vspace{-0.005\textheight}
    \paragraph{Reproducibility Statement:}
        Our instant-NGP NeRF backbone is publicly available through the ``instant-ngp" repository~\citep{muller2022instant}.
        While our diffusion model is not publicly available (as in DreamFusion~\citep{poole2022dreamfusion}), other available models may be used to produce similar results.
        To aid reproducibility, we include a method schematic in Fig.~\ref{fig:amortized-text-to-3d-pipeline} and pseudocode in Alg.~\ref{alg:amortized_training}.
        Our evaluation setup is in Sec.~\ref{sec:how_we_test} along with hyperparameters and other details in App. Sec.~\ref{sec:app-experimental-setup}.

    \vspace{-0.01\textheight}
    \section*{Acknowledgements}\label{sec:ack}
    \vspace{-0.005\textheight}
        \noindent
        We thank Weiwei Sun, Matan Atzmon, and Or Perel for helpful feedback.
        The Python community ~\citep{van1995python, oliphant2007python} made underlying tools, including PyTorch~\citep{paszke2017automatic} \& Matplotlib~\citep{hunter2007matplotlib}.

    \vspace{-0.005\textheight}
    \section*{Disclosure of Funding}
    \vspace{-0.005\textheight}
        \noindent
        NVIDIA funded this work.
        Jonathan Lorraine, Kevin Xie, Xiaohui Zeng, and Towaki Takikawa had funding from student scholarships at the University of Toronto and the Vector Institute, which are not in direct support of this work.
    
    %-------------------------------------------------------------------------
    %\subsection{References}
        
    %\newpage
    %%%%%%%%% REFERENCES
    \vspace{-0.005\textheight}
    {\small
        \bibliography{egbib}
    }
    \hspace{0.01\textwidth}
    \begin{table*}[h]\caption{Glossary and notation}
    \vspace{-0.03\textheight}
    \begin{center}
        \begin{tabular}{c c}
            \toprule
            % $x, y, z, \dots \in \mathbb{C}$ & Scalars\\
            % $\boldsymbol{x}, \boldsymbol{y}, \boldsymbol{z}, \dots \in \mathbb{C}^{n}$ & Vectors\\
            % $\boldsymbol{X}, \boldsymbol{Y}, \boldsymbol{Z}, \dots \in \mathbb{C}^{n \times n}$ & Matrices\\
            % $\boldsymbol{X}^\transpose$ & The transpose of matrix $\boldsymbol{X}$\\
            % $\identity$ & The identity matrix\\
            % \JL{TODO - see commented latex here} &\\
            % \JL{Jargon, like in Table 1 of Hpo-B} & \\
            (A)TT3D & (Amortized) Text-to-3D\\
            NeRF & Neural Radiance Field~\citep{mildenhall2021nerf}\\
            DDM & Denoising Diffusion Model\\
            MLP & Multi-layer Perceptron\\
            DF$27$, DF$411$ & DreamFusion's ~\citep{poole2022dreamfusion} $27$ main text prompts \& the extended $411$ prompts\\
            $n, m \in \mathbb{N}$ & The size of different objects\\
            $x, y, z, \dots \in \mathbb{R}$ & Scalar coordinates\\
            $\boldsymbol{x}, \boldsymbol{y}, \boldsymbol{z}, \dots \in \mathbb{R}^{n}$ & Vectors\\
            % $\boldsymbol{X}, \boldsymbol{Y}, \boldsymbol{Z}, \dots \in \mathbb{R}^{n \times m}$ & Matrices\\
            $\mathcal{X}, \mathcal{Y}, \mathcal{Z}, \dots$ & The domain of $\boldsymbol{x}, \boldsymbol{y}, \boldsymbol{z}, \dots $ \\
            % $\identity$ & The identity matrix\\
            $\position = [x, y, z] \in \positionDom$ & A point\\
            $\density = [\sigma, r, g, b] \in \densityDom$ & The density and color values\\
            $\pointEncoderParams \in \pointEncoderParamsDom$ & The parameters of the point encoder function\\
            $\pointEncoder_\pointEncoderParams: \positionDom \to \encoderRange$ & The point encoder function\\
            $\nerfNet: \encoderRange \to \densityDom$ & The final MLP mapping point encodings to radiance\\
            $\textEmbedding$ & The problem context for amortization\\
            $\textToken \in \textTokenDom$ & A text embedding used to condition the DDM and as problem context\\
            $\mappingNet: \textTokenDom \to \pointEncoderParamsDom$ & The mapping network from problem context to modulations\\
            $\vectorEmbedding \in \mathbb{R}^{n}$ & The intermediary vector-embedding of $\textToken$ in $\mappingNet$\\
            %$\embeddingNet$ & The embedding network from text token embedding $\textToken$ to $\textEmbedding$\\
            $\mathcal{N}, \mathcal{U}, \textnormal{Dir}, \textnormal{Bern}$ & Normal, uniform, Dirichlet, and Bernoulli distributions respectively\\
            $\noise, \hat{\noise}$ & Noise added to rendered frames, or as predicted by the DDM\\
            $\kappa \in \mathbb{R}^{+}$ & The concentration parameter of the Dirichlet distribution\\
            $\alpha \in [0, 1]$ & An interpolation coefficient, sampled from $\textnormal{Dir}(\kappa)$ in training\\
            $\loss$ & A loss function\\
            $\omega$ & A guidance weight\\
            $\beta_1, \beta_2$ & Parameters of the Adam optimizer~\citep{kingma2013auto}\\
            \bottomrule
        \end{tabular}
    \end{center}
    \label{tab:TableOfNotation}
    \vspace{-0.03\textheight}
\end{table*}
    \newpage
    %\hspace{0.01\textwidth}
    % \newpage
    % \hspace{0.01\textwidth}
    %\newpage

    % \clearpage
    
    \appendix
    %\section*{Appendix}
    % \input{appendix/glossary.tex}
    \section{\vspace{-0.001\textheight}Glossary}

    % \vspace{-0.01\textheight}
\section{Experimental Setup}\label{sec:app-experimental-setup}
        % \vspace{-0.01\textheight}
        \subsection{Implementation Details}\label{sec:app-implementation-details}
        \vspace{-0.01\textheight}
            %\textbf{Embedding network $\embeddingNet$:}
            %    \TODO{Describe our embedding networks}
            \noindent
            We replicate DreamFusion~\citep{poole2022dreamfusion} and Magic3D's~\citep{lin2022magic3d} setup where possible and list key details here.
            We recommend reading these papers for additional context.

            \vspace{-0.025\textheight}
            \subsubsection{Point-encoder $\pointEncoder$}
            \vspace{-0.015\textheight}
                We followed Instant NGP~\citep{muller2022instant} to parameterize our NeRF, consisting of dense, multi-resolution voxel grids and dictionaries.
                We only use dense voxel layers unless specified, which trained faster with negligible quality drop.
                % \footnote{We found that training was faster with only dense voxel layers, while the quality drop was negligible.}
                For our multi-resolution voxel grid, we use resolutions of $[\num{9}, \num{14}, \num{22}, \num{36}, \num{58}]$, with $\num{4}$ features per level.
                When active, we use a further three levels of hash grid parameters.
                Each level's features are linearly interpolated according to spatial location and concatenated, leading to a final output feature size of $\num{20}$ with dense voxel grids and $\num{32}$ with the full INGP.

            \vspace{-0.025\textheight}
            \subsubsection{Final NeRF MLP $\nerfNet$}
            \vspace{-0.015\textheight}
                We select a minimal final MLP to maintain evaluation speed, with a single hidden layer with 32 units and a SiLU activation~\citep{hendrycks2016gaussian}.
                The majority of our model's capacity comes from the point-encoder.
                We use a softplus activation for the density output and sigmoid activations on the color.
                
            \subsubsection{Mapping Network $\mappingNet$}\label{sec:app_mapping_network}
            \vspace{-0.015\textheight}
                The mapping network computes a fixed-size vector representation $\vectorEmbedding$ of the task from the text embedding.
                We only use the CLIP embedding for feature grid modulation because it was sufficient and including the T5 embedding increases network size.
                We apply spectral normalization to all linear layers.
                We considered concatenation, hypernetwork, and attention approaches for modulation~\citep{rebain2022attention}:

                \noindent
                \textbf{Concatenation:}
                    The simple strategy of nai\"vely concatenating (a vector-representation $f$ of) the text to the point-encoding -- i.e., $\nerfNet(\pointEncoder_{\pointEncoderParams}(\position), f(\textToken))$ was prohibitively expensive.
                    This is because we require the cost of the final per-point NeRF MLP $\nerfNet$ to be minimal for cheap rendering.
                    % Instead, we efficiently modulate $\pointEncoder$ with $\textToken$ via $\pointEncoderParams$ with a hypernetwork.
                    The concatenation approach of $\nerfNet(\pointEncoder_{\pointEncoderParams}(\position), \vectorEmbedding)$ introduces overhead by increasing per-iteration training time by $37\%$ but doesn't significantly impact quality.
                    In inference, the hypernet method is superior, reducing cost by $\sim 20 - 75\%$, as we only generate the grid parameters $\pointEncoderParams$ once when rendering multiple views of $1$ object, bypassing the use of a single, larger NeRF $\nerfNet$ with concatenation.
                    %\TODO{I think this is missing some words?}

                % \newpage
                \noindent
                \textbf{Hypernetwork:}
                    We first flatten the token and pass it through an MLP to produce a vector-embedding, which is used by a linear layer to output the point-encoder's voxel grid parameters.
                    As the CLIP embedding was already a strong representation, we found that a simple linear layer for the text-embedding to vector-embedding was sufficient.
                    
                    We converged with deeper hypernetworks -- by using spectral normalization on all linear layers -- but this offered no quality benefit while taking longer to train.
                    We also found removing the bias on the final linear layer decreased noise by forcing the result to depend on the prompt.
                    
                    We vary the vector embedding $\vectorEmbedding$'s size in our experiments, which largely dictates our amortized model's capacity.
                    The mapping network $\mappingNet$ dominates the model's memory cost, while the text's vector-embedding largely dictates the mapping network size.
                    Our memory cost scales linearly with the vector-embedding size.
                    We use a vector embedding $\vectorEmbedding$ size of $\num{32}$ for all experiments except interpolation, where we use $\num{2}$.
                    We have experiments where the number of text prompts is both smaller (DF$27$) and larger (DF$411$, compositional prompts) than the vector-embedding.
                
                \noindent
                \textbf{Attention:}
                    We also investigated using an attention-based mapping network with a series of self-attention layers to process text embeddings before feeding into the hypernetworks for each multi-resolution grid level.
                    Our attention performed with comparable quality but trained more slowly.
                    However, we expect modifications to be necessary on more complex prompt sets.

                % \James{Did we use any of the grid factorization techniques in the end? If so, would be good to mention here. E.g. for attention I did a matrix outer-product factorization of the codebook and the voxel grid to keep compute/memory overhead low.}
                % \Jon{I experimented with some simple variants -- especially for the hash-component -- but it was better to use no hash.  Factorizing the multi-resolution grid was not necessary for size 32 latent, but would be neat for much larger.}
                
            \subsubsection{Environment Mapping Network}
                In our experiments, we use a background, a function mapping ray directions -- and text embeddings -- to colors, which we denote as the environment map.
                Specifically, we encode the ray directions, concatenate them with the vector-embedding $\vectorEmbedding$ from the mapping network, and feed them into a final MLP.
                We use a sigmoid activation on the output color and spectral norm on all linear layers.
                We encode the ray directions with a sinusoidal positional encoding~\citep{vaswani2017attention} (frequencies $2^0, 2^1, \dots, 2^{L-1}, L=8$), and no hidden layers — i.e., a linear layer — for our final MLP.

            \subsubsection{Spectral Normalization}
                We found spectral normalization -- which can be implemented trivially in PyTorch on linear layers -- to be critical for mapping net training, but non-essential on other parts.
                In the mapping network, we must use spectral normalization on all linear layers for the hypernetwork and attention approaches or we suffer from numerical instability.
                Using spectral normalization on the linear layers in the environment map, or final NeRF module was unnecessary.
            
            \subsubsection{Sampling Text Prompts}\label{sec:app-sampling-text}
                We cache the CLIP (and T5) embeddings for all experiments to avoid repeated computation and the memory overhead of the large text encoders.
                We use multiple text prompts in each batched update.
                
                \noindent
                \textbf{Interpolations:}
                    We sample interpolated embeddings during training in interpolation experiments (Section~\ref{sec:interpolationExperiments}). 
                    See Section~\ref{sec:app-interpolation-exp} or Figure~\ref{fig:interpolation_hamburgers} for more interpolation setup details.
                    When interpolating between prompts with text-embeddings $\textToken_1$ and $\textToken_2$, we sample a weight $\alpha \in [0, 1]$ and input $\textToken' \!=\! (1 - \alpha) \textToken_1 + \alpha \textToken_2$ to the mapping network.
                
                % \noindent
                % \textbf{Loss-space interpolations:}
                %     We include additional results in Figure~\ref{fig:interpolation_hamburgers}, where we train with interpolations between the loss associated with each text prompt instead of their embeddings, showing we can produce qualitatively different results by amortizing over various training strategies.
                %     Notably, we train over the loss interpolations efficiently via amortization.
            
            \subsubsection{Sampling Rendering Conditions}
                As in DreamFusion~\citep{poole2022dreamfusion}, we randomly sample rendering conditions, including the camera position and lighting conditions.
                We use a bounding sphere of radius $2$ in all experiments.
                We sample the point light location with distance from $\mathcal{U}(\num{1}, \num{3})$ and angle relative to the random camera position of $\mathcal{U}(\num{0}, \nicefrac{\pi}{4})$.
                We sample ``soft" textureless and albedo-only augmentations to allow varying shades during training.
                Also, we sample the camera distance from $\mathcal{U}(\num{2}, \num{3})$ and the focal length from $\mathcal{U}(\num{.7}, \num{1.35})$.
                    
            \subsubsection{Score Distillation Sampling}
                For the DDM's sampling, we sample the time-step from $\mathcal{U}(\num{0.002}, \num{1.0})$ and use a guidance weight of $100$.
                
            \subsubsection{The Objective}\label{sec:app-objective}
                \textbf{The regularizers:}
                    The orientation loss~\citep{verbin2022ref} (as in DreamFusion~\citep{poole2022dreamfusion}) encourages normal vectors of the density field to face the camera when visible, preventing the model from changing colors to be darker during textureless renders by making geometry face ``backward" when shaded.
                    Also, DreamFusion regularizes accumulated alpha value along each ray, encouraging not unnecessarily filling space and aiding in foreground/background separation.
                    We do not use these regularizers for all experiments, as we did not observe failure modes they fixed, and they made no significant change in results over the interval $[10^{-3}, 10^{-1}]$.
                    Larger opacity regularization values resulted in empty scenes, while larger orientation values did not change the initialization from a sphere.
                
                \noindent
                \textbf{The image fidelity:}
                    We train with $32$ points sampled uniformly along each ray for all experiments except interpolations.
                    For interpolations only, we sample $128$ points and reduced batch size to improve quality.
                    Our underlying text-to-image model generates $\num{64} \times \num{64}$ images, leading to $\num{4096}$ rays per rendered frame.
                    At inference time we render with higher points per ray to improve quality for negligible cost.
                
                \noindent
                \textbf{The initialization:}
                    As in DreamFusion, we add an initial spatial density bias to prevent collapsing to an empty scene, shown in Figure~\ref{fig:pig_training}, left.
                    Our density bias on the NeRF MLP output before the softplus activation takes the form:
                    \begin{equation}
                        \textnormal{densityBias}(\position) = 10 \left( 1 - 2\|x\|_2 \right)
                    \end{equation}
                
            \subsubsection{The Optimization}
                We use Adam with a learning rate of $\num{1}\times10^{-1}$ and $\beta_2 = \num{.999}$.
                A wide range momentum $\beta_1$ (up to .95) can yield similar qualities if the step size is jointly tuned, while the quickest convergence occurs at $0$.
                We do not use the linear learning rate warmup or cosine decay from DreamFusion.
                
            % \subsubsection{Memorization \& Zero-shot Generalization \& Finetuning Experiments:}
            %     \TODO{}
                
            \subsubsection{Memorization Experiments}
                Our experiments use the same architecture for per-prompt and amortized training settings to ensure a fair comparison.
                We train models using a batch size of 32 times the number of GPUs used.
                Amortized training uses 8 GPUs while per-prompt uses a single GPU (due to resource constraints), with more details in Section~\ref{sec:app-compute-requirements}.
                The complete set of DreamFusion prompts is located here: \url{https://dreamfusion3d.github.io/gallery.html}
            
            \subsubsection{Generalization Experiments}\label{sec:thePrompts}
                Our prompt selections are motivated by the compositional experiment in DreamFusion's Figure 4~\citep{poole2022dreamfusion}.
                Our experiments with pig prompts with the template ``\emph{a pig \{activity\} \{theme\}}, where the \texttt{activities} and \texttt{themes} are any combination of the following:
                
                \noindent
                \textbf{The \texttt{activities}:}
                    %\newline
                    [
                        ``\emph{riding a bicycle}",
                        ``\emph{sitting on a chair}",
                        ``\emph{playing the guitar}",
                        ``\emph{holding a shovel}",
                        ``\emph{holding a blue balloon}",
                        ``\emph{holding a book}",
                        ``\emph{wielding a katana}",
                        ``\emph{riding a bike}"]

                \noindent
                \textbf{The \texttt{themes}:}
                    %\newline
                    [
                        ``\emph{made out of gold}",
                        ``\emph{carved out of wood}",
                        ``\emph{wearing a leather jacket}",
                        ``\emph{wearing a tophat}",
                        ``\emph{wearing a party hat}",
                        ``\emph{wearing a sombrero}",
                        ``\emph{wearing medieval armor}"]
                
                Our pig holdout, unseen, testing prompts are pairing the $i^{th}$ \texttt{activity} and \texttt{theme}.
                
                Our experiments with animal prompts with the template ``\emph{\{animal\} \{activity\} \{theme\} \{hat\}}, where the \texttt{activities}, \texttt{themes} and \texttt{hats} are any combination of the following:

                \noindent
                \textbf{The \texttt{animals}:}
                    %\newline
                    [
                        ``\emph{a squirrel}",
                        ``\emph{a raccoon}",
                        ``\emph{a pig}",
                        ``\emph{a monkey}",
                        ``\emph{a robot}",
                        ``\emph{a lion}",
                        ``\emph{a rabbit}",
                        ``\emph{a tiger}",
                        ``\emph{an orangutan}",
                        ``\emph{a bear}"]

                \noindent
                \textbf{The \texttt{activities}:}
                    %\newline
                    [
                        ``\emph{riding a motorcycle}",
                        ``\emph{sitting on a chair}",
                        ``\emph{playing the guitar}",
                        ``\emph{holding a shovel}",
                        ``\emph{holding a blue balloon}",
                        ``\emph{holding a book}",
                        ``\emph{wielding a katana}"]

                \noindent
                \textbf{The \texttt{themes}:}
                    %\newline
                    [
                        ``\emph{wearing a leather jacket}",
                        ``\emph{wearing a sweater}",
                        ``\emph{wearing a cape}",
                        ``\emph{wearing medieval armor}",
                        ``\emph{wearing a backpack}",
                        ``\emph{wearing a suit}"]

                \noindent
                \textbf{The \texttt{hats}:}
                    %\newline
                    [
                        ``\emph{wearing a party hat}",
                        ``\emph{wearing a sombrero}",
                        ``\emph{wearing a helmet}",
                        ``\emph{wearing a tophat}",
                        ``\emph{wearing a backpack}",
                        ``\emph{wearing a baseball cap}"]

                \noindent
                Our holdout, unseen, animal testing prompts are selected homogeneously for each training set size.
            
            \subsubsection{Finetuning Experiments}
                We resume training from an amortized training checkpoint while re-initializing the optimizer state.
                For the finetuning experiments, in our mapping network, we only finetune an offset to the output and detach all prior weights that only embed text tokens (because we finetune with one prompt).
                %We detach all preceding terms which find a vector-embedding of the text token embedding, because 
                %
                %or
                %
                %We illustrate the differences qualitatively in \TODO{plot}. 
                %Restricting the finetuning to only spatial parameters helps preserve the cohesion between the different prompts compared to finetuning the whole network.
                Other training details are kept equal to per-prompt training.
            
            \subsubsection{Interpolation Experiments}\label{sec:app-interpolation-exp}
                In interpolations we use $\num{128}$ ray samples and batch size $\num{16}$.

                \noindent
                \textbf{Interpolant concentration:}
                    We sample the interpolation coefficient $\alpha \sim \textnormal{Dir}(\kappa)$ from a Dirichlet distribution with concentration parameter $\kappa$.
                    The Dirichlet distribution allows us to smoothly interpolate from sampling the original text tokens (with concentration $\kappa \approx 0$, to uniformly sampling $\alpha$ (with concentration $\kappa \approx 1$) to focusing on difficult midpoints (with concentration $\kappa > 1$) -- see Figure~\ref{fig:interpolation_ships}.
                    Specifically, in Figures~\ref{fig:interpolation} \& \ref{fig:interpolation_other} we use $\kappa = \num{2.0}$ for $\num{5000}$ steps to stabilize the midpoint, followed by $\kappa = \num{0.5}$ to focus on the original prompts.

                \noindent
                \textbf{Interpolation types:}
                    We provide multiple examples of interpolation types to amortize over that provide qualitatively different results -- see Figure~\ref{fig:interpolation_hamburgers}.

                    A simple strategy is to interpolate over the text embedding used to condition the text-to-image model:
                    \begin{equation}
                        \textToken' = (1 - \alpha) \textToken_1 + \alpha \textToken_2
                    \end{equation}

                    \noindent
                    Another strategy is to interpolate the loss function used between the two prompts.
                    We could evaluate the loss at both prompts and weight the loss:
                    \begin{equation}
                        \loss_{\textnormal{final}} = (1 - \alpha) \loss_{\textnormal{prompt 1}} + \alpha \loss_{\textnormal{prompt 2}}
                    \end{equation}

                    \noindent
                    Instead, to interpolate in the loss, we sample the loss for each prompt with probability $\alpha$, which we equate to training with embedding:
                    \begin{equation}
                        \textToken' = (1 - Z) \textToken_1 + Z \textToken_2 \textnormal{ where } Z \sim \textnormal{Bern}(\alpha)
                    \end{equation}

                    \noindent
                    A third strategy, suggested for images in Magic3D~\citep{lin2022magic3d}, interpolates the DDM's guidance weight.
                    Unlike Magic3D, we amortize over guidance weights, reducing cost while providing continuous interpolation (not allowed via re-training on each weight).
                    Specifically, we guide with:
                    \begin{equation}
                        \hat{\epsilon} = \epsilon_{\textnormal{uncond.}} + (1 - \alpha) \omega_{1} \epsilon_{\textnormal{prompt 1}} + \alpha \omega_{2} \epsilon_{\textnormal{prompt 2}}
                    \end{equation}
                    Here, the $\omega_1$ and $\omega_2$ are notations for the guidance weights for the predicted noise on the $1^{st}$ and $2^{nd}$ prompts respectively, which are fixed and equal in all experiments.
                    This interpolates between using guidance on the first prompt, to guidance on the second prompt.
            
        % \subsection{Hyperparameter Choices}
        %     \TODO{
        %         Do we want some explanation for values we did not leave at defaults?
        %         Commented out table in latex.
        %     }
            % \TODO{
            %     Cover everything from our implementation details, and maybe more in the config.
            %     Describe our specific hyperparameter choices.
            %     Two kinds: those we tune, and those we left fixed at defaults.
            %     For defaults, describe why -- ex., prior work.
            %     For tuned results, add a table with the range we looked at.
            % }
            % \begin{figure}%[H]
            %     \begin{center}
            %         \begin{tabular}{ c| c |c } 
            %             \hline
            %             Hyperparameter              &Value                  & Considered Range\\
            %             \hline
            %             Orientation Reg. Weight     &$\num{0.01}$           & $[10^{-3}, 10^{-1}]$\\
            %             Opacity Reg. Weight         &$\num{0.002}$          & $[10^{-3}, 10^{-1}]$\\
            %             Optimizer                   &Adam                   & \TODO{}\\
            %             Learning rate               &$\num{0.01}$           & \TODO{}\\
            %             $\beta_1 \approx $ Momentum &$\num{0}$              & \TODO{}\\
            %             $\beta_2$                   &$\num{.999}$           & Default\\
            %             \hline
            %         \end{tabular}
            %     \end{center}
            %     \caption{
            %         \Jon{
            %             List our hyperparameters, their value, and the values we tried.
            %         }
            %     }\label{tab:hyperparameters}
            % \end{figure}

        \newcommand{\hshiftTwo}{\hspace{0.075\textwidth}}
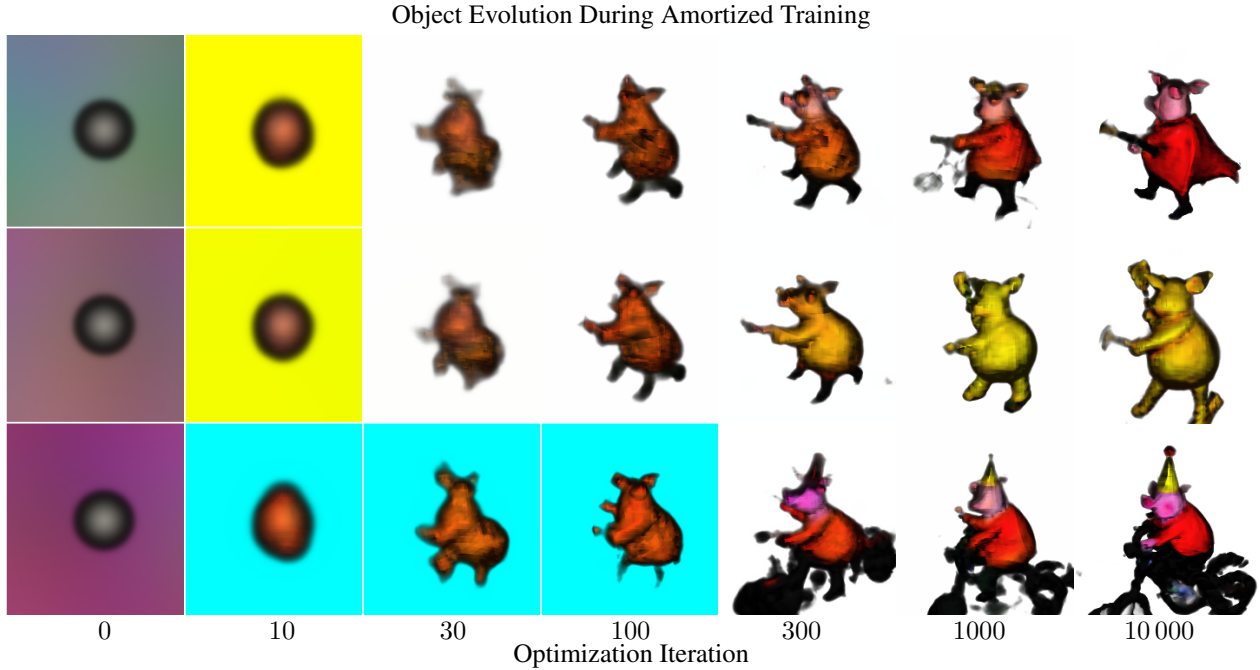
\begin{figure*}
    \vspace{-0.00\textheight}
    \centering
    \begin{tikzpicture}
        \centering
        % Jump down 7 for the first 5.  So 8*7 = 56.  And the differnce is 32, so 80-32 = 48
        \node (img21){\includegraphics[trim={.0cm 48.0cm .0cm 56.0cm}, clip, width=.15\linewidth]{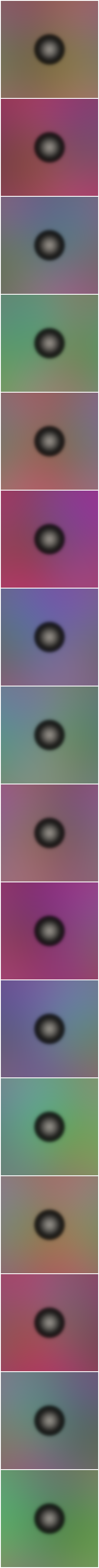}};
        \node [right=of img21, xshift=-1.5cm](img22){\includegraphics[trim={.0cm 48.0cm .0cm 56.0cm}, clip, width=.15\linewidth]{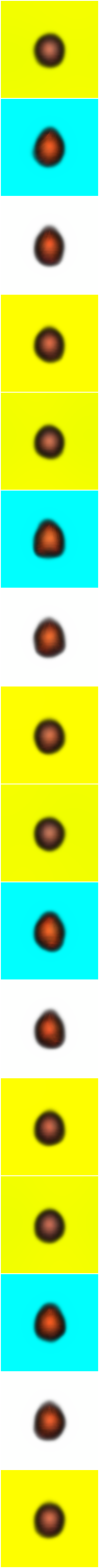}};
        \node [right=of img22, xshift=-1.5cm](img23){\includegraphics[trim={.0cm 48.0cm .0cm 56.0cm}, clip, width=.15\linewidth]{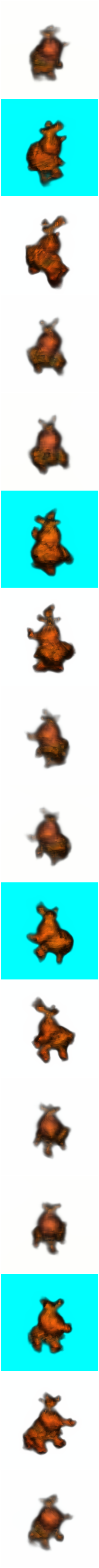}};
        \node [right=of img23, xshift=-1.5cm](img24){\includegraphics[trim={.0cm 48.0cm .0cm 56.0cm}, clip, width=.15\linewidth]{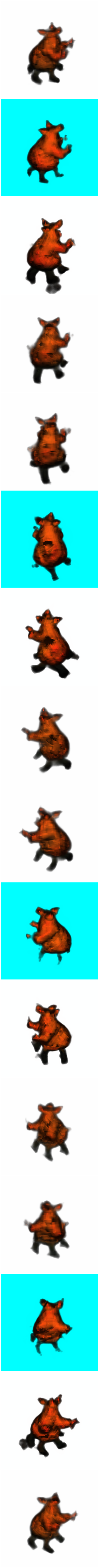}};
        \node [right=of img24, xshift=-1.5cm](img25){\includegraphics[trim={.0cm 48.0cm .0cm 56.0cm}, clip, width=.15\linewidth]{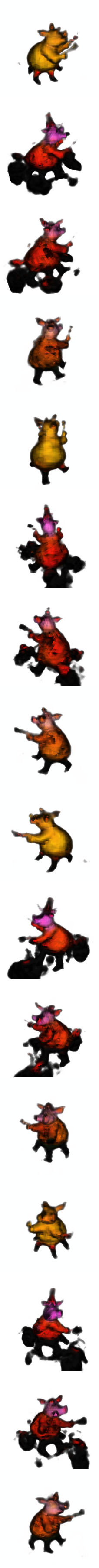}};
        \node [right=of img25, xshift=-1.5cm](img26){\includegraphics[trim={.0cm 80.0cm .0cm 24.0cm}, clip, width=.15\linewidth]{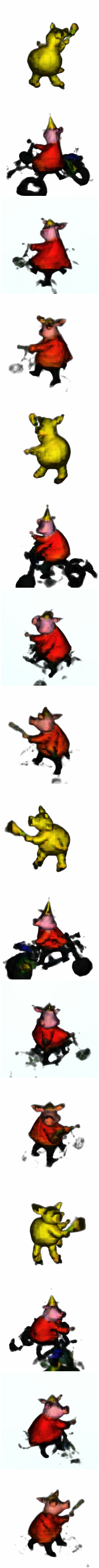}};
        \node [right=of img26, xshift=-1.5cm](img27){\includegraphics[trim={.0cm 80.0cm .0cm 24.0cm}, clip, width=.15\linewidth]{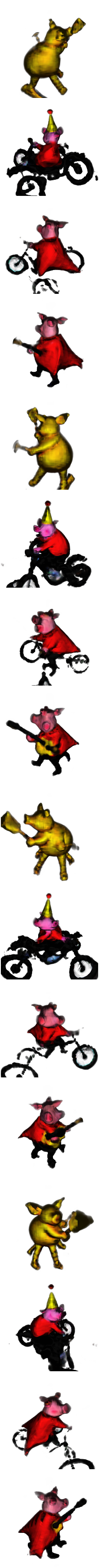}};
        
        % The x-axis of iterations
        \node[below=of img24, node distance=0cm, xshift=.1cm, yshift=1.15cm,font=\color{black}]{\num{0}\hspace{0.04\textwidth} \hshiftTwo \num{10}\hspace{0.03\textwidth} \hshiftTwo \num{30}\hspace{0.03\textwidth} \hshiftTwo \num{100}\hspace{0.02\textwidth} \hshiftTwo \num{300}\hspace{0.02\textwidth} \hshiftTwo \num{1000}\hspace{0.01\textwidth} \hshiftTwo \num{10000}};
        \node[below=of img24, node distance=0cm, xshift=-.1cm, yshift=.85cm,font=\color{black}]{Optimization Iteration};
        
        % The title
        %\node[above=of img24, node distance=0cm, xshift=-.1cm, yshift=-1.15cm,font=\color{black}]{\footnotesize{Training text-to-image samples use embedding $(1 - \alpha) \textToken_1 + \alpha \textToken_2$ where $\alpha \sim \mathcal{U}(0, 1)$}};
        \node[above=of img24, node distance=0cm, xshift=-.1cm, yshift=-1.15cm,font=\color{black}]{Object Evolution During Amortized Training};
    \end{tikzpicture}
    % \vspace{-0.015\textheight}
    \caption{
        We show assorted training trajectories of the rendered objects during compositional training from Figures~\ref{fig:all_quantitative} and \ref{fig:compositional_amortization}.
        \emph{Left:}
            We visualize the initialization strategy described in Section~\ref{sec:app-objective}.
    }
    \vspace{-0.0\textheight}
    \label{fig:pig_training}
\end{figure*}
        \subsection{Compute Requirements}\label{sec:app-compute-requirements}
            %We train our per-prompt models using a single 48GB GPU while the amortized models are largely trained using larger batch sizes with 8 GPUs.
            %The interpolation and finetuning experiments are performed on a single GPU.
            \noindent
            We implement our experiments in PyTorch~\citep{paszke2017automatic}.
            
            \subsubsection{Per-prompt Optimization}
                We do all per-prompt training runs on an NVIDIA A40 GPU, with a batch size of $\num{32}$ for up to $\num{8000}$ steps or $\sim \! \num{4}$ hours.
                DF$27$ (Figure~\ref{fig:all_quantitative}, left) use $\num{27}$ runs, while the compositional prompts (Figure~\ref{fig:animals-compositional}) use $\num{50}$ or $\num{300}$ subsampled runs respectively.
                Each training step costs $\sim \! \num{1}$ second.
            
            \subsubsection{Amortized Training}
                \textbf{Memorization \& generalization:}
                    When amortizing many prompts, we use multiple GPUs to train with a larger batch size, causing amortized and per-prompt training to have different update costs.
                    So we report the total rendered frames to compare compute accurately.
                    Updates are roughly $\num{1}$ second in each setup.
                    
\begin{figure*}%[h]
            \vspace{-0.00\textheight}
            %\vspace{-6mm}
            \centering
            \begin{tikzpicture}
                \centering
                % \node (img1){\includegraphics[trim={.5cm .5cm .5cm .5cm},clip,width=.99\linewidth]{images/df27_ft256.png}};
                \node (img1){\includegraphics[trim={.5cm .5cm .5cm .5cm},clip,width=.99\linewidth]{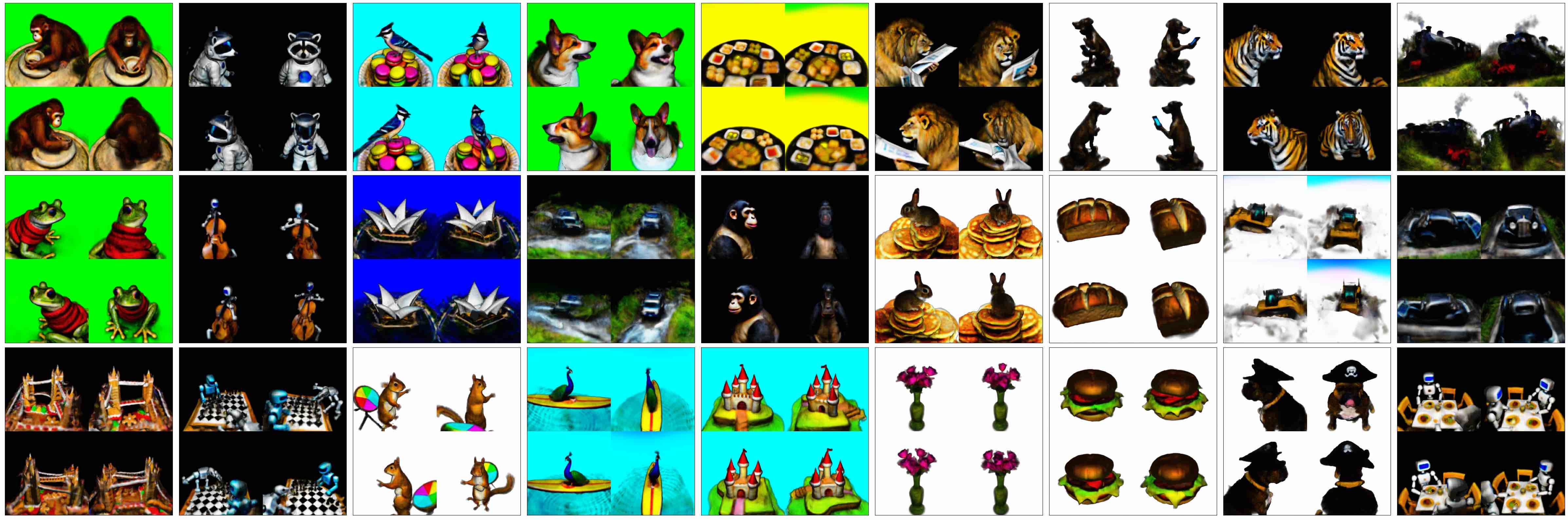}};
                %\node[left=of img1, node distance=0cm, rotate=90, xshift=2.0cm, yshift=-.5cm, font=\color{black}] {y-axis};
                %\node[below=of img1, node distance=0cm, xshift=-.1cm, yshift=.5cm,font=\color{black}]{x-axis};
            \end{tikzpicture}
            %\vspace{-0.0\textheight}
            %\caption{
            \vspace{-0.02\textheight}
            \captionof{figure}{
            % \caption{
                Our method, ATT3D, uses a single model to produce 3D scenes with varying geometric and texture details from the set of $27$ prompts in the main DreamFusion paper~\citep{poole2022dreamfusion}.
                The quality is comparable to existing single prompt training and requires far fewer training resources (Fig.~\ref{fig:all_quantitative}).
                %We generalize to unseen prompts with no additional training (Fig.~\ref{fig:compositional_amortization}) and have useful interpolations between 3D scenes (Fig.~\ref{fig:interpolation}).
                %\Kevin{Teaser picture with all memorized DF27 prompts}
            }\label{fig:memorize-df27}
            \vspace{-0.02\textheight}
        \end{figure*}

                    We perform the DF27 (Figures~\ref{fig:all_quantitative}, \ref{fig:memorize-df27}) and DF$411$ (Figures~\ref{fig:df400_qualitative}, \ref{fig:df400_full}) runs on $8$ NVIDIA A40 GPUs, each with a batch size of $32$.
                    We train DF$27$ for $\num{13000}$ steps ($\sim 4$ hours) and DF$411$ for $\num{100000}$ steps (about a day).
                    
                    The compositional runs (Figures~\ref{fig:compositional_amortization}, \ref{fig:all_quantitative}, \ref{fig:animals-compositional}) were performed on $4$ NVIDIA A100 GPUs, with a batch size of $\num{32}$ per GPU, for $\num{40000}$ steps or about $10$ hours.

                \noindent
                \textbf{Interpolations (Figure~\ref{fig:interpolation}):}
                    We use a single NVIDIA A40 GPU as in per-prompt training.
                
                \noindent
                \textbf{Finetuning (Figure~\ref{fig:finetuning_qualitative}):}
                    We use a single NVIDIA A40 GPU as in per-prompt training.
                    
            \subsubsection{Inference}
                At inference -- delineated from training in Figure~\ref{fig:amortized-text-to-3d-pipeline} -- we generate grid parameters in $< \num{1}$ second and render frames in real-time due to our small final NeRF $\nerfNet$ and efficient point-encoding $\pointEncoder$.
                We use more ray samples at inference than training due to negligible cost and enhanced fidelity.
                Modulation generation occurs once and is reused for each view \& location query, creating negligible overhead with many views or high-resolution images.
                During training with $1$ view and image size $64$ (batch size $8$), hypernet modulations introduced an overhead of $24\%$ more time per iteration, which could be avoided if our weights do not need to be generated.
                With $1$ view and image size $256$ (batch size $1$) in inference, the modulation introduced an $11\%$ overhead in rendering time, dropping to $< 1\%$ with $30$ views.

\section{Results}\label{sec:app-results}
    \subsection{Additional Experiments \& Visualizations}\label{sec:app-visualizations}

        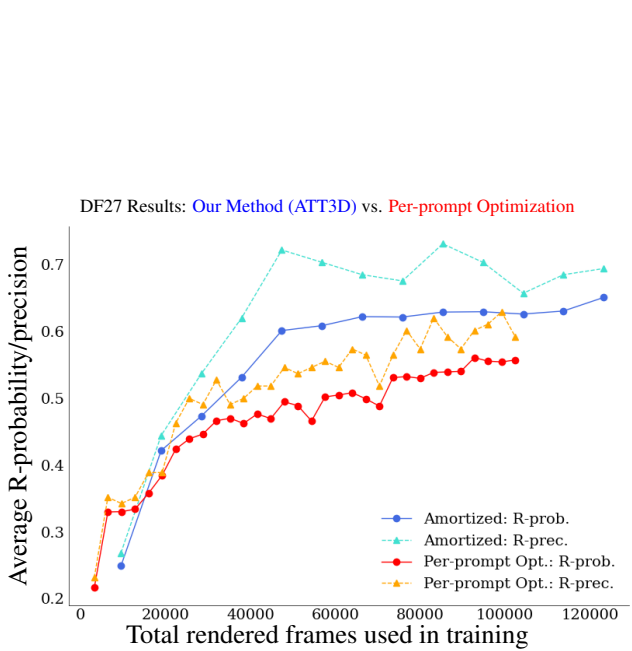
\begin{figure}
            \vspace{-0.02\textheight}
            \centering
            \hspace{-0.02\textwidth}
            \begin{tikzpicture}
                \centering
                \node (img11){\includegraphics[trim={.0cm .75cm .0cm .75cm},clip,width=.95\linewidth]{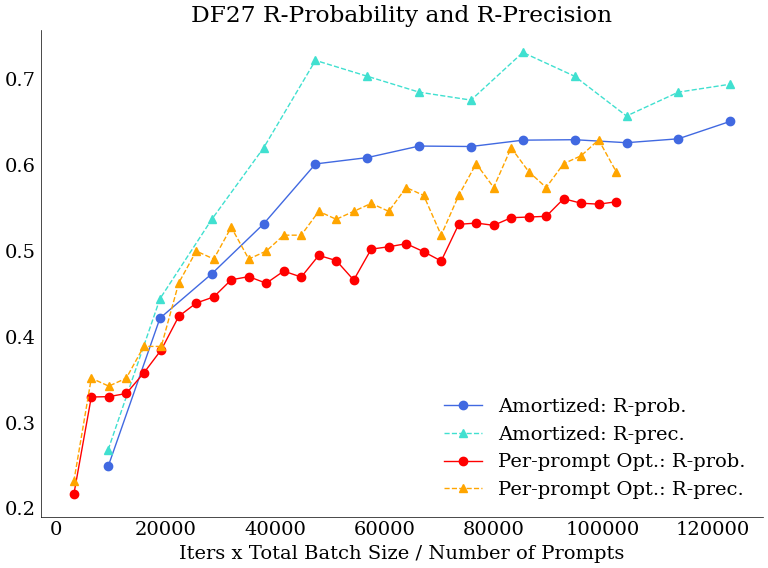}};
                
                \node[left=of img11, node distance=0cm, rotate=90, xshift=2.5cm, yshift=-.9cm, font=\color{black}] {Average R-probability/precision};
                \node[below=of img11, node distance=0cm, xshift=-.1cm, yshift=1.2cm,font=\color{black}]{Total rendered frames used in training};
                %{Iterations $\cdot$ \# GPUs vs. Iterations $\cdot$ \# Prompts};
                \node[above=of img11, node distance=0cm, xshift=-.1cm, yshift=-1.1cm,font=\color{black}]{\scriptsize{DF27 Results: {\color{blue}Our Method (ATT3D)} vs. {\color{red}Per-prompt Optimization}}};
            \end{tikzpicture}
            \vspace{-0.01\textheight}
            \caption{
                We show the same plot as Figure~\ref{fig:all_quantitative} with the addition of R-precision.
                \textbf{Takeaway:} Results with R-precision are similar to –- but noisier than -- R-probability when we have few prompts.
            }
            \vspace{-0.02\textheight}
            \label{fig:r-prec-vs-r-prob}
        \end{figure}

        \begin{figure}%[h!]
            \vspace{-0.02\textheight}
            \centering
            \begin{tikzpicture}
                \centering
                \node (img1){\includegraphics[trim={.0cm 15.0cm .0cm .85cm},clip,width=.75\linewidth]{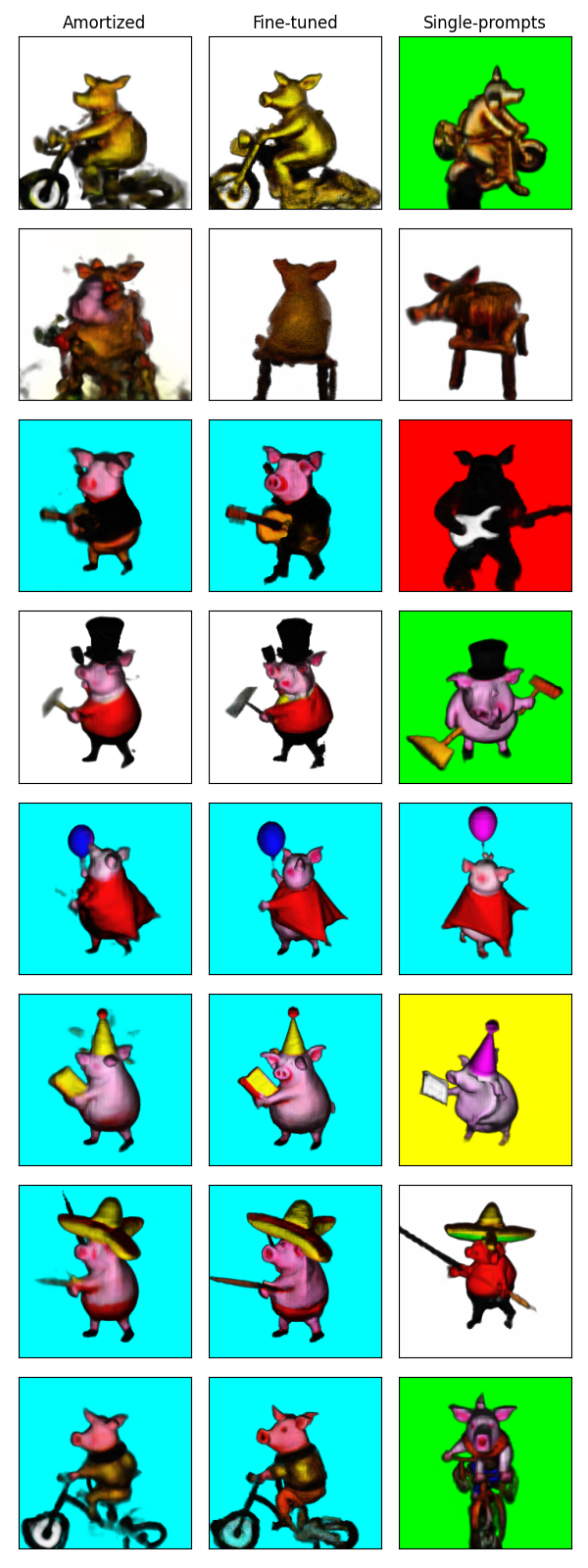}};
                \node[above=of img1, node distance=0cm, xshift=.05cm, yshift=-1.2cm,font=\color{black}]{\scriptsize{{\color{blue}Amortized}} \hspace{0.0225\textwidth} \scriptsize{{\color{blue}+ Finetuning}} \hspace{0.0225\textwidth} \scriptsize{{\color{red}Per-prompt}}};
                \node[above=of img1, node distance=0cm, xshift=-.1cm, yshift=-.75cm,font=\color{black}]{Training Style};
            \end{tikzpicture}
            \vspace{-0.02\textheight}
            \caption{
                We qualitatively compare the unseen ``testing" results from the various training strategies in Figures~\ref{fig:all_quantitative} and \ref{fig:compositional_amortization}, with {\color{blue}our method in blue} and {\color{red}baselines in red}.
                Notably, {\color{blue}amortized} requires no test time optimization, while {\color{blue}finetuning} uses a small amount, and {\color{red}per-prompt} uses a large amount to tune from scratch.
            }
            \vspace{-0.02\textheight}
            \label{fig:final_pig_comparison}
        \end{figure}
        
        % Large prompt set grid visualizations 
        \begin{figure*}%[h!]
            %\vspace{-0.0\textheight}
            \centering
            %\hspace{-0.06\textwidth}
            \begin{tikzpicture}
                \centering
                \node (img1){\includegraphics[trim={.0cm .0cm .0cm .0cm},clip,width=1.0\linewidth]{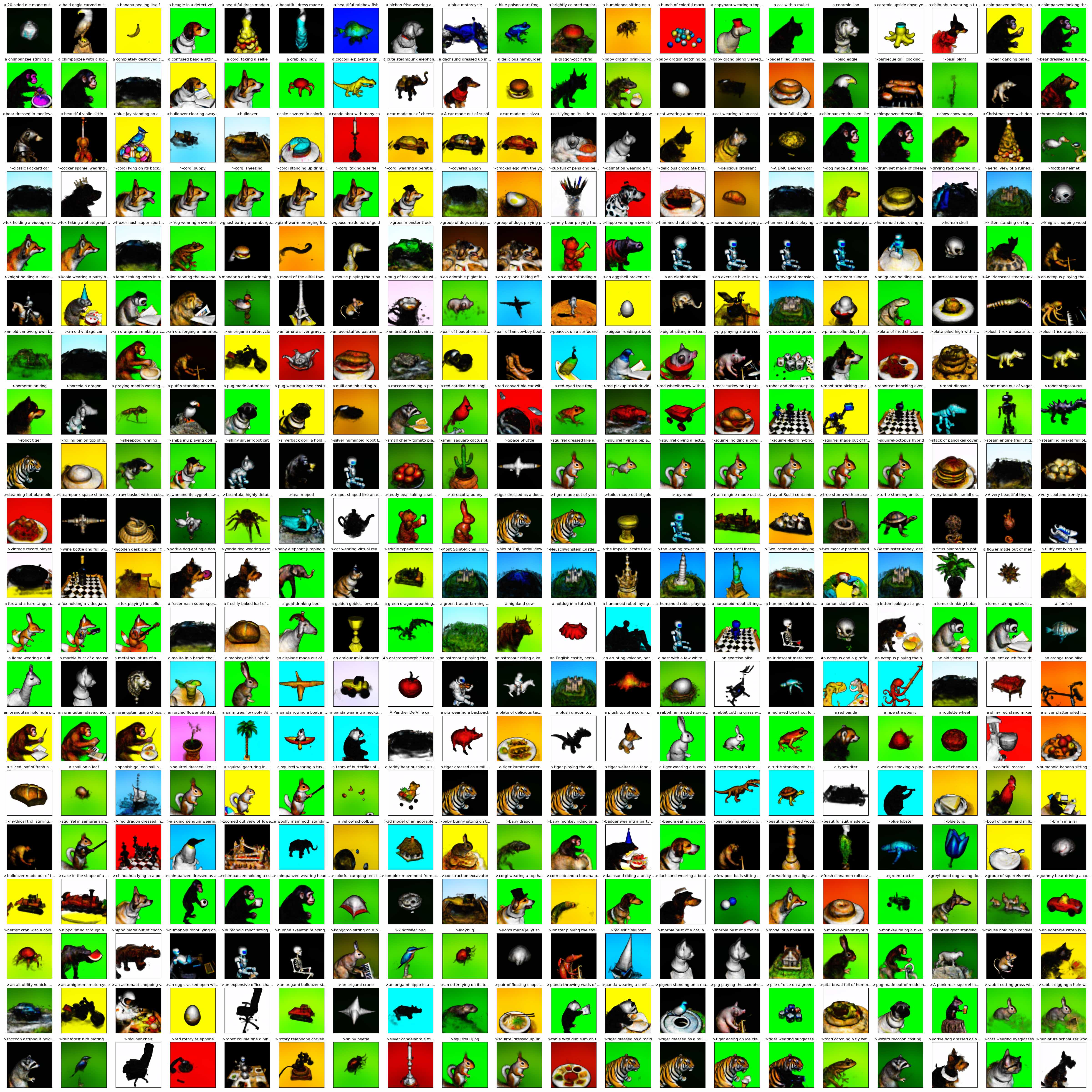}};
                %\node (img1){\includegraphics[trim={.0cm .0cm .0cm .0cm},clip,width=1.0\linewidth]{images/additional/df400_grid.jpg}};
                % CHANGED THIS LINE TO SAVE SPACE
            \end{tikzpicture}
            %\vspace{-0.0\textheight}
            \caption{
                We show full results from our method on the DF$411$ prompt set, which we truncate for Figure~\ref{fig:df400_qualitative}.
                There are various examples of the model re-using object components across prompts -- see Figure~\ref{fig:feature-reuse}.
            }
            %\vspace{-0.0\textheight}
            \label{fig:df400_full}
        \end{figure*}

        \begin{figure*}
            %\vspace{-0.0\textheight}
            \centering
            \begin{tikzpicture}
                \centering
                \node (img11){\includegraphics[trim={.0cm .0cm .0cm .0cm},clip,width=.22\linewidth]{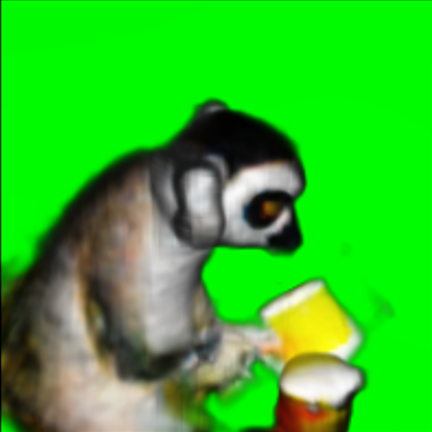}};
                \node [right=of img11, xshift=-1cm](img12){\includegraphics[trim={.275cm .275cm .275cm .25cm},clip,width=.22\linewidth]{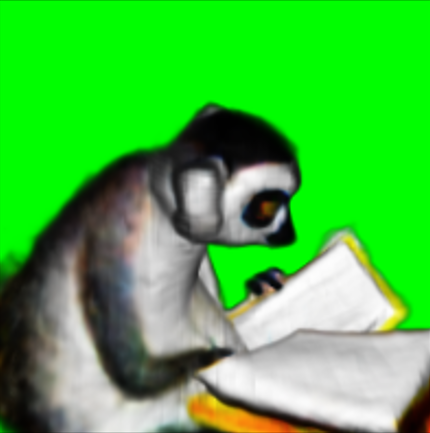}};
                \node[left=of img11, node distance=0cm, rotate=90, xshift=1.5cm, yshift=-.9cm, font=\color{black}] {\scriptsize{``\emph{a lemur drinking boba}"}};
                \node[right=of img12, node distance=0cm, rotate=270, xshift=-1.4cm, yshift=-.9cm, font=\color{black}] {\scriptsize{``\emph{a lemur taking notes}"}};
                
                \node[above=of img12, node distance=0cm, xshift=-1.75cm, yshift=-1.1cm,font=\color{black}]{Component Re-use in DF$411$};
                
                \node [below=of img11, yshift=1cm](img21){\includegraphics[trim={.2cm .2cm .2cm .2cm},clip,width=.22\linewidth]{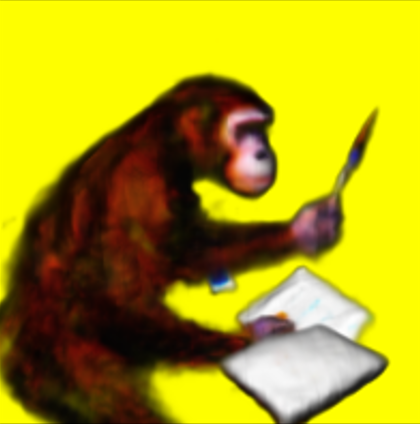}};
                \node [right=of img21, xshift=-1cm](img22){\includegraphics[trim={.275cm .275cm .275cm .275cm},clip,width=.22\linewidth]{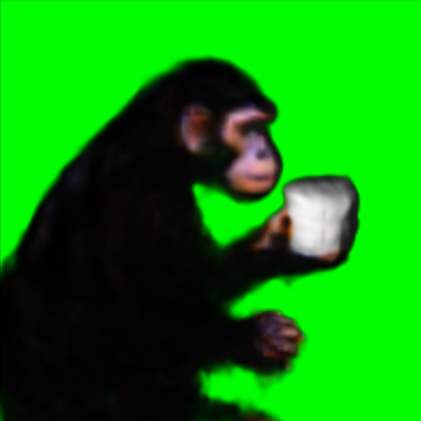}};
                \node[left=of img21, node distance=0cm, rotate=90, xshift=1.8cm, yshift=-.9cm, font=\color{black}] {\scriptsize{``\emph{orangutan holding a paintbrush}"}};
                \node[right=of img22, node distance=0cm, rotate=270, xshift=-1.5cm, yshift=-.9cm, font=\color{black}] {\scriptsize{``\emph{chimpanzee holding a cup}"}};
                
                % \node [below=of img21, yshift=1cm](img31){\includegraphics[trim={.2cm .2cm .2cm .2cm},clip,width=.45\linewidth]{example-image-a}};
                % \node [right=of img31, xshift=-1cm](img32){\includegraphics[trim={.25cm .25cm .25cm .25cm},clip,width=.45\linewidth]{example-image-a}};
                % \node [right=of img22, xshift=-1cm](img23){\includegraphics[trim={.0cm .0cm .0cm .0cm},clip,width=.3\linewidth]{example-image-a}};
            \end{tikzpicture}
            %\vspace{-0.0\textheight}
            \caption{
                We show examples of prompts in which our model (from the DF$411$ run in Figures~\ref{fig:df400_qualitative} and \ref{fig:df400_full}) re-uses components, showing a means by which amortization saves compute.
                \emph{Top:}
                    The lemur is re-used with different activities.
                \emph{Bottom:}
                    The orangutan is re-colored to a chimpanzee and given a different activity.
            }
            %\vspace{-0.0\textheight}
            \label{fig:feature-reuse}
        \end{figure*}
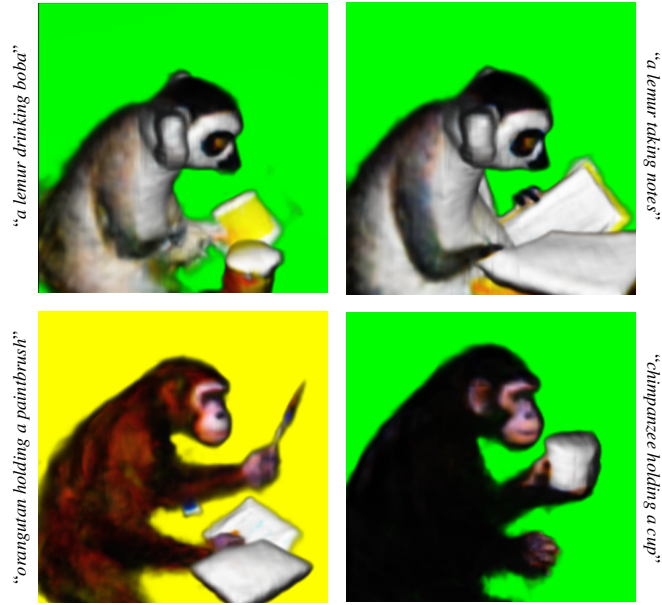

        \begin{figure*}
            \vspace{-0.02\textheight}
            \centering
            \begin{tikzpicture}
                \centering
                \node (img21){\includegraphics[trim={.0cm .0cm .0cm .0cm}, clip, width=.14\linewidth]{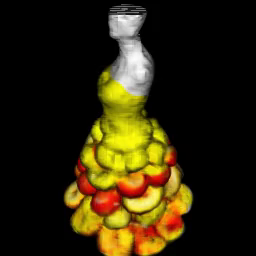}};
                \node [right=of img21, xshift=-1.4cm](img22){\includegraphics[trim={.0cm .0cm .0cm .0cm}, clip, width=.14\linewidth]{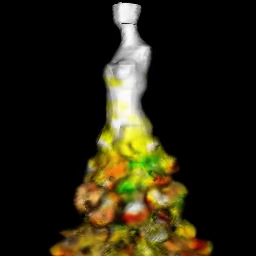}};
                \node [right=of img22, xshift=-1.4cm](img23){\includegraphics[trim={.0cm .0cm .0cm .0cm}, clip, width=.14\linewidth]{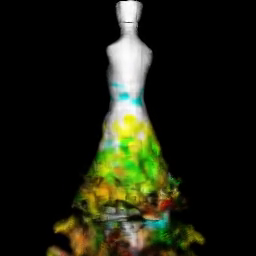}};
                \node [right=of img23, xshift=-1.4cm](img24){\includegraphics[trim={.0cm .0cm .0cm .0cm}, clip, width=.14\linewidth]{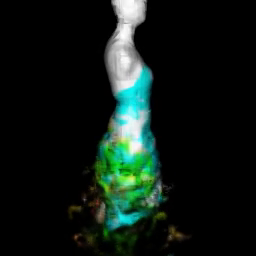}};
                \node [right=of img24, xshift=-1.4cm](img25){\includegraphics[trim={.0cm .0cm .0cm .0cm}, clip, width=.14\linewidth]{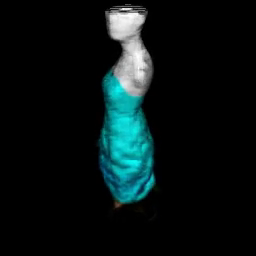}};
                \node[above=of img21, node distance=0cm, xshift=.0cm, yshift=-1.15cm,font=\color{black}]{``\emph{...dress of fruit...}"};
                \node[above=of img25, node distance=0cm, xshift=.0cm, yshift=-1.15cm,font=\color{black}]{``\emph{...dress of bags...}"};

                \node[above=of img23, node distance=0cm, xshift=.0cm, yshift=-.75cm,font=\color{black}]{Interpolated embeddings not viewed during training};
                \node[above=of img23, node distance=0cm, xshift=.0cm, yshift=-1.15cm,font=\color{black}]{I.e., interpolants have no training};
               
                % \node[left=of img21, node distance=0cm, xshift=.9cm, yshift=-.3cm, rotate=90, font=\color{black}]{No Training};
                %\node[left=of img21, node distance=0cm, xshift=1.0cm, yshift=.5cm, rotate=90, font=\color{black}]{\tiny{{\color{green}Success}}};

                \node [below=of img21, yshift=1.25cm](img31){\includegraphics[trim={.0cm .0cm .0cm .0cm}, clip, width=.14\linewidth]{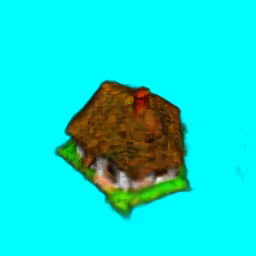}};
                \node [right=of img31, xshift=-1.4cm](img32){\includegraphics[trim={.0cm .0cm .0cm .0cm}, clip, width=.14\linewidth]{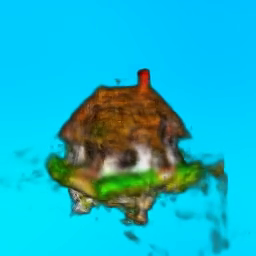}};
                \node [right=of img32, xshift=-1.4cm](img33){\includegraphics[trim={.0cm .0cm .0cm .0cm}, clip, width=.14\linewidth]{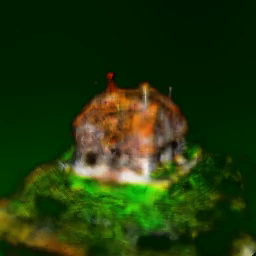}};
                \node [right=of img33, xshift=-1.4cm](img34){\includegraphics[trim={.0cm .0cm .0cm .0cm}, clip, width=.14\linewidth]{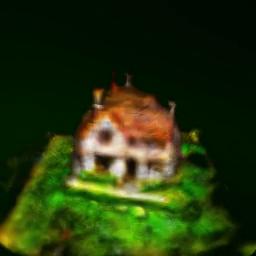}};
                \node [right=of img34, xshift=-1.4cm](img35){\includegraphics[trim={.0cm .0cm .0cm .0cm}, clip, width=.14\linewidth]{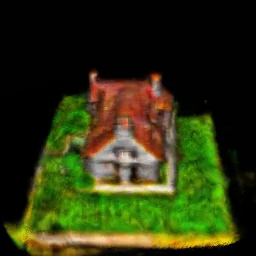}};
                %\node[left=of img31, node distance=0cm, xshift=.9cm, yshift=.5cm, rotate=90, font=\color{black}]{\footnotesize{Zero-shot}};
                \node[below=of img31, node distance=0cm, xshift=.0cm, yshift=1.15cm,font=\color{black}]{``\emph{...cottage...}"};
                \node[below=of img35, node distance=0cm, xshift=.0cm, yshift=1.15cm,font=\color{black}]{``\emph{...house...}"};
                %\node[left=of img31, node distance=0cm, xshift=1.0cm, yshift=.5cm, rotate=90, font=\color{black}]{\tiny{{\color{green}Success}}};

                \node [right=of img25, xshift=-.9cm, yshift=-.0cm](img81){\includegraphics[trim={0.0cm 0.0cm 0.0cm 0.0cm},clip,width=.12\linewidth]{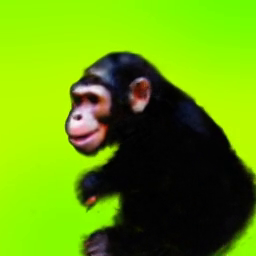}};
                \node[above=of img81, node distance=0cm, xshift=-.1cm, yshift=-1.25cm,font=\color{black}]{\footnotesize{``\emph{a chimpanzee}"}};
                \node [right=of img81, xshift=-1.0cm, yshift=-.0cm](img82){\includegraphics[trim={0.0cm 0.0cm 0.0cm 0.0cm},clip,width=.12\linewidth]{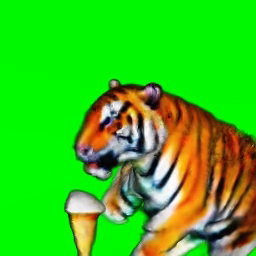}};
                \node[above=of img82, node distance=0cm, xshift=.1cm, yshift=-1.25cm,font=\color{black}]{\footnotesize{``\emph{...eating an icecream}"}};
                \node[right=of img82, node distance=0cm, xshift=-.85cm, yshift=.65cm, rotate=270, font=\color{black}]{Training};
                
                \node [right=of img35, xshift=.15cm, yshift=-.1cm](img91){\includegraphics[trim={0.0cm 0.0cm 0.0cm 0.0cm},clip,width=.12\linewidth]{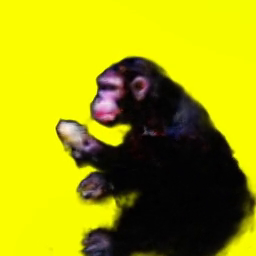}};
                \node[above=of img91, node distance=0cm, xshift=.25cm, yshift=-1.25cm,font=\color{black}]{\footnotesize{``\emph{a chimpanzee}" + ``\emph{eating an icecream}"}};
                \node[right=of img91, node distance=0cm, xshift=-.85cm, yshift=1.0cm, rotate=270, font=\color{black}]{No Training};
            \end{tikzpicture}
            \vspace{-0.02\textheight}
            \caption{
                We investigate generalization on the DF411 run (App. Fig.~\ref{fig:df400_full}).
                \emph{Left}:
                    Generalization to interpolated embeddings, which produces suboptimal results that we improve by amortizing over interpolants as in Figure~\ref{fig:interpolation}.
                \emph{Right}:
                    Generalization to compositional embeddings.
                \textbf{Takeaway}: The generalization is promising, yet could be improved, motivating training on large compositional sets in Figures~\ref{fig:all_quantitative} \& \ref{fig:animals-compositional}, and training on interpolants as in Figures~\ref{fig:interpolation}, \ref{fig:interpolation_hamburgers}, \ref{fig:interpolation_ships}, \& \ref{fig:interpolation_other}.
            }
            \vspace{-0.02\textheight}
            \label{fig:df411-zero-shot}
        \end{figure*}
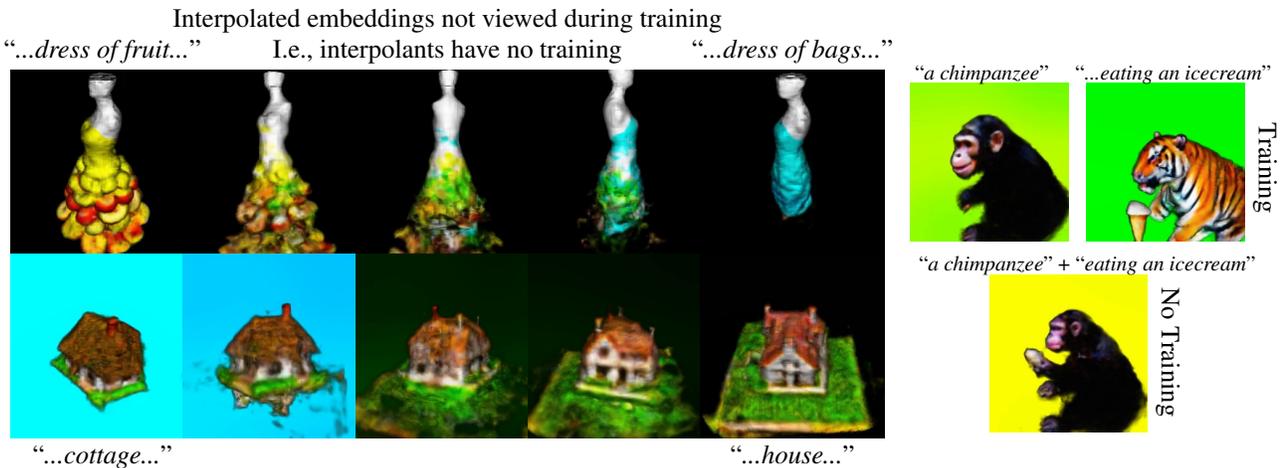

        \newcommand{\hshift}{\hspace{0.12\textwidth}}
        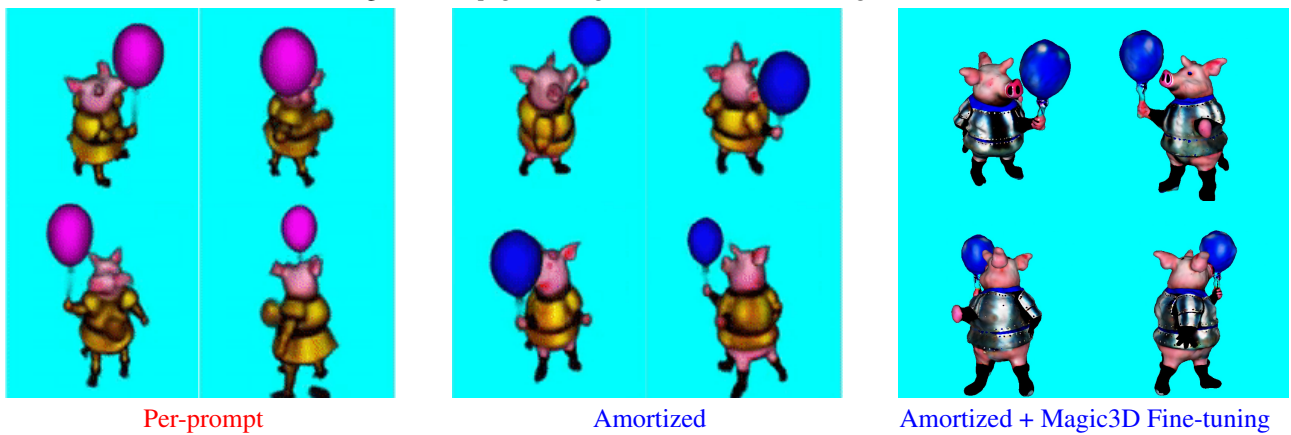
\begin{figure*}[ht!]
            \vspace{-0.00\textheight}
            \centering
            \begin{tikzpicture}
                \centering
                \node (img1){\includegraphics[trim={.0cm .0cm .0cm .75cm},clip,width=.95\linewidth]{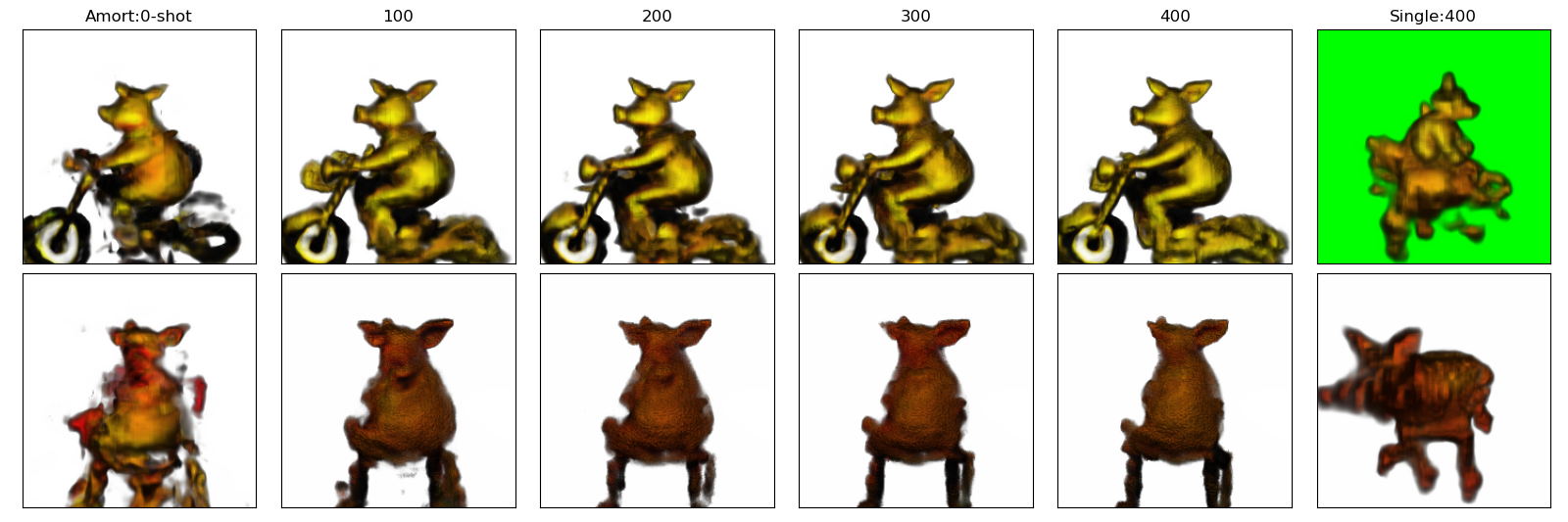}};
                
                \node[above=of img1, node distance=0cm, xshift=-.1cm, yshift=-.8cm,font=\color{black}]{Finetuning iteration};
                \node[above=of img1, node distance=0cm, xshift=.1cm, yshift=-1.15cm,font=\color{black}]{0\hspace{0.01\textwidth} \hshift 100 \hshift 200 \hshift 300 \hshift \textbf{400} \hshift \textbf{400}};
                
                \node[left=of img1, node distance=0cm, rotate=90, xshift=2.7cm, yshift=-1.1cm, font=\color{black}] {\scriptsize{``\emph{... motorcycle of gold"}}};
                \node[left=of img1, node distance=0cm, rotate=90, xshift=0.1cm, yshift=-1.1cm, font=\color{black}] {\scriptsize{``\emph{... chair carved of wood"}}};
                \node[left=of img1, node distance=0cm, rotate=90, xshift=1.15cm, yshift=-.65cm, font=\color{black}] {Text prompt};
                
                % \node[below=of img1, node distance=0cm, xshift=.0cm, yshift=.9cm,font=\color{black}]{Training Method};
                \node[below=of img1, node distance=0cm, xshift=-2.0cm, yshift=1.3cm,font=\color{black}]{{\color{blue} Amortized (first 5 columns)}};
                \node[below=of img1, node distance=0cm, xshift=6.85cm, yshift=1.3cm,font=\color{black}]{{\color{red}Per-prompt}};
        
                \node [below=of img1, xshift=-7.75cm, yshift=.2cm] (img11){\includegraphics[trim={.0cm .0cm .0cm .0cm},clip,width=.15\linewidth]{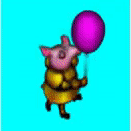}};
                \node [right=of img11, xshift=-1.3cm](img12){\includegraphics[trim={.0cm .0cm .0cm .0cm},clip,width=.15\linewidth]{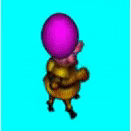}};
                \node [below=of img11, yshift=1.3cm](img21){\includegraphics[trim={.0cm .0cm .0cm .0cm},clip,width=.15\linewidth]{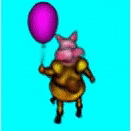}};
                \node [right=of img21, xshift=-1.3cm](img22){\includegraphics[trim={.0cm .0cm .0cm .0cm},clip,width=.15\linewidth]{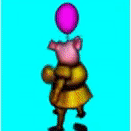}};
                \node[below=of img22, node distance=0cm, xshift=-1.25cm, yshift=1.1cm,font=\color{black}]{{\color{red}Per-prompt}};

                \node [right=of img12, xshift=-.5cm](img111){\includegraphics[trim={.0cm .0cm .0cm .0cm},clip,width=.15\linewidth]{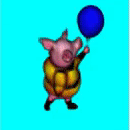}};
                \node [right=of img111, xshift=-1.3cm](img112){\includegraphics[trim={.0cm .0cm .0cm .0cm},clip,width=.15\linewidth]{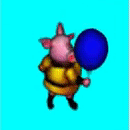}};
                \node [below=of img111, yshift=1.3cm](img121){\includegraphics[trim={.0cm .0cm .0cm .0cm},clip,width=.15\linewidth]{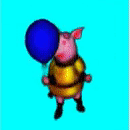}};
                \node [right=of img121, xshift=-1.3cm](img122){\includegraphics[trim={.0cm .0cm .0cm .0cm},clip,width=.15\linewidth]{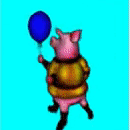}};
                \node[below=of img122, node distance=0cm, xshift=-1.25cm, yshift=1.1cm,font=\color{black}]{{\color{blue}Amortized}};

                \node [right=of img112, xshift=-.5cm](img211){\includegraphics[trim={.0cm .0cm .0cm .0cm},clip,width=.15\linewidth]{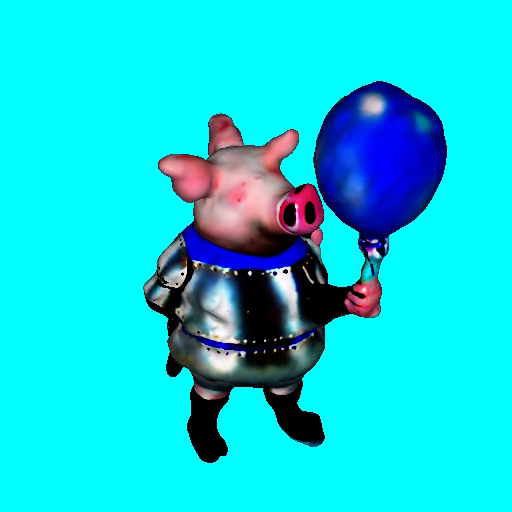}};
                \node [right=of img211, xshift=-1.3cm](img212){\includegraphics[trim={.0cm .0cm .0cm .0cm},clip,width=.15\linewidth]{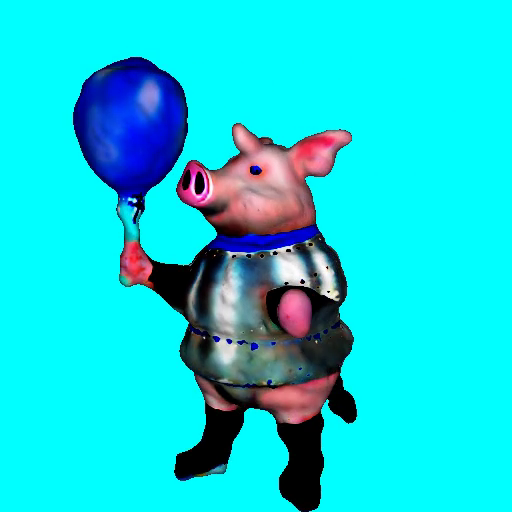}};
                \node [below=of img211, yshift=1.3cm](img221){\includegraphics[trim={.0cm .0cm .0cm .0cm},clip,width=.15\linewidth]{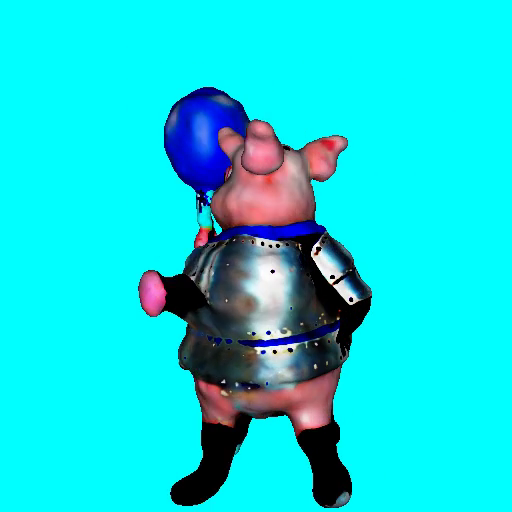}};
                \node [right=of img221, xshift=-1.3cm](img222){\includegraphics[trim={.0cm .0cm .0cm .0cm},clip,width=.15\linewidth]{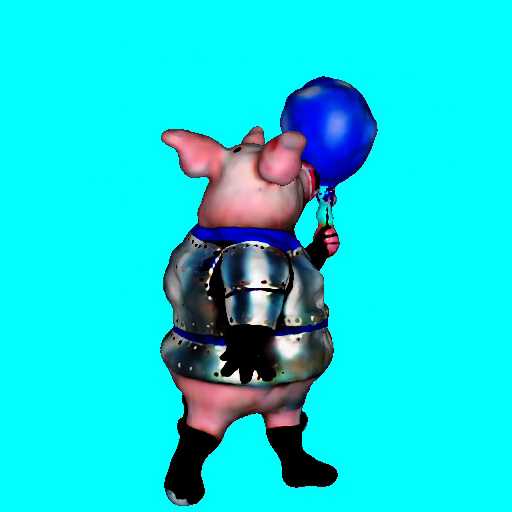}};
                \node[below=of img222, node distance=0cm, xshift=-1.4cm, yshift=1.1cm,font=\color{black}]{{\color{blue}Amortized + Magic3D Fine-tuning}};
        
                \node[above=of img112, node distance=0cm, xshift=-1.5cm, yshift=-1.15cm,font=\color{black}]{Various strategies on ``\emph{a pig wearing medieval armor holding a blue balloon}"};
            \end{tikzpicture}
            \vspace{-0.03\textheight}
            \caption{
                We display the results of finetuning held-out, unseen testing prompts from Fig.~\ref{fig:compositional_amortization}.
                \emph{Top:}
                    For amortization, we finetune from the final optimization value, while for per-prompt, we finetune the model from a random initialization.
                    We achieve higher quality with fewer finetuning updates.
                \emph{Bottom:}
                    Per-prompt optimization fails to recover a blue balloon, and can not be recovered with finetuning.
                    In contrast, amortized optimization recovers the correct balloon and can be fine-tuned using Magic3D's second optimization stage~\citep{lin2022magic3d}.
            }
            \vspace{-0.02\textheight}
            \label{fig:finetuning_qualitative}
        \end{figure*}
        % \begin{figure*}[h!]
        %     %\vspace{-0.0\textheight}
        %     \centering
        %     \begin{tikzpicture}
        %         \centering
        %         \node (img1){\includegraphics[trim={.0cm .0cm .0cm .0cm},clip,width=.90\linewidth]{images/additional/coco153_grid.png}};
        %     \end{tikzpicture}
        %     %\vspace{-0.0\textheight}
        %     \caption{
        %       \TODO{
        %             COCO 153 grid visualization
        %         }
        %     }
        %     %\vspace{-0.0\textheight}
        %     \label{fig:TODO}
        % \end{figure*}

        % \begin{figure}[h!]
        %     %\vspace{-0.0\textheight}
        %     \centering
        %     \begin{tikzpicture}
        %         \centering
        %         \node (img1){\includegraphics[trim={.0cm .0cm .0cm .0cm},clip,width=.90\linewidth]{example-image-a}};
        %     \end{tikzpicture}
        %     %\vspace{-0.0\textheight}
        %     \caption{
        %       \TODO{
        %             Other examples of failures of single prompt, and perhaps us fixing them.
        %         }
        %     }
        %     %\vspace{-0.0\textheight}
        %     \label{fig:TODO3}
        % \end{figure}

        % \begin{figure}[h!]
        %     %\vspace{-0.0\textheight}
        %     \centering
        %     \begin{tikzpicture}
        %         \centering
        %         \node (img1){\includegraphics[trim={.0cm .0cm .0cm .0cm},clip,width=.90\linewidth]{example-image-a}};
        %     \end{tikzpicture}
        %     %\vspace{-0.0\textheight}
        %     \caption{
        %       \TODO{
        %             Any other interpolations we thought were interesting
        %         }
        %     }
        %     %\vspace{-0.0\textheight}
        %     \label{fig:TODO4}
        % \end{figure}
            
        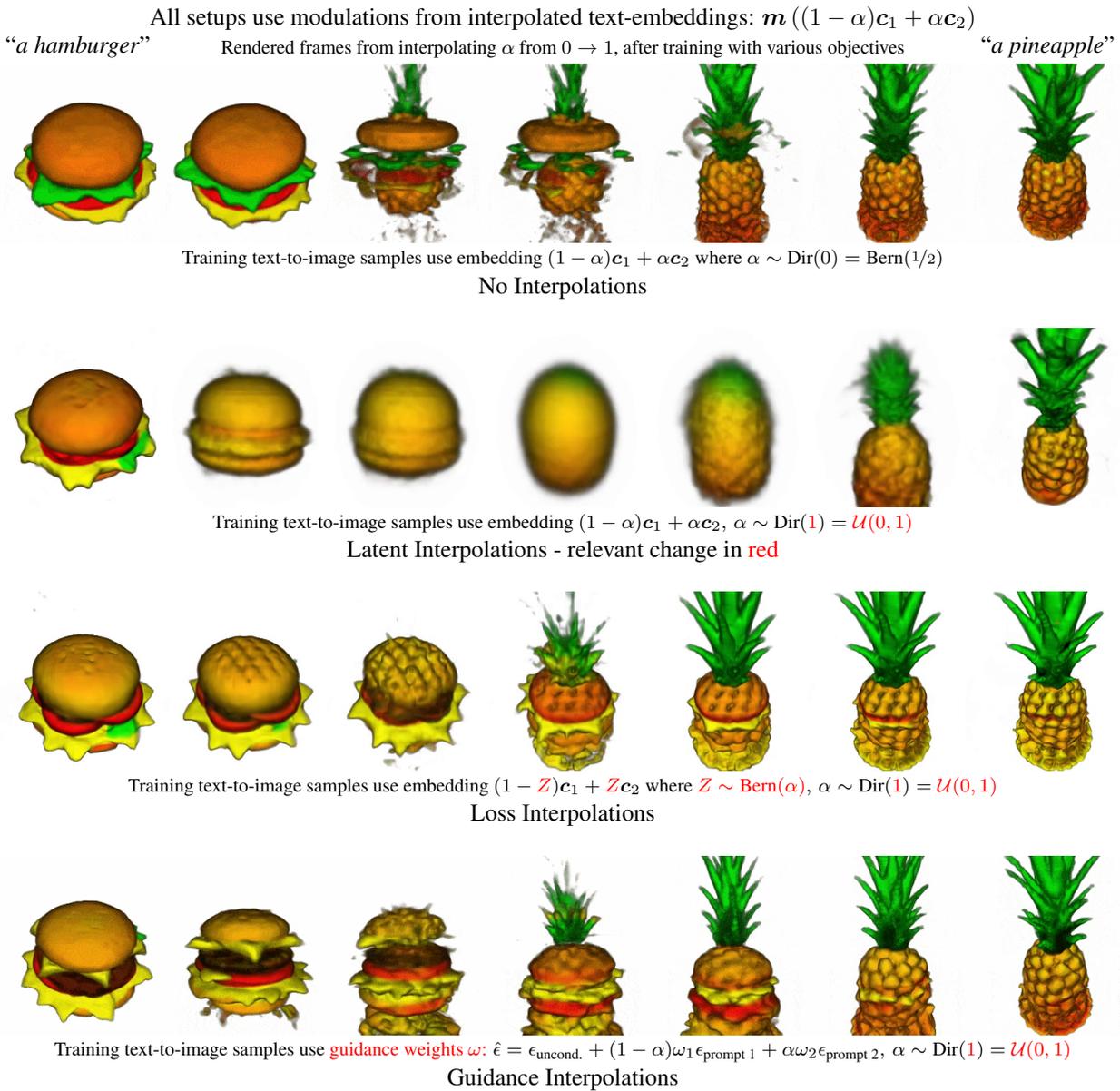
\begin{figure*}%[ht!]
            %\vspace{-0.0225\textheight}
            \centering
            \begin{tikzpicture}
                \centering
                \node (img41){\includegraphics[trim={.0cm .0cm .0cm .0cm}, clip, width=.15\linewidth]{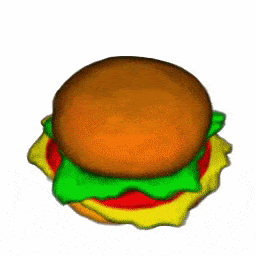}};
                \node [right=of img41, xshift=-1.5cm](img42){\includegraphics[trim={.0cm .0cm .0cm .0cm}, clip, width=.15\linewidth]{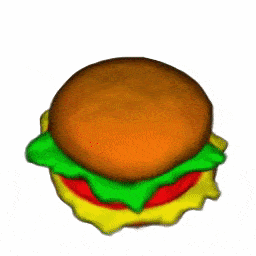}};
                \node [right=of img42, xshift=-1.5cm](img43){\includegraphics[trim={.0cm .0cm .0cm .0cm}, clip, width=.15\linewidth]{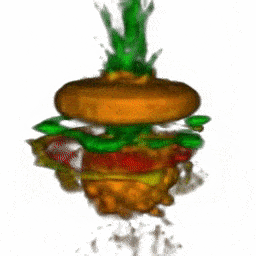}};
                \node [right=of img43, xshift=-1.5cm](img44){\includegraphics[trim={.0cm .0cm .0cm .0cm}, clip, width=.15\linewidth]{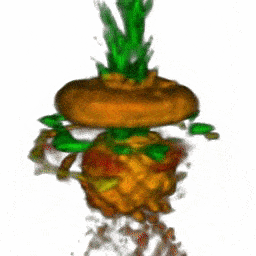}};
                \node [right=of img44, xshift=-1.5cm](img45){\includegraphics[trim={.0cm .0cm .0cm .0cm}, clip, width=.15\linewidth]{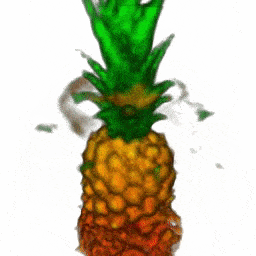}};
                \node [right=of img45, xshift=-1.5cm](img46){\includegraphics[trim={.0cm .0cm .0cm .0cm}, clip, width=.15\linewidth]{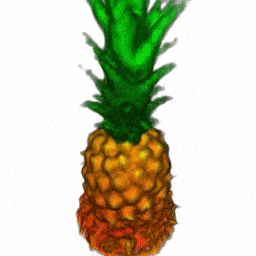}};
                \node [right=of img46, xshift=-1.5cm](img47){\includegraphics[trim={.0cm .0cm .0cm .0cm}, clip, width=.15\linewidth]{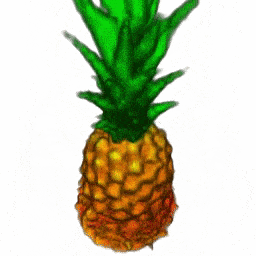}};
                \node[below=of img44, node distance=0cm, xshift=-.1cm, yshift=1.15cm,font=\color{black}]{\footnotesize{Training text-to-image samples use embedding $(1 - \alpha) \textToken_1 + \alpha \textToken_2$ where $\alpha \sim \textnormal{Dir}(0) = \textnormal{Bern}(\nicefrac{1}{2})$}};
                \node[below=of img44, node distance=0cm, xshift=-.1cm, yshift=.75cm,font=\color{black}]{No Interpolations};

                \node [below=of img41] (img21){\includegraphics[trim={.0cm .0cm .0cm .0cm}, clip, width=.15\linewidth]{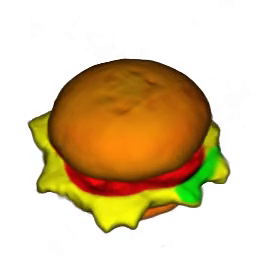}};
                \node [right=of img21, xshift=-1.5cm](img22){\includegraphics[trim={.0cm .0cm .0cm .0cm}, clip, width=.15\linewidth]{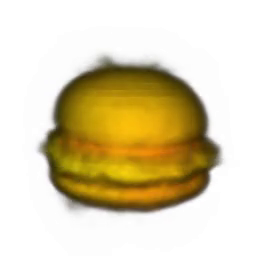}};
                \node [right=of img22, xshift=-1.5cm](img23){\includegraphics[trim={.0cm .0cm .0cm .0cm}, clip, width=.15\linewidth]{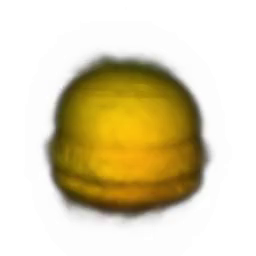}};
                \node [right=of img23, xshift=-1.5cm](img24){\includegraphics[trim={.0cm .0cm .0cm .0cm}, clip, width=.15\linewidth]{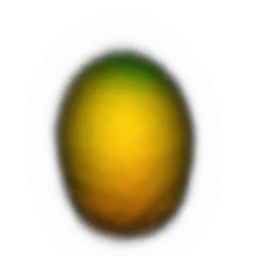}};
                \node [right=of img24, xshift=-1.5cm](img25){\includegraphics[trim={.0cm .0cm .0cm .0cm}, clip, width=.15\linewidth]{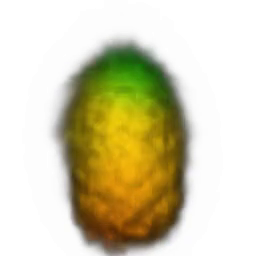}};
                \node [right=of img25, xshift=-1.5cm](img26){\includegraphics[trim={.0cm .0cm .0cm .0cm}, clip, width=.15\linewidth]{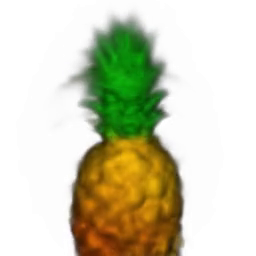}};
                \node [right=of img26, xshift=-1.5cm](img27){\includegraphics[trim={.0cm .0cm .0cm .0cm}, clip, width=.15\linewidth]{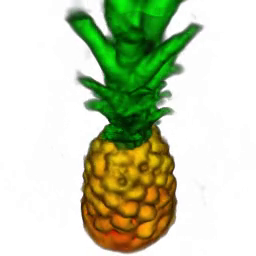}};
                \node[below=of img24, node distance=0cm, xshift=-.1cm, yshift=1.15cm,font=\color{black}]{\footnotesize{Training text-to-image samples use embedding $(1 - \alpha) \textToken_1 + \alpha \textToken_2, \, \alpha \sim \textnormal{Dir}({\color{red}1}) = {\color{red}\mathcal{U}(0, 1)}$}};
                \node[below=of img24, node distance=0cm, xshift=-.1cm, yshift=.75cm,font=\color{black}]{Latent Interpolations - relevant change in {\color{red}red}};

                \node [below=of img21, yshift=.0cm](img31){\includegraphics[trim={.0cm .0cm .0cm .0cm}, clip, width=.15\linewidth]{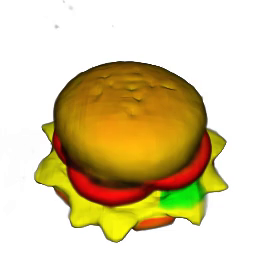}};
                \node [right=of img31, xshift=-1.5cm](img32){\includegraphics[trim={.0cm .0cm .0cm .0cm}, clip, width=.15\linewidth]{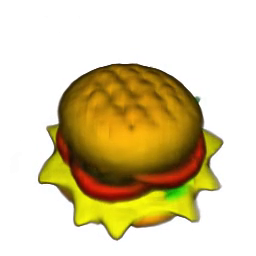}};
                \node [right=of img32, xshift=-1.5cm](img33){\includegraphics[trim={.0cm .0cm .0cm .0cm}, clip, width=.15\linewidth]{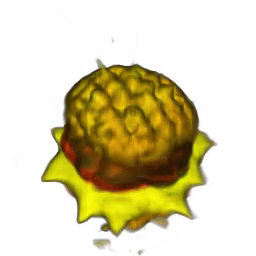}};
                \node [right=of img33, xshift=-1.5cm](img34){\includegraphics[trim={.0cm .0cm .0cm .0cm}, clip, width=.15\linewidth]{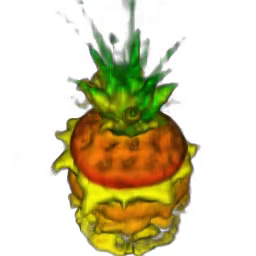}};
                \node [right=of img34, xshift=-1.5cm](img35){\includegraphics[trim={.0cm .0cm .0cm .0cm}, clip, width=.15\linewidth]{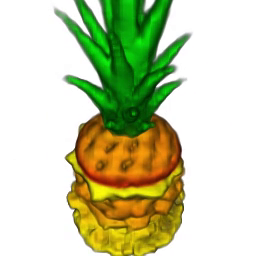}};
                \node [right=of img35, xshift=-1.5cm](img36){\includegraphics[trim={.0cm .0cm .0cm .0cm}, clip, width=.15\linewidth]{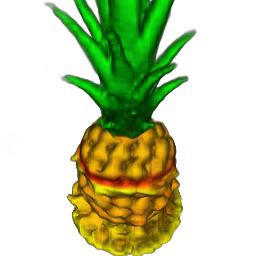}};
                \node [right=of img36, xshift=-1.5cm](img37){\includegraphics[trim={.0cm .0cm .0cm .0cm}, clip, width=.15\linewidth]{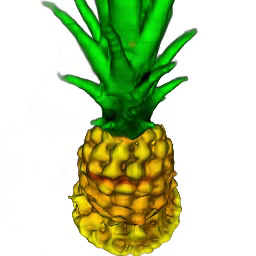}};
                \node[below=of img34, node distance=0cm, xshift=-.1cm, yshift=1.15cm,font=\color{black}]{\footnotesize{Training text-to-image samples use embedding $(1 - {\color{red}Z}) \textToken_1 + {\color{red}Z} \textToken_2$ where ${\color{red}Z \sim \textnormal{Bern}(\alpha)}, \, \alpha \sim \textnormal{Dir}({\color{red}1}) = {\color{red}\mathcal{U}(0, 1)}$}};
                \node[below=of img34, node distance=0cm, xshift=-.1cm, yshift=.75cm,font=\color{black}]{Loss Interpolations};

                \node [below=of img31] (img51){\includegraphics[trim={.0cm .0cm .0cm .0cm}, clip, width=.15\linewidth]{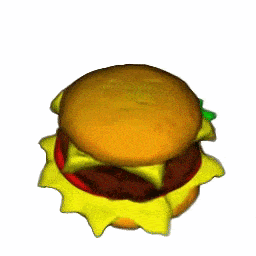}};
                \node [right=of img51, xshift=-1.5cm](img52){\includegraphics[trim={.0cm .0cm .0cm .0cm}, clip, width=.15\linewidth]{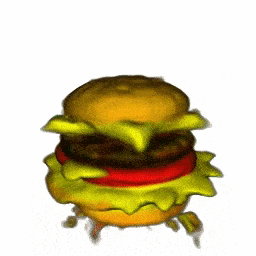}};
                \node [right=of img52, xshift=-1.5cm](img53){\includegraphics[trim={.0cm .0cm .0cm .0cm}, clip, width=.15\linewidth]{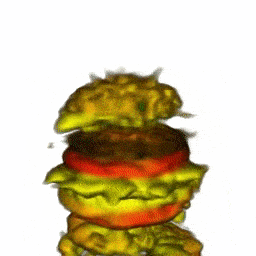}};
                \node [right=of img53, xshift=-1.5cm](img54){\includegraphics[trim={.0cm .0cm .0cm .0cm}, clip, width=.15\linewidth]{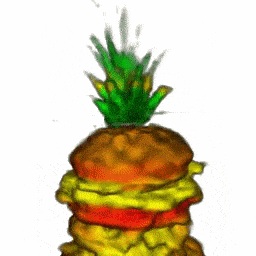}};
                \node [right=of img54, xshift=-1.5cm](img55){\includegraphics[trim={.0cm .0cm .0cm .0cm}, clip, width=.15\linewidth]{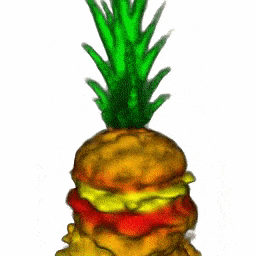}};
                \node [right=of img55, xshift=-1.5cm](img56){\includegraphics[trim={.0cm .0cm .0cm .0cm}, clip, width=.15\linewidth]{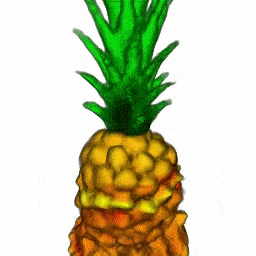}};
                \node [right=of img56, xshift=-1.5cm](img57){\includegraphics[trim={.0cm .0cm .0cm .0cm}, clip, width=.15\linewidth]{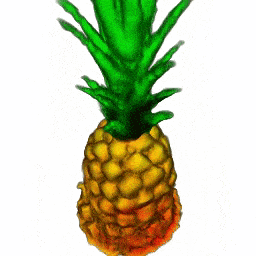}};
                \node[below=of img54, node distance=0cm, xshift=-.1cm, yshift=1.15cm,font=\color{black}]{\footnotesize{Training text-to-image samples use {\color{red}guidance weights $\omega$:} $\hat{\epsilon} = \epsilon_{\textnormal{uncond.}} + (1 - \alpha) \omega_{1} \epsilon_{\textnormal{prompt 1}} + \alpha \omega_{2} \epsilon_{\textnormal{prompt 2}}, \, \alpha \sim \textnormal{Dir}({\color{red}1}) = {\color{red}\mathcal{U}(0, 1)}$}};
                \node[below=of img54, node distance=0cm, xshift=-.1cm, yshift=.75cm,font=\color{black}]{Guidance Interpolations};

                \node[above=of img41, node distance=0cm, xshift=-.1cm, yshift=-1.15cm,font=\color{black}]{``\emph{a hamburger}"};
                \node[above=of img47, node distance=0cm, xshift=-.1cm, yshift=-1.15cm,font=\color{black}]{``\emph{a pineapple}"};
                \node[above=of img44, node distance=0cm, xshift=-.1cm, yshift=-1.15cm,font=\color{black}]{\footnotesize{Rendered frames from interpolating $\alpha$ from $0 \to 1$, after training with various objectives}};
                \node[above=of img44, node distance=0cm, xshift=-.1cm, yshift=-.75cm,font=\color{black}]{All setups use modulations from interpolated text-embeddings: $\mappingNet\left((1-\alpha) \textToken_1 + \alpha \textToken_2\right)$};
            \end{tikzpicture}
            %\vspace{-0.015\textheight}
            \caption{
                We contrast amortizing over different types of interpolations as described in Section~\ref{sec:app-interpolation-exp}.
                For all examples, we give the mapping network $\mappingNet$ the interpolated embedding $(1 - \alpha) \textToken_1 + \alpha \textToken_2$.
                However, we vary the embedding used by the text-to-image model.
                \textbf{Takeaway:}
                    We can amortize over various training methods to produce qualitatively different results.
                \emph{Top:}
                    We use no interpolants during training, which can just dissolve between the endpoints.
                \emph{Latent Interpolation:}
                    We simply interpolate between the latent embeddings used for image sampling.
                \emph{Loss Interpolation:}
                    We interpolate the loss function used in training between the prompts, producing objects simultaneously solving both losses.
                \emph{Guidance Interpolation:}
                    We interpolate the guidance weight applied to the prompts, as explored in Magic3D (without amortization)~\citep{lin2022magic3d}.
            }
            %\vspace{-0.015\textheight}
            \label{fig:interpolation_hamburgers}
        \end{figure*}

        \begin{figure*}%[ht!]
            %\vspace{-0.0225\textheight}
            \centering
            \begin{tikzpicture}
                \centering
                
                \node (img21){\includegraphics[trim={.0cm .0cm .0cm .0cm}, clip, width=.15\linewidth]{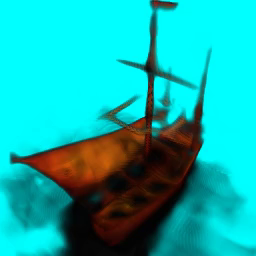}};
                \node [right=of img21, xshift=-1.5cm](img22){\includegraphics[trim={.0cm .0cm .0cm .0cm}, clip, width=.15\linewidth]{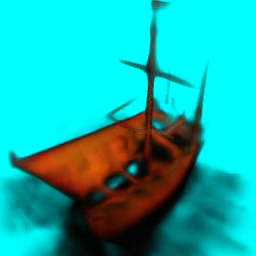}};
                \node [right=of img22, xshift=-1.5cm](img23){\includegraphics[trim={.0cm .0cm .0cm .0cm}, clip, width=.15\linewidth]{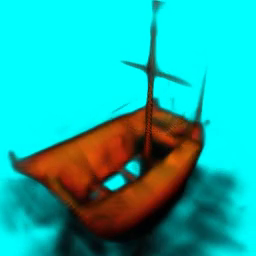}};
                \node [right=of img23, xshift=-1.5cm](img24){\includegraphics[trim={.0cm .0cm .0cm .0cm}, clip, width=.15\linewidth]{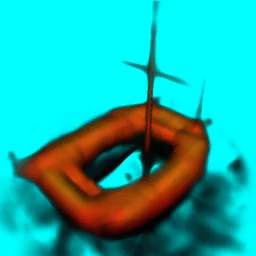}};
                \node [right=of img24, xshift=-1.5cm](img25){\includegraphics[trim={.0cm .0cm .0cm .0cm}, clip, width=.15\linewidth]{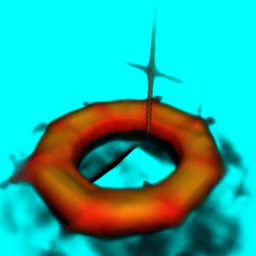}};
                \node [right=of img25, xshift=-1.5cm](img26){\includegraphics[trim={.0cm .0cm .0cm .0cm}, clip, width=.15\linewidth]{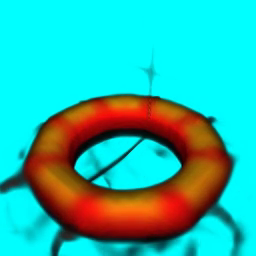}};
                \node [right=of img26, xshift=-1.5cm](img27){\includegraphics[trim={.0cm .0cm .0cm .0cm}, clip, width=.15\linewidth]{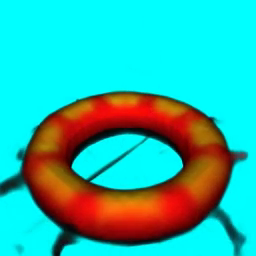}};
                \node[above=of img21, node distance=0cm, xshift=.3cm, yshift=-1.15cm,font=\color{black}]{``\emph{a wooden pirate ship}"};
                \node[above=of img27, node distance=0cm, xshift=-.1cm, yshift=-1.15cm,font=\color{black}]{``\emph{a rubber life raft}"};
                %\node[above=of img24, node distance=0cm, xshift=-.1cm, yshift=-1.15cm,font=\color{black}]{$\kappa$ from {\color{red}small} $\to$ {\color{red}large} during training};
                \node[above=of img24, node distance=0cm, xshift=-.1cm, yshift=-.6cm,font=\color{black}]{Rendered Frames Interpolating $\alpha$ from $0 \to 1$, where training $\alpha \sim \textnormal{Dir}(\kappa)$ with varying $\kappa$};

                \node [below=of img21, yshift=1.26cm](img31){\includegraphics[trim={.0cm .0cm .0cm .0cm}, clip, width=.15\linewidth]{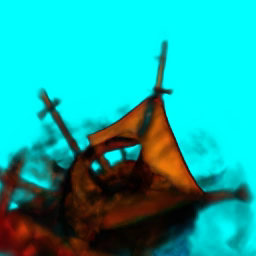}};
                \node [right=of img31, xshift=-1.5cm](img32){\includegraphics[trim={.0cm .0cm .0cm .0cm}, clip, width=.15\linewidth]{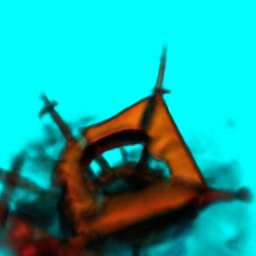}};
                \node [right=of img32, xshift=-1.5cm](img33){\includegraphics[trim={.0cm .0cm .0cm .0cm}, clip, width=.15\linewidth]{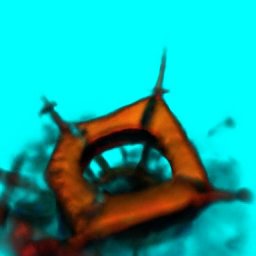}};
                \node [right=of img33, xshift=-1.5cm](img34){\includegraphics[trim={.0cm .0cm .0cm .0cm}, clip, width=.15\linewidth]{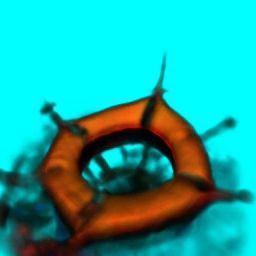}};
                \node [right=of img34, xshift=-1.5cm](img35){\includegraphics[trim={.0cm .0cm .0cm .0cm}, clip, width=.15\linewidth]{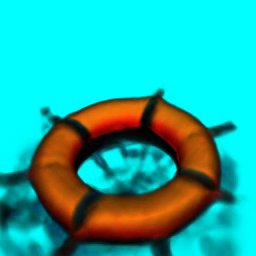}};
                \node [right=of img35, xshift=-1.5cm](img36){\includegraphics[trim={.0cm .0cm .0cm .0cm}, clip, width=.15\linewidth]{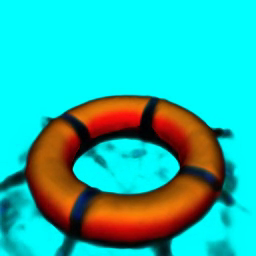}};
                \node [right=of img36, xshift=-1.5cm](img37){\includegraphics[trim={.0cm .0cm .0cm .0cm}, clip, width=.15\linewidth]{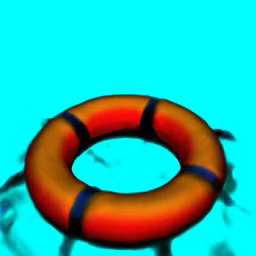}};
                %\node[below=of img34, node distance=0cm, xshift=-.1cm, yshift=1.15cm,font=\color{black}]{$\kappa$ from {\color{red}large} $\to$ {\color{red}small} during training};
                
                \node [below=of img31, yshift=1.26cm](img41){\includegraphics[trim={.0cm .0cm .0cm .0cm}, clip, width=.15\linewidth]{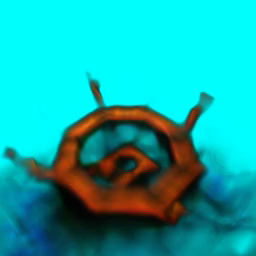}};
                \node [right=of img41, xshift=-1.5cm](img42){\includegraphics[trim={.0cm .0cm .0cm .0cm}, clip, width=.15\linewidth]{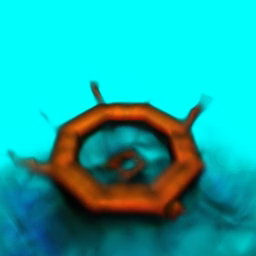}};
                \node [right=of img42, xshift=-1.5cm](img43){\includegraphics[trim={.0cm .0cm .0cm .0cm}, clip, width=.15\linewidth]{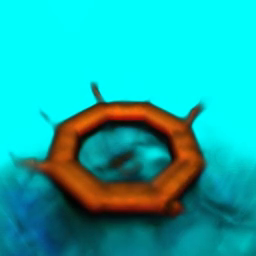}};
                \node [right=of img43, xshift=-1.5cm](img44){\includegraphics[trim={.0cm .0cm .0cm .0cm}, clip, width=.15\linewidth]{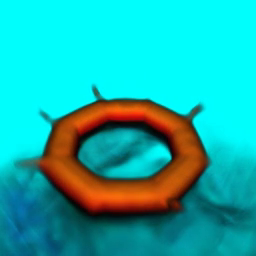}};
                \node [right=of img44, xshift=-1.5cm](img45){\includegraphics[trim={.0cm .0cm .0cm .0cm}, clip, width=.15\linewidth]{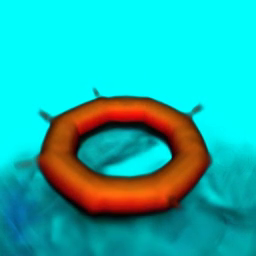}};
                \node [right=of img45, xshift=-1.5cm](img46){\includegraphics[trim={.0cm .0cm .0cm .0cm}, clip, width=.15\linewidth]{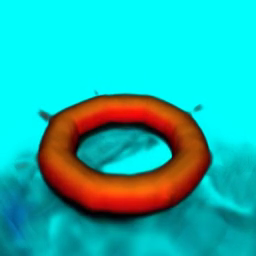}};
                \node [right=of img46, xshift=-1.5cm](img47){\includegraphics[trim={.0cm .0cm .0cm .0cm}, clip, width=.15\linewidth]{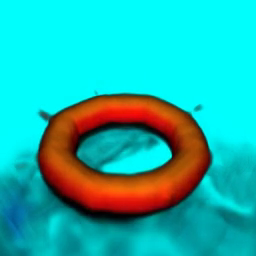}};
                %\node[below=of img34, node distance=0cm, xshift=-.1cm, yshift=1.15cm,font=\color{black}]{$\kappa$ from {\color{red}large} $\to$ {\color{red}small} during training};
                %\node[below=of img31, node distance=0cm, xshift=-.1cm, yshift=1.15cm,font=\color{black}]{``\emph{a baby dragon}"};
                %\node[below=of img37, node distance=0cm, xshift=-.1cm, yshift=1.15cm,font=\color{black}]{``\emph{a green dragon}"};

                \node[left=of img31, node distance=0cm, rotate=90, xshift=1.4cm, yshift=-.5cm, font=\color{black}]{$\kappa$ during training};
                \node[left=of img21, node distance=0cm, rotate=90, xshift=1.3cm, yshift=-.9cm,  font=\color{black}]{small $\to$ large};
                \node[left=of img31, node distance=0cm, rotate=90, xshift=1.6cm, yshift=-.9cm,  font=\color{black}]{\scriptsize{$\kappa = \num{1} \implies \alpha \sim \mathcal{U}(\num{0}, \num{1})$}};
                \node[left=of img41, node distance=0cm, rotate=90, xshift=.8cm, yshift=-.9cm,  font=\color{black}]{large $\to$ small};
            \end{tikzpicture}
            %\vspace{-0.015\textheight}
            \caption{
                We display the results for differing strategies for changing the concentration parameter $\kappa$ for the distribution of the interpolation weights $\alpha$.
                Note that a concentration of $\kappa = \num{1}$ is simply a uniform distribution: $\textnormal{Dir}(\num{1}) = \mathcal{U}(0,1)$.
                For both results, we train for $\num{5000}$ steps with an initial concentration $\kappa$, which we then change for the final $\num{5000}$ steps.
                \textbf{Takeaway:}
                    The initial shapes learned strongly influence subsequent training, and a ``large" concentration $\kappa$ focuses on the midpoint, while a ``small" concentration focuses on the endpoints.
                    If we want the original prompts in the interpolation, then we should start with $\kappa$ small, while if we desire a steering-wheel-life-raft satisfying both losses, we should start with $\kappa$ large.
            }
            %\vspace{-0.015\textheight}
            \label{fig:interpolation_ships}
        \end{figure*}
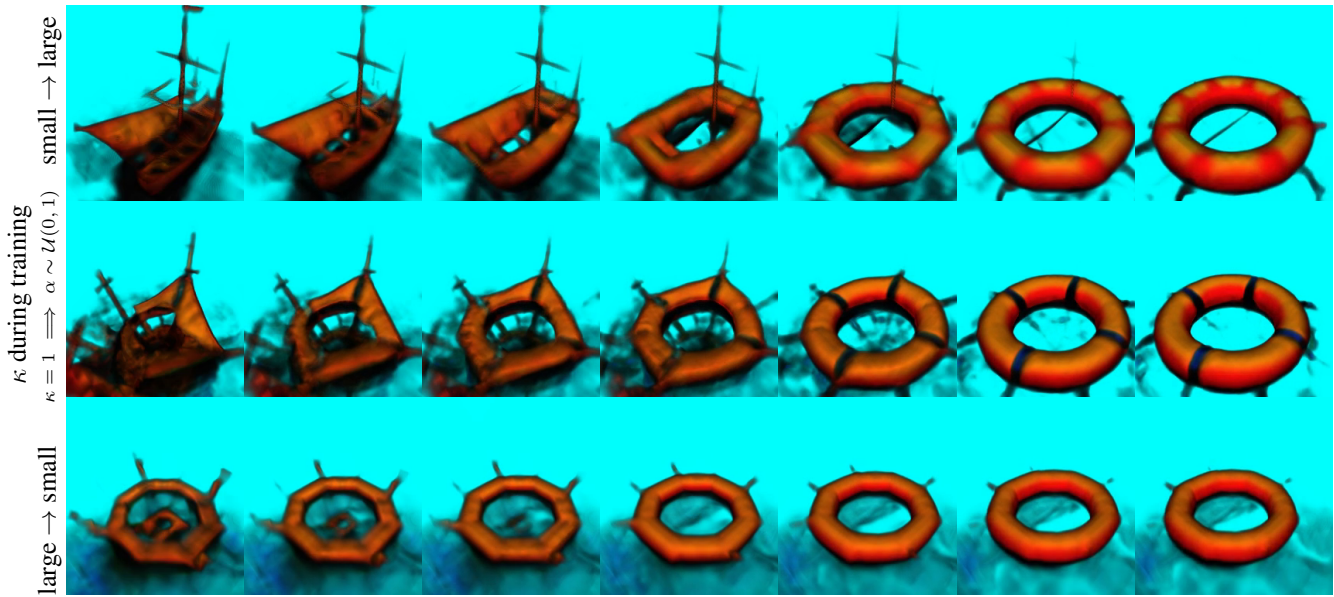

        \begin{figure*}%[ht!]
            \vspace{-0.03\textheight}
            \centering
            \begin{tikzpicture}
                \centering
                % \node [right=of img81, xshift=-1.26cm](img82){\includegraphics[trim={.0cm .3cm .0cm .3cm}, clip, width=.142\linewidth]{images/interpolation_trees/frame1.png}};
                
                \node [below=of img21, yshift=.65cm](img31){\includegraphics[trim={.0cm .3cm .0cm .3cm}, clip, width=.142\linewidth]{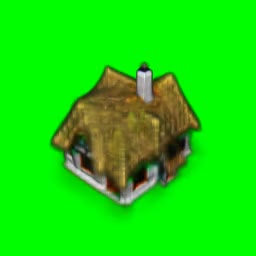}};
                \node [right=of img31, xshift=-1.26cm](img32){\includegraphics[trim={.0cm .3cm .0cm .3cm}, clip, width=.142\linewidth]{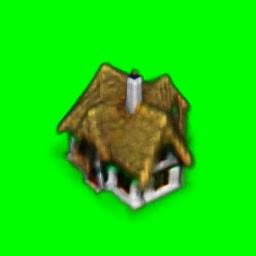}};
                \node [right=of img32, xshift=-1.26cm](img33){\includegraphics[trim={.0cm .3cm .0cm .3cm}, clip, width=.142\linewidth]{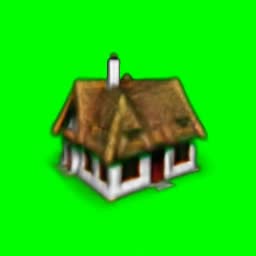}};
                \node [right=of img33, xshift=-1.26cm](img34){\includegraphics[trim={.0cm .3cm .0cm .3cm}, clip, width=.142\linewidth]{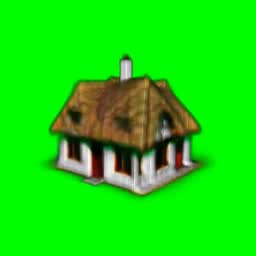}};
                \node [right=of img34, xshift=-1.26cm](img35){\includegraphics[trim={.0cm .3cm .0cm .3cm}, clip, width=.142\linewidth]{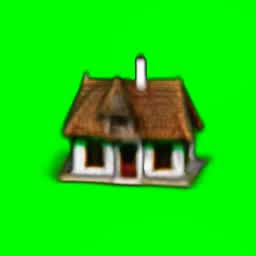}};
                \node [right=of img35, xshift=-1.26cm](img36){\includegraphics[trim={.0cm .3cm .0cm .3cm}, clip, width=.142\linewidth]{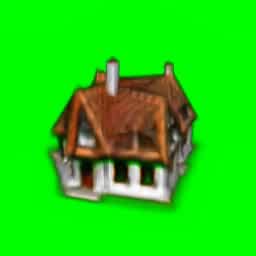}};
                \node [right=of img36, xshift=-1.26cm](img37){\includegraphics[trim={.0cm .3cm .0cm .3cm}, clip, width=.142\linewidth]{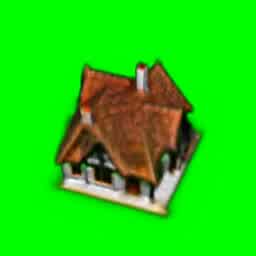}};
                \node[above=of img31, node distance=0cm, xshift=2.1cm, yshift=-1.2cm,font=\color{black}]{``\emph{... an adorable cottage with a thatched roof}"};
                \node[above=of img37, node distance=0cm, xshift=-1.0cm, yshift=-1.2cm,font=\color{black}]{``\emph{...  a house in Tudor Style}"};

                \node [below=of img31, yshift=.65cm](img51){\includegraphics[trim={.0cm .3cm .0cm .3cm}, clip, width=.142\linewidth]{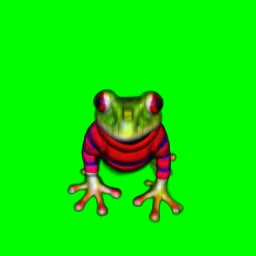}};
                \node [right=of img51, xshift=-1.26cm](img52){\includegraphics[trim={.0cm .3cm .0cm .3cm}, clip, width=.142\linewidth]{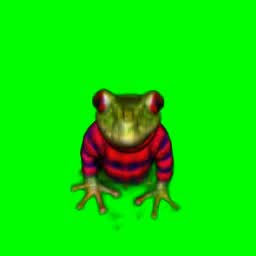}};
                \node [right=of img52, xshift=-1.26cm](img53){\includegraphics[trim={.0cm .3cm .0cm .3cm}, clip, width=.142\linewidth]{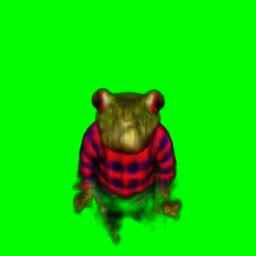}};
                \node [right=of img53, xshift=-1.26cm](img54){\includegraphics[trim={.0cm .3cm .0cm .3cm}, clip, width=.142\linewidth]{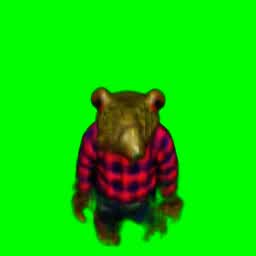}};
                \node [right=of img54, xshift=-1.26cm](img55){\includegraphics[trim={.0cm .3cm .0cm .3cm}, clip, width=.142\linewidth]{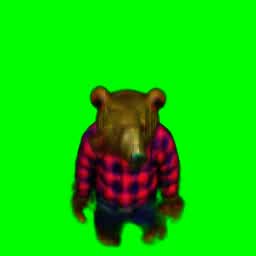}};
                \node [right=of img55, xshift=-1.26cm](img56){\includegraphics[trim={.0cm .3cm .0cm .3cm}, clip, width=.142\linewidth]{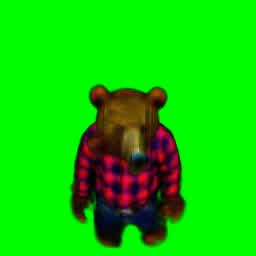}};
                \node [right=of img56, xshift=-1.26cm](img57){\includegraphics[trim={.0cm .3cm .0cm .3cm}, clip, width=.142\linewidth]{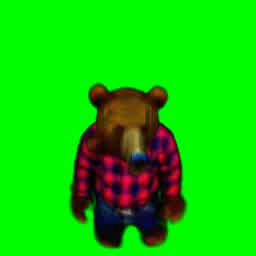}};
                \node[above=of img51, node distance=0cm, xshift=.75cm, yshift=-1.2cm,font=\color{black}]{``\emph{a frog wearing a sweater}"};
                \node[above=of img57, node distance=0cm, xshift=-1.25cm, yshift=-1.2cm,font=\color{black}]{``\emph{a bear dressed as a lumberjack}"};

                \node [below=of img51, yshift=.65cm](img511){\includegraphics[trim={.0cm .3cm .0cm .3cm}, clip, width=.142\linewidth]{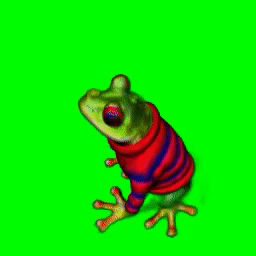}};
                \node [right=of img511, xshift=-1.26cm](img521){\includegraphics[trim={.0cm .3cm .0cm .3cm}, clip, width=.142\linewidth]{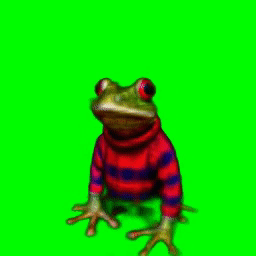}};
                \node [right=of img521, xshift=-1.26cm](img531){\includegraphics[trim={.0cm .3cm .0cm .3cm}, clip, width=.142\linewidth]{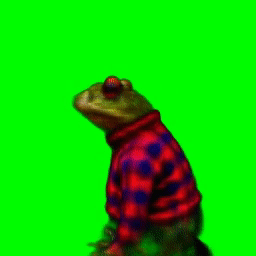}};
                \node [right=of img531, xshift=-1.26cm](img541){\includegraphics[trim={.0cm .3cm .0cm .3cm}, clip, width=.142\linewidth]{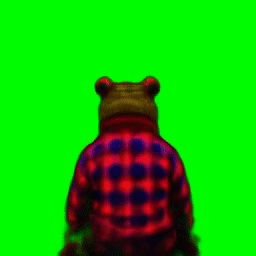}};
                \node [right=of img541, xshift=-1.26cm](img551){\includegraphics[trim={.0cm .3cm .0cm .3cm}, clip, width=.142\linewidth]{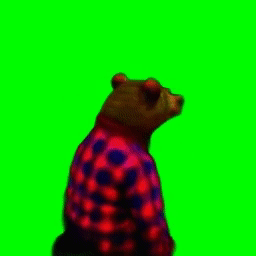}};
                \node [right=of img551, xshift=-1.26cm](img561){\includegraphics[trim={.0cm .3cm .0cm .3cm}, clip, width=.142\linewidth]{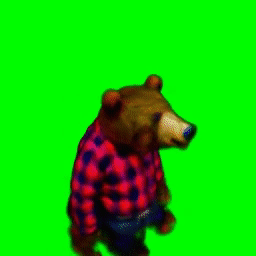}};
                \node [right=of img561, xshift=-1.26cm](img571){\includegraphics[trim={.0cm .3cm .0cm .3cm}, clip, width=.142\linewidth]{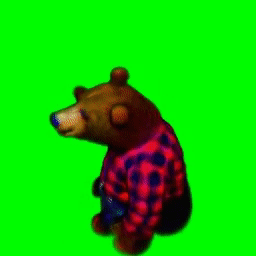}};
                % \node[above=of img511, node distance=0cm, xshift=.75cm, yshift=-1.2cm,font=\color{black}]{``\emph{a frog wearing a sweater}"};
                % \node[above=of img571, node distance=0cm, xshift=-1.25cm, yshift=-1.2cm,font=\color{black}]{``\emph{a bear dressed as a lumberjack}"};

                \node [below=of img511, yshift=.65cm](img71){\includegraphics[trim={.0cm .3cm .0cm .3cm}, clip, width=.142\linewidth]{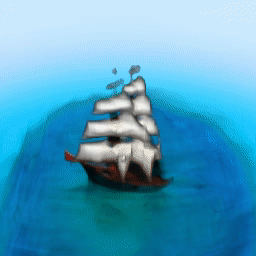}};
                \node [right=of img71, xshift=-1.26cm](img72){\includegraphics[trim={.0cm .3cm .0cm .3cm}, clip, width=.142\linewidth]{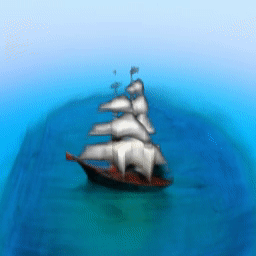}};
                \node [right=of img72, xshift=-1.26cm](img73){\includegraphics[trim={.0cm .3cm .0cm .3cm}, clip, width=.142\linewidth]{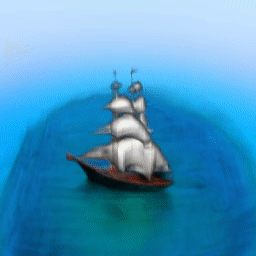}};
                \node [right=of img73, xshift=-1.26cm](img74){\includegraphics[trim={.0cm .3cm .0cm .3cm}, clip, width=.142\linewidth]{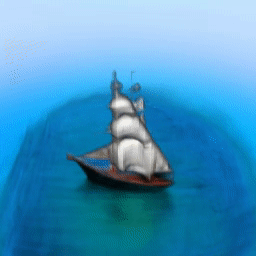}};
                \node [right=of img74, xshift=-1.26cm](img75){\includegraphics[trim={.0cm .3cm .0cm .3cm}, clip, width=.142\linewidth]{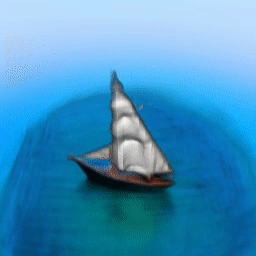}};
                \node [right=of img75, xshift=-1.26cm](img76){\includegraphics[trim={.0cm .3cm .0cm .3cm}, clip, width=.142\linewidth]{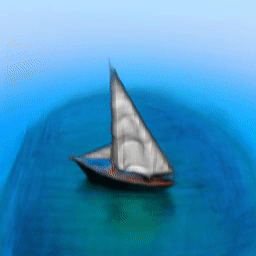}};
                \node [right=of img76, xshift=-1.26cm](img77){\includegraphics[trim={.0cm .3cm .0cm .3cm}, clip, width=.142\linewidth]{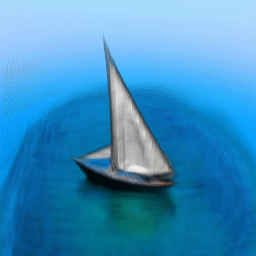}};
                \node[above=of img71, node distance=0cm, xshift=.55cm, yshift=-1.2cm,font=\color{black}]{``\emph{... a majestic sailboat}"};
                \node[above=of img77, node distance=0cm, xshift=-.55cm, yshift=-1.2cm,font=\color{black}]{``\emph{a spanish galleon...}"};

                \node [below=of img71, yshift=.65cm](img61){\includegraphics[trim={.0cm .3cm .0cm .3cm}, clip, width=.142\linewidth]{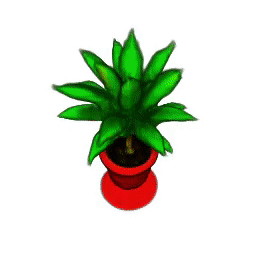}};
                \node [right=of img61, xshift=-1.26cm](img62){\includegraphics[trim={.0cm .3cm .0cm .3cm}, clip, width=.142\linewidth]{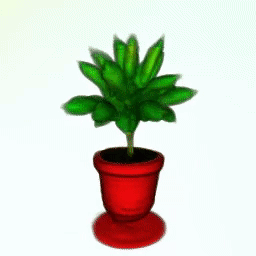}};
                \node [right=of img62, xshift=-1.26cm](img63){\includegraphics[trim={.0cm .3cm .0cm .3cm}, clip, width=.142\linewidth]{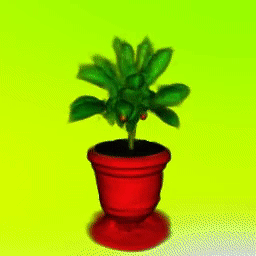}};
                \node [right=of img63, xshift=-1.26cm](img64){\includegraphics[trim={.0cm .3cm .0cm .3cm}, clip, width=.142\linewidth]{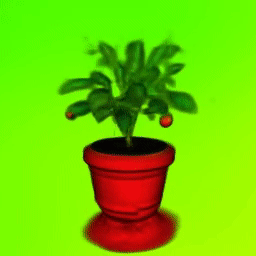}};
                \node [right=of img64, xshift=-1.26cm](img65){\includegraphics[trim={.0cm .3cm .0cm .3cm}, clip, width=.142\linewidth]{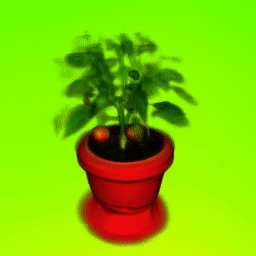}};
                \node [right=of img65, xshift=-1.26cm](img66){\includegraphics[trim={.0cm .3cm .0cm .3cm}, clip, width=.142\linewidth]{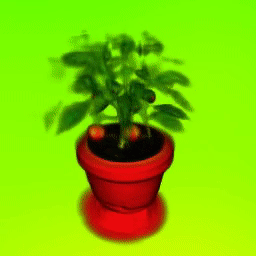}};
                \node [right=of img66, xshift=-1.26cm](img67){\includegraphics[trim={.0cm .3cm .0cm .3cm}, clip, width=.142\linewidth]{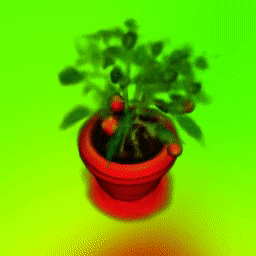}};
                \node[above=of img61, node distance=0cm, xshift=.75cm, yshift=-1.2cm,font=\color{black}]{``\emph{a ficus planted in a pot}"};
                \node[above=of img67, node distance=0cm, xshift=-1.25cm, yshift=-1.2cm,font=\color{black}]{``\emph{a small cherry tomato plant...}"};

                \node [below=of img61, yshift=.65cm](img41){\includegraphics[trim={.0cm .3cm .0cm .3cm}, clip, width=.142\linewidth]{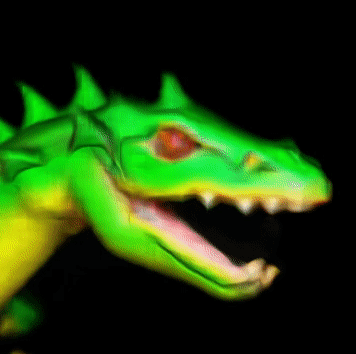}};
                \node [right=of img41, xshift=-1.26cm](img42){\includegraphics[trim={.0cm .3cm .0cm .3cm}, clip, width=.142\linewidth]{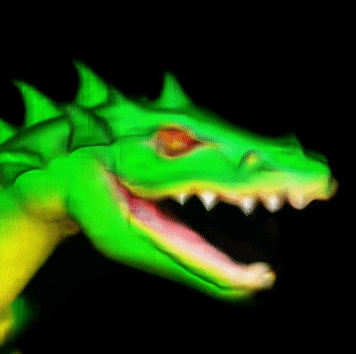}};
                \node [right=of img42, xshift=-1.26cm](img43){\includegraphics[trim={.0cm .3cm .0cm .3cm}, clip, width=.142\linewidth]{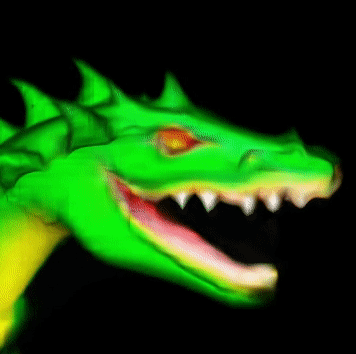}};
                \node [right=of img43, xshift=-1.26cm](img44){\includegraphics[trim={.0cm .3cm .0cm .3cm}, clip, width=.142\linewidth]{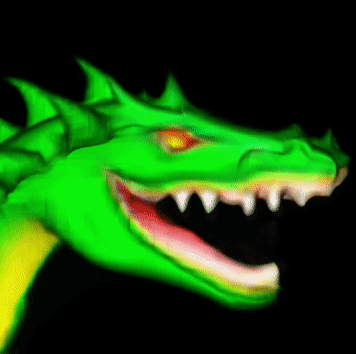}};
                \node [right=of img44, xshift=-1.26cm](img45){\includegraphics[trim={.0cm .3cm .0cm .3cm}, clip, width=.142\linewidth]{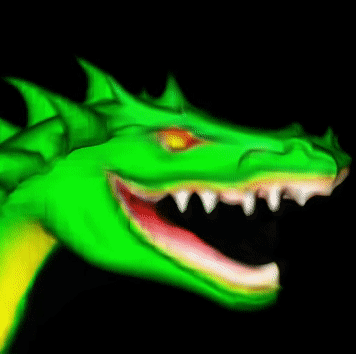}};
                \node [right=of img45, xshift=-1.26cm](img46){\includegraphics[trim={.0cm .3cm .0cm .3cm}, clip, width=.142\linewidth]{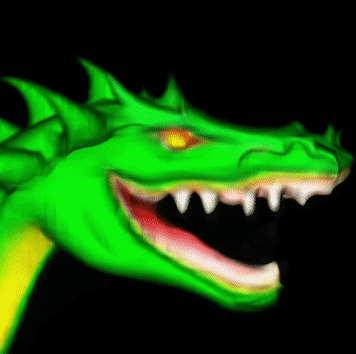}};
                \node [right=of img46, xshift=-1.26cm](img47){\includegraphics[trim={.0cm .3cm .0cm .3cm}, clip, width=.142\linewidth]{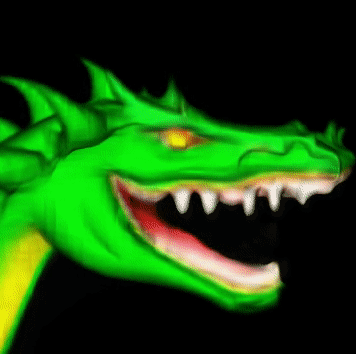}};
                \node[above=of img41, node distance=0cm, xshift=.15cm, yshift=-1.2cm,font=\color{black}]{``\emph{a baby dragon}"};
                \node[above=of img47, node distance=0cm, xshift=-.15cm, yshift=-1.2cm,font=\color{black}]{``\emph{a green dragon}"};

                \node [below=of img41, yshift=.65cm](img21){\includegraphics[trim={.0cm .3cm .0cm .3cm}, clip, width=.142\linewidth]{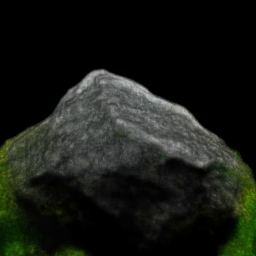}};
                \node [right=of img21, xshift=-1.26cm](img22){\includegraphics[trim={.0cm .3cm .0cm .3cm}, clip, width=.142\linewidth]{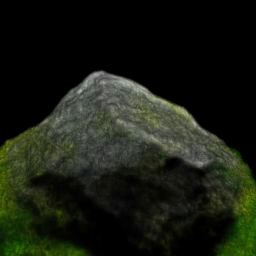}};
                \node [right=of img22, xshift=-1.26cm](img23){\includegraphics[trim={.0cm .3cm .0cm .3cm}, clip, width=.142\linewidth]{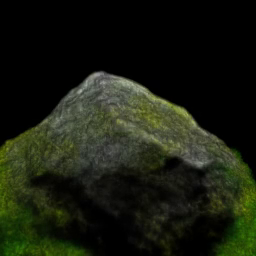}};
                \node [right=of img23, xshift=-1.26cm](img24){\includegraphics[trim={.0cm .3cm .0cm .3cm}, clip, width=.142\linewidth]{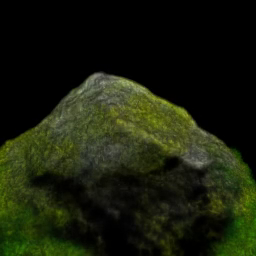}};
                \node [right=of img24, xshift=-1.26cm](img25){\includegraphics[trim={.0cm .3cm .0cm .3cm}, clip, width=.142\linewidth]{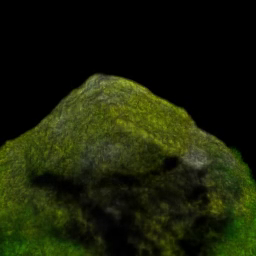}};
                \node [right=of img25, xshift=-1.26cm](img26){\includegraphics[trim={.0cm .3cm .0cm .3cm}, clip, width=.142\linewidth]{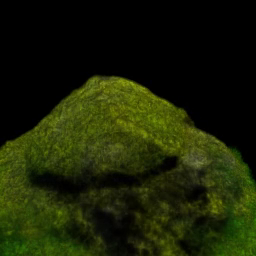}};
                \node [right=of img26, xshift=-1.26cm](img27){\includegraphics[trim={.0cm .3cm .0cm .3cm}, clip, width=.142\linewidth]{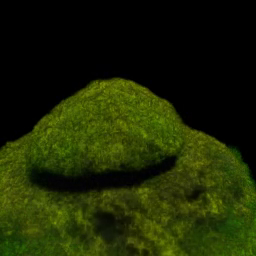}};
                \node[above=of img21, node distance=0cm, xshift=-.15cm, yshift=-1.2cm,font=\color{black}]{``\emph{jagged rock}"};
                \node[above=of img27, node distance=0cm, xshift=-.0cm, yshift=-1.2cm,font=\color{black}]{``\emph{mossy rock}"};

                % \node [below=of img21, yshift=1.26cm](img31){\includegraphics[trim={.0cm .3cm .0cm .3cm}, clip, width=.142\linewidth]{images/interpolation_set_3/frame0.png}};
                % \node [right=of img31, xshift=-1.26cm](img32){\includegraphics[trim={.0cm .3cm .0cm .3cm}, clip, width=.142\linewidth]{images/interpolation_set_3/frame1.png}};
                % \node [right=of img32, xshift=-1.26cm](img33){\includegraphics[trim={.0cm .3cm .0cm .3cm}, clip, width=.142\linewidth]{images/interpolation_set_3/frame2.png}};
                % \node [right=of img33, xshift=-1.26cm](img34){\includegraphics[trim={.0cm .3cm .0cm .3cm}, clip, width=.142\linewidth]{images/interpolation_set_3/frame3.png}};
                % \node [right=of img34, xshift=-1.26cm](img35){\includegraphics[trim={.0cm .3cm .0cm .3cm}, clip, width=.142\linewidth]{images/interpolation_set_3/frame4.png}};
                % \node [right=of img35, xshift=-1.26cm](img36){\includegraphics[trim={.0cm .3cm .0cm .3cm}, clip, width=.142\linewidth]{images/interpolation_set_3/frame5.png}};
                % \node [right=of img36, xshift=-1.26cm](img37){\includegraphics[trim={.0cm .3cm .0cm .3cm}, clip, width=.142\linewidth]{images/interpolation_set_3/frame6.png}};
                % \node[below=of img31, node distance=0cm, xshift=-.1cm, yshift=1.15cm,font=\color{black}]{``\emph{a baby dragon}"};
                % \node[below=of img37, node distance=0cm, xshift=-.1cm, yshift=1.15cm,font=\color{black}]{``\emph{a green dragon}"};
            \end{tikzpicture}
            %\vspace{-0.015\textheight}
            \caption{
                We include additional results for using our method to amortize over (loss) interpolants between prompts.
                We alternate between a fixed and varied camera view.
                We show examples of varied buildings, characters, vehicles, plants, landscapes, or a simple animation of ``\emph{a baby dragon}" aging into an adult.
                % \TODO{Fix spacing to match main and render fixed + multi-view}
                % We display interpolations in the text embedding $\textToken$ for a model trained with our amortized method.
                % We show our models output -- with no additional training -- with interpolated embedding $(1 - \alpha) \textToken_1 + \alpha \textToken_2$.
                % \emph{Top:} We generate a continuum of novel landscape assets.
                % \emph{Bottom:} We create a simplistic animation of ``\emph{a baby dragon}" maturing into ``\emph{a green dragon}."
            }
            %\vspace{-0.015\textheight}
            \label{fig:interpolation_other}
        \end{figure*}
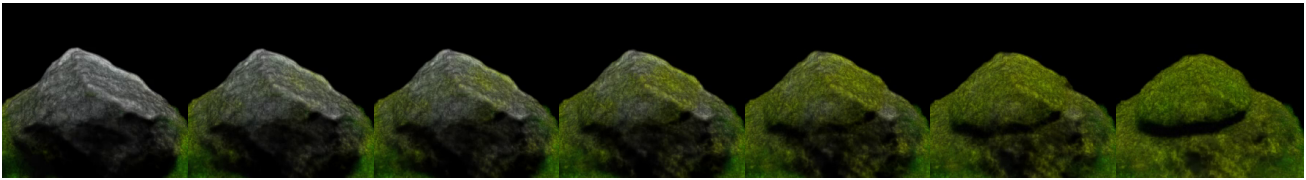

\end{document}